\documentclass[12pt]{article}
\usepackage[natbibapa]{apacite}
\usepackage{amsmath}
\usepackage{times}
\usepackage{graphicx}
\usepackage{color}
\usepackage{tabularx}
\newcolumntype{Y}{>{\centering\arraybackslash}X}
\usepackage{multirow}
\usepackage{rotating}
\usepackage{bbm}
\usepackage{latexsym}
\usepackage{hyperref}

\usepackage{calc}

\usepackage{setspace}
\usepackage{caption}
\newcommand{\captionfonts}{\normalsize}

\makeatletter  
\long\def\@makecaption#1#2{%
  \vskip\abovecaptionskip
  \sbox\@tempboxa{{\captionfonts #1: #2}}%
  \ifdim \wd\@tempboxa >\hsize
    {\captionfonts #1: #2\par}
  \else
    \hbox to\hsize{\hfil\box\@tempboxa\hfil}%
  \fi
  \vskip\belowcaptionskip}
\makeatother   
\usepackage{enumitem}

\usepackage{graphicx}
\usepackage{epstopdf}
\usepackage{subcaption}
\usepackage{appendix}

\usepackage{mathtools}
\usepackage{amsmath}
\usepackage{siunitx} %for degree sign

\graphicspath{{images/}}

\renewcommand{\thefootnote}{\normalsize \fnsymbol{footnote}} 

\usepackage{fancyhdr}
\begin{document}

% \listoftodos[Notes]

\newpage

\hspace{13.9cm}1

\ \vspace{20mm}\\

\begin{doublespace}
{\LARGE How Convolutional Neural Network Architecture Biases Learned Opponency and Colour Tuning}
\end{doublespace}
% \renewcommand{\baselinestretch}{1}
% Uncovering the connection between architecture and early processing in deep CNNs
% Characterising spatial and spectral processing in convolutional neural networks: a physiological perspective
% A physiological perspective on deep CNNs: how architecture influences function
% A physiologically inspired analysis of deep CNNs: how architecture influences learned spatial and spectral response
% How convolutional neural architecture influences learned spatial and spectral response

\ \\
{\bf \large Ethan Harris$\footnotemark[1]\footnotemark[2]$, Daniela Mihai$\footnotemark[1]\footnotemark[2]$, Jonathon Hare$\footnotemark[1]\footnotemark[2]$}\\
{$\footnotemark[1]$Equal contribution}\\
{$\footnotemark[2]$Vision Learning and Control, Electronics and Computer Science, University of Southampton.}\\

\renewcommand{\thefootnote}{\arabic{footnote}} 

%\ \\[-2mm]

\thispagestyle{fancy}
\markboth{}{How Convolutional Neural Network Architecture Biases Learned Opponency and Colour Tuning}
\ \vspace{-0mm}\\
%
%Abstract
\begin{center} {\bf Abstract} \end{center}
Recent work suggests that changing Convolutional Neural Network (CNN) architecture by introducing a bottleneck in the second layer can yield changes in learned function.
To understand this relationship fully requires a way of quantitatively comparing trained networks.
The fields of electrophysiology and psychophysics have developed a wealth of methods for characterising visual systems which permit such comparisons.
Inspired by these methods, we propose an approach to obtaining spatial and colour tuning curves for convolutional neurons, which can be used to classify cells in terms of their spatial and colour opponency.
We perform these classifications for a range of CNNs with different depths and bottleneck widths.
Our key finding is that networks with a bottleneck show a strong functional organisation: almost all cells in the bottleneck layer become both spatially and colour opponent, cells in the layer following the bottleneck become non-opponent.
The colour tuning data can further be used to form a rich understanding of how colour is encoded by a network.
As a concrete demonstration, we show that shallower networks without a bottleneck learn a complex non-linear colour system, whereas deeper networks with tight bottlenecks learn a simple channel opponent code in the bottleneck layer.
We further develop a method of obtaining a hue sensitivity curve for a trained CNN which enables high level insights that complement the low level findings from the colour tuning data.
We go on to train a series of networks under different conditions to ascertain the robustness of the discussed results.
Ultimately, our methods and findings coalesce with prior art, strengthening our ability to interpret trained CNNs and furthering our understanding of the connection between architecture and learned representation.
Trained models and code for all experiments are available at \url{https://github.com/ecs-vlc/opponency}.

%%%%%%%%%%%

\section{Introduction}

The tendency for learning machines to exhibit oriented-edge receptive fields, similar to those found in nature, has long been observed \citep{lehky1988network,olshausen1996emergence,BELL19973327,NIPS2006_3018,Wang2015ModelingTO,krizhevsky2012imagenet,lindsey2019unified,olah2020overview}. However, learning machines rarely exhibit the functional organisation found in nature. In convolutional neural networks, we typically find oriented-edge receptive fields in early layers, rather than a progression from centre-surround receptive fields to oriented-edge receptive fields as is common in biological vision \citep{hubel2004brain}. In an important work, \citet{lindsey2019unified} demonstrate that the addition of a bottleneck to a deep convolutional network can induce centre-surround receptive fields, suggesting a causal link between anatomical constraints and the nature of learned visual processing.
In order to refine our understanding of this causal relationship, we pursue an electrophysiological interpretation of convolutional networks which incorporates opponency and colour tuning.
% cells with centre-surround or oriented edge receptive fields are examples of spatially opponent cells
% spatially opponent  opponency one of the many important functional archetypes that have long been observed in nature.
% At present,
% % a lack of methods for automatic classification of pertinent cell types limits the scope of studies in this space,
% practitioners must resort to either qualitative analyses \citep{lindsey2019unified} or manual annotation \citep{olah2020overview}.
% This prevents the large scale analysis that would be required to fully understand the relationship between data, architecture, and learned representation. In this work, we seek biologically inspired methods for automatically constructing informative descriptors of the function learned by a convolutional network.

Cells with centre-surround and oriented edge receptive fields are spatially opponent. From the classic work of \citet{kuffler1953discharge}, \citet{hubel1962receptive,hubel2004brain} and others (summarised in \citet{troy2002receptive} and \citet{martinez2003complex}), these neurons form the building blocks of feature extraction in the primary visual cortex. Formally, a neuron that is excited by a particular stimulus and inhibited by another in the same stimulus space is said to be opponent to that space. For example, if a neuron is excited by stimulus in some part of the visual field and inhibited in another, it is spatially opponent. Alternatively, if a neuron is excited by stimulus of a certain wavelength and inhibited by stimulus of another, it is spectrally opponent.
Spectral opponency, first hinted at by the complementary colour system from \citet{goethe1840theory} and later detailed by \citet{hering1920grundzuge}, was only observed and characterised at a cellular level around 1960 \citep{de1958response,wagner1960opponent,wiesel1966spatial,naka1966s,daw1967goldfish}. Combined, the theories of spatially opponent feature extraction in the visual cortex \citep{kuffler1953discharge,hubel1962receptive,hubel2004brain,troy2002receptive}, trichromacy \citep{young1802ii,helmholtz1852lxxxi,maxwell1860iv}, and spectral opponency \citep{DeValois:66} constitute a deep understanding of the early layers of visual processing in nature.
% However, despite the sheer number of experimental discoveries, we have only a sparse view of the space of visual intelligence in the natural world. This limits our ability to consider precisely which physiological or environmental differences lead to the variation in visual processing across different species.

The notional elegance of the above theories has served to motivate much of the progress made in computer vision, most notably including the development of multi-layer (deep) Convolutional Neural Networks (CNNs) \citep{le1990handwritten,bottou1994comparison,lecun1995convolutional} that are now so focal in our collective interests. Multi-layer CNNs are learning models designed to mimic the functional properties, namely spatial feature extraction and retinotopy, of the retina, Lateral Geniculate Nucleus (LGN), and primary visual cortex. By virtue of the ease with which one can train such models, multi-layer CNNs offer a unique opportunity to study the emergence of visual phenomena across the full gamut of constraints and conditions of interest. It is widely observed that trained convolutional neurons experience the same kinds of receptive fields as those found in nature, and that the learned features become successively more abstract with depth \citep{krizhevsky2012imagenet,zeiler2014visualizing,olah2017feature,olah2020overview}. However, we do not typically see structural organisation of these cell types. For example, edge and colour information is confounded in the first layer of ZFNet \citep{zeiler2014visualizing}, with some colour information also encoded in the second layer. Furthermore, as addressed by \citet{lindsey2019unified}, none of the convolutional neurons have centre-surround receptive fields of the kind observed in retinal ganglion cells.
\citet{rafegas2018color} analysed colour selectivity in a deep CNN, finding cells which are excited by two groups of stimuli that are roughly opposite in hue. To classify these cells as opponent would additionally require an understanding of the stimuli which inhibit each cell. 
There has been some exploration of the role of inhibition in deep CNNs
% There is some evidence that inhibition plays an important role in deep CNNs 
\citep{olah2018building}, although we are not aware of any demonstration that learned convolutional cells are ever truly opponent in the sense that they are both inhibited below and excited above a baseline by some stimuli.
% \todo[inline]{I think we also need to address counter arguments against CNNS - like that paper I linked to in messenger; we must make it clear that we're not so much (a) interested in trying to build deep models that have superhuman performance, or (b) models that completely \emph{model} the brain. Our goal is to explore what factors give rise to certain features that are observed in real brains though simple models of small parts of neuroanatomy}

With the exception of recent developments in meta learning \citep[e.g.][]{zoph2016neural,tan2019efficientnet}, new convolutional architectures are typically designed with the aim of increasing either width \citep{zagoruyko2016wide} or depth \citep{szegedy2015going,he2016deep} whilst preventing the vanishing gradient problem with: auxiliary losses \citep{szegedy2015going}, skip connections \citep{he2016deep}, dense connections \citep{huang2017densely}, or stochastic depth \citep{huang2016deep} to name a few. However, the finding by \citet{lindsey2019unified} that network architecture can impact the fundamental `type' of function that is learned (rather than simply affecting capacity) suggests a new approach to both architecture design and interpretability. Specifically, if we can improve our understanding of the bias introduced by the network architecture, we may be able to design new architectures with specific goals in mind or better interpret the performance of pre-existing ones.

% In vision science, much progress has been made through the study of visual illusions. The work of \citet{purves2011we} presents a strong suggestion that the specifics of colour processing are heavily influenced by the statistics of visual stimuli in the natural world. For example, the variation in perceived line length as a function of angle described by \citet{pollock1952apparent} can be explained by the statistics of lines in natural scenes \citep{howe2002range}. \citet{gomez2018convolutional} find evidence that CNNs are susceptible to the same visual illusions as those that fool human observers. This again suggests that the subtleties of visual intelligence are born out of a need to perform certain visual tasks well in a particular data space. Elsewhere in the deep learning literature, illusory stimuli are typically seen as adversarial examples which can be used to `fool' a network into giving an incorrect output \citep{goodfellow2014explaining}. A better understanding of the relationship between the data space and the type of function that is learned could lead to more efficient methods of increasing the robustness of learning machines (\citet{goodfellow2014explaining} suggest the costly approach of training the network on its own adversarial examples).

Clearly, research in this space has the potential to impact our understanding of both deep learning and the neuroscience of vision.
In order to realise this potential, large-scale studies are needed which properly establish the connections between the model architecture, the data space, and the kind of visual processing that is learned.
% Presently, the barrier to conducting such studies is the lack of automated approaches to network analysis. For example,
\citet{lindsey2019unified} mainly rely on qualitative assessment for the identification of centre-surround and oriented edge receptive fields,
% Although this is sufficient for the observation that centre-surround receptive fields are obtained, it
% This does not scale well or enable a more detailed assessment which would be required to, for example, compare models trained on different data sets.
% That said, \citet{lindsey2019unified}
but do propose some quantitative analyses such as the variance in gradient with respect to different inputs as a measure of the linearity of the neuron. The highly detailed analyses of \citet{olah2020zoom} give a comprehensive understanding of the function of particular neurons or circuits in deep networks, however, each functional unit or group is currently identified manually. The procedure proposed by \citet{rafegas2018color} could be automated but involves the costly process of determining image patches which most excite each cell. The Brain-Score project from \citet{SchrimpfKubilius2018BrainScore} is an attempt at providing an assessment of the similarity between a given network and various neural and behavioural recordings from primates.
% Because the Brain-Score relies purely on similarity measures it is
This is uninformative in the sense that it does not provide any information regarding precisely how the function of the network is similar to that of the primate visual system. The same could be said of the work of \citet{gomez2018convolutional}, who find evidence that CNNs are susceptible to the same visual illusions as those that fool human observers.

In this paper, we develop a framework for automatically classifying convolutional cells in terms of their spatial and colour opponency, based on electrophysiological definitions from the neuroscience literature. In addition, we propose a method of obtaining a hue sensitivity curve for a given network, inspired by similar methods in psychophysics.
Combined, these approaches provide a descriptor of the functions learned by CNNs that provides rich insight into how they encode information.
We apply our framework on a colour variant of the model from \citet{lindsey2019unified} and demonstrate that, following the introduction of a bottleneck, different cell types tend to be organised according to their depth in the network, with no such organisation found in networks without a bottleneck. We detail the relationship between data, architecture, and learned representation through a series of control experiments.
% \todo{update this and maybe change, not such a fan of the list}
% \begin{itemize}
%     \item By analysing networks with random weights, we show that all observed opponency is learned.
%     \item By analysing networks trained on greyscale images, we show that our cell classification approach accords with the approach taken by \citet{lindsey2019unified}.
%     \item By training networks on images that have been rotated in hue space, we show that the type of spectral opponency depends on both the statistics of the data and the input colour space.
%     \item By training networks on images in CIELAB colour space, we show that emergent opponency is not unique to models with RGB inputs.
%     \item By training networks on the ImageNet data set \citep{russakovsky2015imagenet}, we show that our finding is not limited to networks trained on CIFAR-10 \citep{Krizhevsky09learningmultiple}.
%     \item By training networks on shuffled mosaic images, we show that the emergence of spatial opponency depends on the spatial consistency of the input.
%     \item By training networks on images with shuffled colour channels, we show that spectral opponency depends on the spectral consistency of the input.
% \end{itemize}
In total, we have trained 2490 models over 9 different settings, all of which have been made publicly available, alongside code for all of our experiments, via PyTorch-Hub at \url{https://github.com/ecs-vlc/opponency}.

\section{The Physiology and Psychophysics of Early Colour Vision}

% \begin{figure}
%     \centering
%     \begin{subfigure}{0.25\linewidth}
%         \includegraphics[width=\textwidth]{zhao_drifting.pdf}
%         \caption{Drifting Gratings}
%     \end{subfigure}
%     \hspace{0.1\linewidth}
%     \begin{subfigure}{0.25\linewidth}
%         \includegraphics[width=\textwidth]{zhao_flashing.pdf}
%         \caption{Flashing Bar}
%     \end{subfigure}
%     \caption{Spatial tuning curves for cells in the Mouse Lateral Geniculate Nucleus (LGN) from \citet{zhao2013orientation}.}
%     \label{fig:zhao}
% \end{figure}

% \todo[inline]{Much more here, discuss different viewpoints on opponency etc.}

% Holmgren - measurements of physiological effects of light on the retina - discovery of current - electroretinogram 
% https://physoc.onlinelibrary.wiley.com/doi/abs/10.1113/jphysiol.1927.sp002410 - first ME measurements (film recordings)

Since the advent of Holmgren's electrophysiology experiments in 1866 which first showed the flow of electrical current in the retina, vision scientists have sought to understand the cellular mechanisms that allow us to see.
In a series of articles, \citet{AdrianMatthewsPt1,AdrianMatthewsPt2,AdrianMatthewsPt3} started to explore the electrical response of the retina to light.
% : the relationship between light on the eye and the signal transmitted by the optic nerve \citep{AdrianMatthewsPt1}, the dynamical processes of retinal excitation (e.g. responses to short flashes of light rather than long exposures) \citep{AdrianMatthewsPt2}, and the interactions between retinal neurons caused by spatial non-uniformity and dynamics \citep{AdrianMatthewsPt3}. These experiments were necessarily crude given the equipment available at the time
% % (measurements taken using photographic film recording a capiliary electrometer, connected via a valve amplifier to two points at either end of the optic-nerve bundle)
% , but did start to provide insight into the kinds of stimuli that activate the visual system.
In later experiments,
% micro-electrode recordings were used to
practitioners explored how single cells respond to different stimuli; for example, \citeauthor{doi:10.1152/ajplegacy.1938.121.2.400}'s early measurements of the response of single optic nerve fibres to illumination \citep{doi:10.1152/ajplegacy.1938.121.2.400}, \citeauthor{doi:10.1113/jphysiol.1953.sp004829}'s `fly detectors' \citep{doi:10.1113/jphysiol.1953.sp004829}, \citeauthor{Lettvin1959}'s `bug perceivers' \citep{Lettvin1959}, and \citeauthor{hubel1962receptive}'s classic experiments in understanding receptive field structure \citep{hubel1962receptive}.
As a consequence of these experiments, a number of different observations and subsequent classifications of the behavioural characteristics of single cells and cell populations have been made. Although many of these classifications have been disproved or disputed, a number have stood the test of time. In particular, there is now a good shared understanding of how cells in the early parts of the visual systems of a range of primates respond to different spatial and spectral stimuli. This understanding covers the main pathway from the retina, through the Lateral Geniculate Nucleus (LGN) and into early parts of the visual cortex (e.g. V1 and V2).
In the following subsections we highlight the key findings from previous physiological studies that directly relate to the work presented in this paper.
% In particular, we draw the readers attention to the different classifications of cells based on observations regarding their response to specific classes of stimuli.

\subsection{Spatial opponency in cells}
Following \citet{AdrianMatthewsPt3}, \citet{doi:10.1152/ajplegacy.1938.121.2.400, Hartline:40} discovered evidence for different types of cellular behaviour to stimuli, and in particular found that inhibitory interactions were sometimes revealed when multiple receptors were excited~\citep{Hartline1952ThePO}. \citet{kuffler1953discharge} and \citet{doi:10.1113/jphysiol.1953.sp004829} investigated this finding further, and discovered cells with spatial receptive fields that are opponent to each other.
% These cells were characterised such that, in some spatial area, they were excited above the background rate by certain stimuli, and in other areas they are inhibited by certain stimuli.
These early results, obtained by presenting spots of light to different parts of the receptive field, showed an antagonism (opponency) between an inner centre and outer surround. Nowadays, it is widely accepted that such `centre-surround' cells can be found in the retina and LGN \citep{hubel2004brain}.
In contrast, the majority of cells in V1 are orientation tuned \citep{livingstone1984anatomy}.
One approach to analysing this spatial selectivity involves the presentation of drifting high contrast sinusoidal gratings \citep{levick1982analysis,de1982spatial,lennie1990chromatic,johnson2001spatial,johnson2008orientation,zhao2013orientation}.
For example, one can characterise orientation selectivity through presentation of gratings with fixed frequency and contrast at a range of orientations \citep{levick1982analysis,lennie1990chromatic,johnson2008orientation,zhao2013orientation}.
Similarly, a spatial frequency tuning curve can be obtained through the use of a fixed orientation and contrast \citep{de1982spatial,johnson2001spatial}.
These analyses again grant a notion of spatial antagonism (spatial opponency herein) in the cortex, where there exists a grating configuration that excites the cell and an opponent grating configuration which inhibits the cell \citep{SHAPLEY2011701}. Note that, although non-typical, presentation of grating stimuli have also been used to detect centre-surround organisation in the retina \citep[e.g.]{bilotta1989spatial} since these are cells which are highly tuned to frequency but not orientation selective.

\subsection{Colour vision and colour opponency}
With respect to colour vision, the first major physiological finding relates to the discovery of two broad classes of cell that respond to colour: those that exhibit opponent spectral sensitivity, and those (non-opponent) cells that do not. Experiments by \citet{DeValois:66} discovered `spectrally opponent' cells in the LGN of a trichromatic primate which are excited by particular single-wavelength stimuli and inhibited by others.
% More specifically, for a cell to be considered to be inhibited, its response must fall below its `background rate' to an empty stimulus. For excitation to occur, the response must be at some level above the background rate.
Additionally, \citet{DeValois:66} discovered that broadly speaking
% (see \citet{SHAPLEY2011701} and \citet{Shevell:17} for an in-depth exposition of how colour opponency is currently understood)
the cells could be grouped into those that were excited by red and inhibited by green (and vice-versa), and cells that were excited by blue and inhibited by yellow (and vice-versa). 
Indeed, these cells would appear to align with \citeauthor{hering1920grundzuge}'s unique hues (red, green, blue, and yellow) \citep{hering1920grundzuge}, which are unique in the sense that none of them can be viewed as a combination of the others.
However, the experiments from \citet{derrington1984chromatic} reveal that the cardinal axes of the chromatic response in the macaque LGN are not aligned to \citeauthor{hering1920grundzuge}'s unique hues but to cone responses.
The consequence of this finding is that spectrally opponent cells in early primate vision are best described as `cone opponent'.
It has similarly been argued that so called red / green opponency is better described as magenta / cyan and that these should be viewed as complementary colours, rather than opponent \citep{pridmore2005theory,pridmore2011complementary}.
For a more in-depth exposition of the contention between the physiological and psychophysical understanding of spectral opponency see \citet{Shevell:17}.
Cells that are `spectrally non-opponent' have also been observed in primate LGN; these are cells which are not sensitive to specific wavelengths but respond to broad range of wavelengths in the same way (either inhibitory or excitatory) \citep{de1958electrical,jacobs1964single}.
In V1, it has been suggested that cells described as selective to orientation but not colour by \citet{livingstone1984anatomy} are in fact colour opponent but with unbalanced cone inputs such that they respond to general changes in luminance \citep{lennie1990chromatic,johnson2001spatial}.

More recently, techniques such as functional Magnetic Resonance Imaging (fMRI) have been used to explore population coding of vision and colour related processes~\citep[e.g.][]{Engel199768,10.1093/cercor/bhv021,BOYNTON2002R838, Wade2008fMRIMO}. 
In particular, studies have shown strong responses in V1 to stimuli that are preferred by spectrally opponent cells \citep{kleinschmidt1996functional,Engel199768,schluppeck2002color}.
% \citet{}'s work is of particular relevance, demonstrating strong responses in the human V1 and V2 to red / green and blue / yellow stimuli.
The work of \citet{Wade2008fMRIMO} validates that the early visual system of the macaque (where many of the single-cell measurements of colour vision have been taken) correlates strongly with humans in terms of overall population responses to chromatic contrast; this is important to our work since we seek functional archetypes that are of general efficacy in visual intelligence.
% for this paper because we make comparisons of artificial models to experimental observations from a range of experiments with different primates.
% \todo{say something a bit different here. we don't compare networks to primates as such, but we do care about definitions from primates, good that these cell types are ubiquitous in nature}
It is, however, worth noting that \citeauthor{Wade2008fMRIMO} also show that in later areas of the visual pathway the topographical organisation of the macaque is fundamentally different.

Following \citeauthor{DeValois:66}'s initial findings, there has been a realisation that cells responsive to colour could be further grouped into `single opponent' and `double opponent' cells.
% The definition of these groups has shifted over the years, and the most modern interpretation is that single opponent cells do not have any structured spatial opponency, whereas double opponent cells do.
The defining characteristic of double opponent cells is that they respond strongly to colour patterns but are non-responsive or weakly responsive to full-field colour stimuli (e.g. solid colour across the receptive field, slow gradients or low frequency changes in colour)~\citep{SHAPLEY2011701}.
In the retina, double opponency presents as spectrally opponent cells with centre-surround organisation \citep{troy2002receptive}. In the primary visual cortex, there are both the spectrally opponent cells with oriented receptive fields mentioned above and non-oriented double opponent cells in the cytochrome oxidase rich blobs \citep{livingstone1984anatomy}.
% Conversely, single opponent cells respond to large areas of colour, and to the interiors of colour patches.
Note that one interpretation is that double opponent cells are both spatially and spectrally opponent.

\subsection{Linearity of retinal ganglion cell response}
There is a connection between anatomy and the relative presence of linear and non-linear cells in the retina. For example, midget cells, which are well approximated by a linear model \citep{smith1992responses}, are the most prevalent ganglion cell type in the human retina \citep{dacey1993mosaic}. In contrast, the most prevalent ganglion cell type in the mouse retina is a non-linear feature detector that is thought to act as an overhead predator detection mechanism \citep{zhang2012most}, not dissimilar to the aforementioned `fly detectors' and `bug perceivers'.
In their experiments with CNNs, \citet{lindsey2019unified} suggest that the contrast between the anatomy of the primate and mouse visual systems can be considered in terms of network depth. The authors subsequently present evidence that the natural differences in function derive from these associated differences in visual system anatomy. In particular, deeper networks learn linear features in early layers, whereas shallower networks learn non-linear features.

\section{Opponency in Artificial Vision}

The notion of a spatially opponent receptive field has a long history in computer vision. Notably, the Marr-Hildreth algorithm for edge detection \citep{marr1980theory} performs a Laplacian of Gaussian (often approximated by a Difference of Gaussian (DoG)) which resembles the function performed by centre-surround ganglion cells in the retina. Oriented-edge receptive fields were also modelled in early approaches to visual recognition. In particular, edge orientation histograms \citep{mcconnell1986method,freeman1995orientation} and later histograms of oriented gradients \citep{dalal2005histograms} are similar in principle to a layer of neurons with oriented-edge receptive fields with different rotation, frequency, and phase. DoG and edge orientation assignment are also integral components of the well-known Scale Invariant Feature Transform (SIFT) descriptor \citep{lowe1999object}.

In addition to approaches which directly model opponent receptive fields, several studies have shown emergent opponency in learning machines. For example, \citet{lehky1988network} found evidence for orientation selectivity in a neural network trained with back-propagation to determine the curvature of simple surfaces in procedurally generated images.
\citet{olshausen1996emergence} demonstrated the emergence of basis functions which resemble oriented receptive fields when learning an efficient sparse linear code for a set of images.
Similar results are presented by \citet{BELL19973327} who show that a nonlinear `infomax' network which performs Independent Component Analysis (ICA), trained on images of natural scenes, produces sets of visual filters that show orientation and spatial selectivity.
% Sparse coding and the ICA algorithm are based on the efficient coding hypothesis \citep{barlow1961possible} which suggests that the goal of early vision is to reduce the redundancy of the input, and hence encourage emergent structure. The downside of these early approaches to efficient coding is that they express images in only a single layer, and hence are unable to learn higher order features.
% 
\citet{lehky1999seeing} use a four layer neural network to map cone responses to a population of Gaussian tuning curves in CIE colour space and demonstrate colour opponent neurons in the hidden layers. \citet{karklin2003learning} propose a hierarchical probabilistic approach to learning a non-linear efficient code. The authors demonstrate the emergence of higher order features such as object location, scale, and texture.
Alternatively, \citet{NIPS2006_3018} introduced Recursive ICA, where the outputs of a previous application of ICA are transformed such that it may be re-applied. The authors again demonstrate the emergence of these higher order features when applying their model to natural images.
\citet{Wang2015ModelingTO} use Recursive ICA to automatically learn visual features that accord with those found in the early visual cortex. The authors subsequently model the object recognition pathway using Gnostic Fields \citep{kanan2013recognizing, kanan2014fine}, a brain-inspired model of object categorisation.
\citet{Wang2015ModelingTO} demonstrate that the features in the first ICA layer, trained on natural images,
are oriented-edges with the colour opponent characteristics typical of V1 neurons (dark-light, yellow-blue, red-green).
The second layer filters are sensitive to edges of different frequency and orientation, reminiscent of complex cells in V1. Cells which exhibit responses similar to simple and complex neurons in V1 can only be observed in the two ICA layers.

In this work we are primarily concerned with opponency in deep CNNs,
% , which are similar to ICA in the sense that learned filters correspond to features in the input. In fact,
for which some early approaches used variants of ICA to learn the filters \citep{le2011ica}.
% The first key difference between modern CNNs and ICA is that
Modern CNNs are trained using the back-propagation algorithm, similar to the work of \citet{lehky1988network,lehky1999seeing}, such that the features learned are dependent on the objective function of the model.
% , rather than merely the statistics of the input data.
% The second key difference is that
In addition, CNNs are typically constructed with many more layers of non-linear feature extraction than the one or two layers used in ICA. As a result, CNNs permit a notion of functional organisation: `what happens where' rather than just `what happens'. Due to the connections between CNNs and ICA, one might reasonably expect CNNs to exhibit emergent opponency. This is indeed the case, with multiple works pointing out that learned filters in early layers appear to be spatially and colour selective \citep{krizhevsky2012imagenet,zeiler2014visualizing,lindsey2019unified,rafegas2018color,olah2020overview}.
% In contrast to the typical qualitative assessment,

\citet{rafegas2018color} propose an automated measurement of the spectral selectivity of convolutional neurons. For their approach, the authors find image patches which maximally excite each neuron and construct an index with high values when these patches are consistent in colour.
% and low values otherwise. This provides a measure of the extent to which each neuron is selective to colour.
The authors further
% classify a cell as `double colour' if it is excited by two distinct spectral groups. Finally, the authors
suggest that a neuron is double opponent if it is selective to two distinct colours that are roughly opposite in hue. Note that these definitions of opponency are not direct correlates of the previously discussed definition. The key difference is that the electrophysiological definition requires an understanding of the stimuli which inhibit cells in addition to the stimuli which excite them. This is important since although cells which are excited by two colours may be projecting the input on an opponent axis, they may also just be activating for both colours indiscernibly.
% Put simply, it may not be possible to distinguish between the two colours using only the output of a neuron classified as opponent by the definition from \citet{rafegas2018color}.
The double colour selective neurons found by \citet{rafegas2018color} are typically red-cyan, blue-yellow and magenta-green. These do not closely reflect the opponent axes of the primate LGN. This is to be expected since cone opponency observed in nature translates to channel opponency in a convolutional model, and so we can reasonably expect the opponent axes to be aligned with extreme RGB values rather than cone responses or \citeauthor{hering1920grundzuge}'s unique hues (although note that these are a subset of the RGB extrema).
% As well as being concerned with a more specific definition of opponency, we have a different aim to the works discussed here.
% In particular, we are not focused on constructing a deep, nuanced understanding of a single architecture or trained model (as is the case with \citet{rafegas2018color}). Instead, our goal is to use the uncontroversial idea that opponent cells are powerful functional units as a way to
% % construct an informative characterisation of the type of function that is performed by a trained CNN. Equipped with this characterisation, we can then seek an understanding of
% explore the connection between architecture, data, and learned representation.

\section{Methods}\label{sec:methods}

In this section we detail our methodology for classifying convolutional cells according to their spatial and colour processing.
Generalising the discussed physiological definitions, to classify a cell as opponent we require: a set of stimuli, the ability to measure the response of the cell to each stimulus, and a measurement of the baseline response of the cell (in order to establish excitation and inhibition). The response of each neuron to the input is readily available in a deep network, and we define the baseline as the response of the cell to a black input (a matrix of zeros). If there exists a stimulus for which the cell is excited (responds above the baseline) and a stimulus for which the cell is inhibited (responds below the baseline), then the cell is opponent to the axis of variance of the stimuli set. We first describe the two stimuli sets that we will use for the classification of spatial and colour opponency. We go on to discuss automatic classification of a cell as double opponent and how we can infer the specific `type' of an opponent cell.
In addition, we introduce an approach for studying the hue sensitivity curve of a deep network, inspired by \citet{bedford1958wavelength}
% , which bridges the gap between our spectral analysis and the known characteristic sensitivity of human observers.
The experiments laid out in this section will form our core results. We will later perform a control study to determine how well these results extend to different settings.
% \todo{This could be better, doesn't flow very well with everything up to now}
% , for a depiction of the characterisation of a single cell see Appendix~\ref{app:single}.
% \todo{Following the definitions given by XXX, to ...}

\paragraph{Spatial opponency}

\begin{figure}
    \centering
    \begin{subfigure}{0.15\linewidth}
        \includegraphics[width=\linewidth]{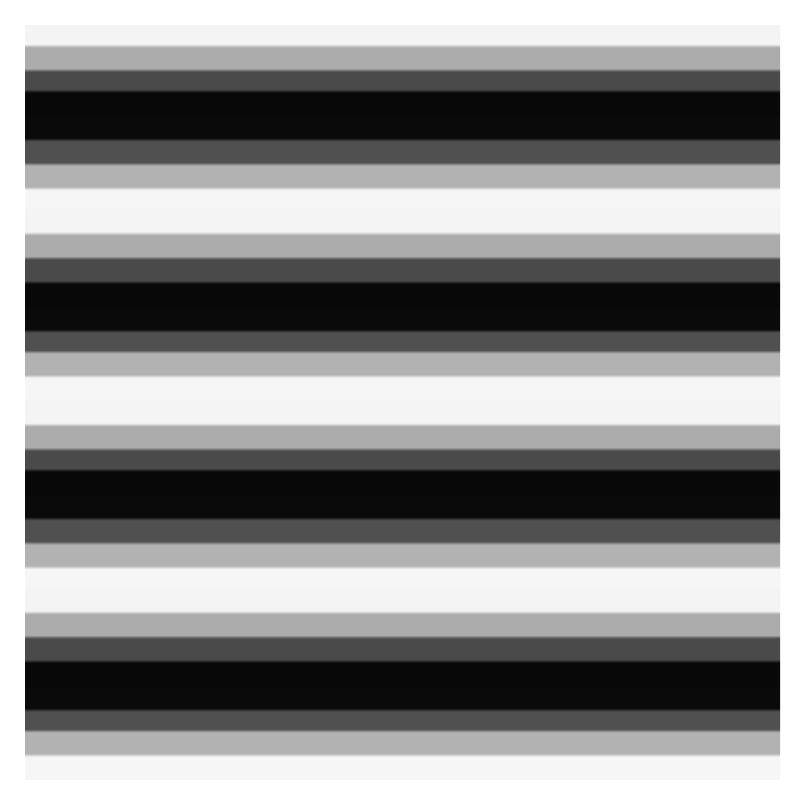}
        \caption{$\theta=\ang{0}$}
    \end{subfigure}
    \hspace{0.05\linewidth}
    \begin{subfigure}{0.15\linewidth}
        \includegraphics[width=\linewidth]{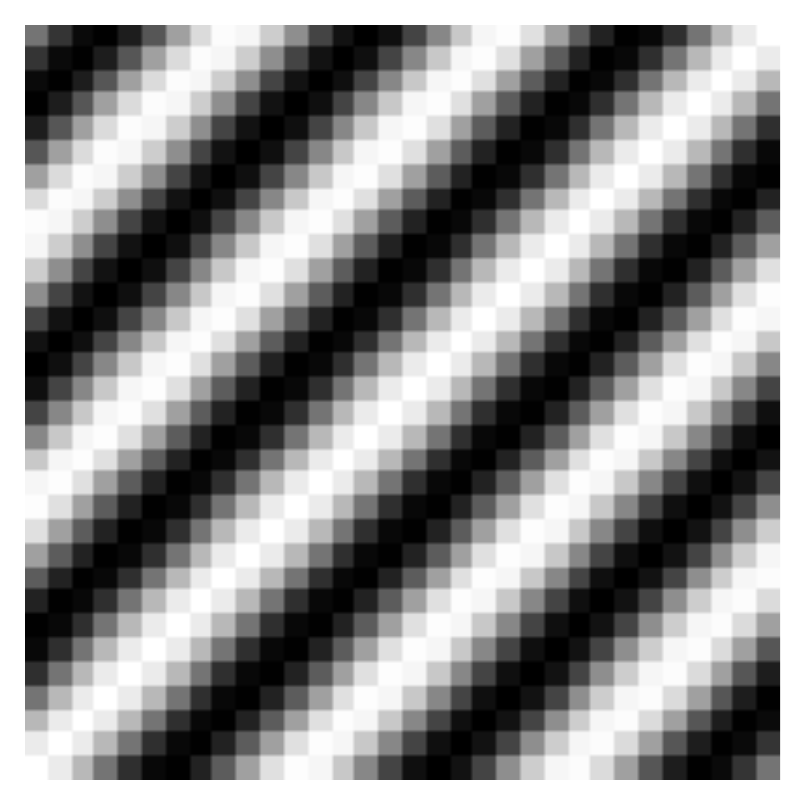}
        \caption{$\theta=\ang{45}$}
    \end{subfigure}
    \hspace{0.05\linewidth}
    \begin{subfigure}{0.15\linewidth}
        \includegraphics[width=\linewidth]{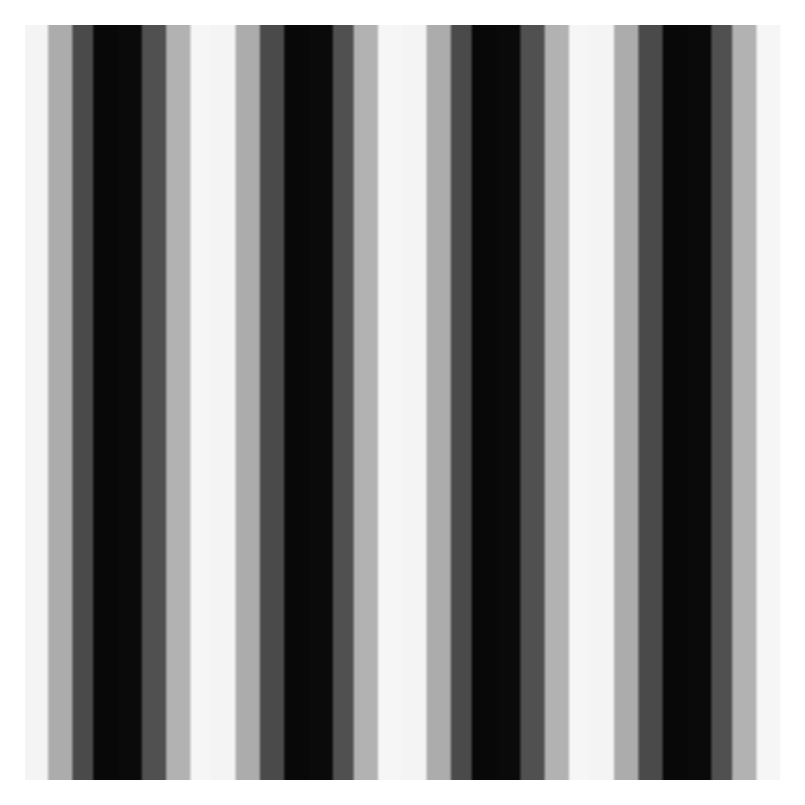}
        \caption{$\theta=\ang{90}$}
    \end{subfigure}
    \hspace{0.05\linewidth}
    \begin{subfigure}{0.15\linewidth}
        \includegraphics[width=\linewidth]{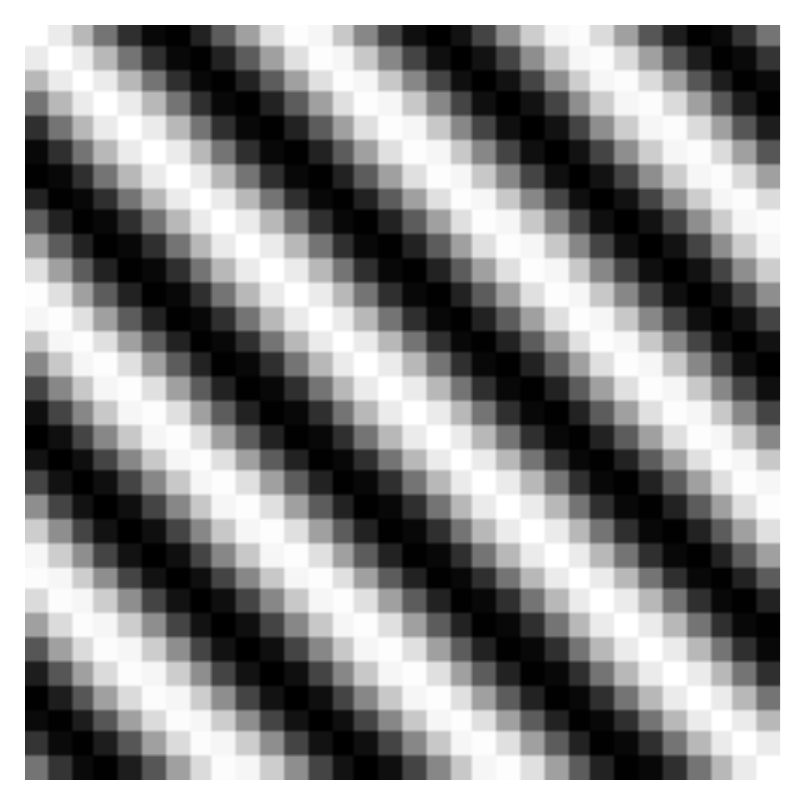}
        \caption{$\theta=\ang{135}$}
    \end{subfigure}
    \caption{Examples of grating patterns used as stimuli for the spatial opponency experiments. These samples have been generated using PsychoPy \citep{peirce2019psychopy2}, with different angles ($\theta$), frequency of $4$, and phase of $0$.}
    \label{fig:greygratings}
\end{figure}

% \begin{figure}
%     \centering
%     \includegraphics[width=0.9\linewidth]{spatial.pdf}
%     \caption{Our approach for obtaining a spatial tuning curve for cells in a deep network.}
%     \label{fig:spatial}
% \end{figure}

To classify spatial opponency, we require a set of stimuli that vary spatially. Following \citet{johnson2001spatial}, we construct a set of high contrast greyscale gratings produced from a sinusoidal function for a range of rotations, frequencies, and phases. Figure~\ref{fig:greygratings} gives some example stimuli generated using PsychoPy \citep{peirce2019psychopy2} with various degrees of rotation. We compute the response of each cell to each stimulus and compare these to the baseline to obtain a tuning curve which can be used to perform an automatic classification.
% Figure~\ref{fig:spatial} depicts our process for obtaining the tuning curve of a cell.
In addition to spatially opponent, we have spatially non-opponent and spatially unresponsive. A non-opponent cell is one which may be excited or inhibited by the stimuli but does not cross the baseline. An unresponsive cell is one which is neither excited nor inhibited by any of the stimuli. We do not use any form of tolerance when making the above classifications. The reason for this is that each cell will activate in its own space, and so the relative effect of a fixed tolerance could vary greatly between cells. As a consequence of this design decision, it is likely that the output of any unresponsive cells lies in a clipped region of the activation function.
For example, if using Rectified Linear Units (ReLUs), the cell response may remain at $0$ for all of the stimuli. Such cells are either highly tuned to a particular, complex stimulus or merely unresponsive to all stimuli.
That said, our interests here are primarily bound to the existence and distribution of opponent and non-opponent cells only; we are not aware of any demonstration that unresponsive cells are found in nature.
Recall that although the described stimuli are oriented edges, they can still be used to infer centre-surround spatial opponency since a cell with a characteristic centre-surround receptive field would be highly tuned to a particular frequency and phase, but responsive to a broad range of angles.

% Opponency

% To explore spatial opponency, we can use a similar set-up to our experiments with spectral opponency,
% % rephrase this after order change
% measuring cell response to a series of high contrast greyscale gratings produced from a sinusoidal function for a range of rotations, frequencies and phases following \citet{johnson2001spatial} (see Appendix~\ref{app:gratings} for an example). 
% % Put the example here
% We can subsequently classify a cell as spatially opponent, non-opponent or unresponsive by comparing the maximum and minimum responses against the baseline in the same way as before.
% Further, we can characterise whether a cell is orientation tuned by isolating the grating frequency which gives the largest response for all orientations and phases, then computing the average response per orientation for that frequency across all phases. % do we actually do this anywhere??

% More here:
% - diagram from poster
% - pseudo-code for grating generation???
% - range of phases, angles and frequencies considered

% Orientation tuning?

% Anything else?

\paragraph{Colour opponency}

% \begin{figure}
%     \centering
%     \includegraphics[width=0.9\linewidth]{spectral.pdf}
%     \caption{Our approach for obtaining a spectral tuning curve for cells in a deep network.}
%     \label{fig:spectral}
% \end{figure}

To classify spectral opponency, \citet{DeValois:66} vary the stimuli according to wavelength. For our experiments we propose using stimuli which vary in hue, rather than wavelength. The reason for this is that the trained networks will expect an RGB input and there is no exact mapping from wavelength to RGB.
We could consider a more biologically valid colour representation such as the cone response space used by \citet{lehky1999seeing} but opt for RGB as it is the standard practice in deep learning. We sample colours in the Hue, Saturation, Lightness (HSL) colour space for all integer hue values with saturation of $1.0$ and lightness of $0.5$. We then convert our stimuli to RGB before forwarding to the network and constructing the colour tuning curve.
We can perform classification by following the same process of comparing to the baseline as in the spatial setting. We use the terms hue opponency, and colour opponency interchangeably to refer to the different cell types found through this process.

\paragraph{Double opponency}

As discussed, we can automatically classify a cell as double opponent if it is both colour and spatially opponent. Our interests here lie in whether or not double opponent cells emerge in convolutional networks trained with a classification objective. Note that it has been observed that most spectrally opponent cells in macaque V1 are also orientation selective \citep{johnson2008orientation}, that is, they are double opponent. Unlike in the single opponent cases, we do not define a notion of double non-opponency or double unresponsiveness (although such classifications could be made if required).

\paragraph{Excitatory and inhibitory colours}

Using the colour tuning curve, we can further determine the hue which most excites or inhibits each cell. Since cells are typically equipped with a non-linear activation function, there may be a wide range of stimuli for which they produce the lowest response. As such, we use the pre-activation output to infer the most inhibitory stimulus.
This excitation and inhibition data will allow us to plot the distribution of colours to which cells in networks are tuned.
Note that this distribution is insufficient to describe the type of opponency since it does not permit an understanding of whether there are distinct classes of opponent cell. For example, the distribution of excitation and inhibition does not distinguish between two groups of cells that are red / green opponent and blue / yellow opponent respectively, or many groups of cells that are red / green opponent, green / blue opponent, blue / red opponent etc.
One option would be to applying a clustering technique to the most excitatory and inhibitory responses. However, this would introduce additional challenges through the need for appropriate algorithm and hyper-parameter choice.
Instead, we can additionally study the conditional distribution of maximal excitation, given maximal inhibition by some colours in a chosen range. We suggest evaluation of these conditional distributions for the following hue ranges: red ($[315\si{\degree},45\si{\degree})$), yellow ($[45\si{\degree},75\si{\degree})$), green ($[75\si{\degree},165\si{\degree})$), cyan ($[165\si{\degree},195\si{\degree})$), blue ($[195\si{\degree},285\si{\degree})$), and magenta ($[285\si{\degree},315\si{\degree})$).
By enabling direct assessment of the inhibition / excitation pairs, this will give a much deeper understanding of the kinds of opponency present in the networks being analysed.
% \todo{Saying ReLU a lot, haven't defined the model yet, more generality needed}
% \todo{Make this sound more like the plot doesn't exist yet...}

% Maybe mention complementary colours here???

\paragraph{Hue Sensitivity}

% \begin{figure}
%     \centering
%     \includegraphics[width=\linewidth]{wavelength.pdf}
%     \caption{Plot of the Red, Green and Blue components of the wavelength to RGB conversion. This function is piece-wise differentiable.}
%     \label{fig:wavelength}
% \end{figure}

In addition to the hue tuning curve, we can consider the hue sensitivity of a network. Specifically, we look to replicate the experiments of \citet{bedford1958wavelength}, who showed that the change needed to elicit a just-noticeable difference in hue to a human observer is a complex function of wavelength. In \citet{long2006spectral} the authors further suggest that the reason for this non-uniform spectral sensitivity derives from the statistics of natural scenes, showing that the curve predicted from a data set of natural images bares a strong resemblance to that obtained for a human observer.
Another way to explain the discrimination curve is in terms of cone responses \citep{zhaoping2011human}. This is more direct since scene statistics can be seen as indirectly controlling wavelength discrimination through evolutionary modifications of cone properties.
It is expected that such a sensitivity curve, though over hue rather than wavelength, will enable a more holistic view of colour tuning.

% \todo{Why do we do this? Because it gives another, more holistic, view on how our networks encode colour}
To perform a similar experiment to \citet{bedford1958wavelength}, note that the just-noticeable difference method is inversely related to the gradient of the perceived colour with respect to wavelength, which can be seen as a form of sensitivity.
% To see this, consider that we have a fixed change in perception (that is, the change required to be noticed) and a variable change in wavelength, giving a finite difference measure that is inversely proportional to the gradient of the perceived colour with respect to the wavelength.
By virtue of automatic differentiation, it is trivial to obtain the gradient of the activation in a layer of our network with respect to the RGB input.
Since the conversion from HSL to RGB is piece-wise differentiable, we can further obtain the approximate gradient of the activation with respect to hue.
Note that we use the hidden layer activation of a network rather than a notion of `perceived' colour, so it is unclear whether these results should reflect the biological data. Furthermore, in light of the above, one might expect that the predominant features of the sensitivity curve should derive from the relative responses of the RGB channels as a function of hue.
% We use wavelength here simply so that we can make a direct comparison between the spectral `sensitivity' of our network, that of a human observer, and the function predicted from the statistics of natural scenes by \citet{long2006spectral}. It would also be entirely possible to construct this curve as a function of hue rather than wavelength. We suggest that for most applications hue would be desirable in light of the crude nature of the mapping from wavelength to RGB.

% Opponency

% To classify cells according to their spectral opponency, we can simulate the experimental procedure of \citet{DeValois:66}. Specifically, we first present the network with uniform coloured images and measure the response of the target cell. By sampling colour patches according to hue we can show the network a range of stimuli and construct a response curve. We then classify each cell as either `spectrally opponent' or `spectrally non-opponent' by considering this curve relative to a background rate, defined as the response of the cell to a zero image. A spectrally non-opponent cell is one for which all responses are either above or below the baseline. A spectrally opponent cell is one for which the response is above the baseline for some colours and below the baseline for others. We further define an additional class, spectrally unresponsive, for cells which respond the same regardless of the hue of the input.

% More here:
% - diagram from poster
% - additional details, do we use any tolerance? why hue instead of wavelength?

% Most excitatory / inhibitory colours

% Colour sensitivity?

\section{Results}\label{sec:results}

We now present the results for our core experiments with Retina-Net models trained on colour CIFAR-10. We will later perform a control study and provide an in-depth discussion of the implications of these results, our aim in this section is merely to present the core findings of this work.

\subsection{Retina-Net}

\begin{figure}
    \centering
    \begin{subfigure}{\linewidth}
        \includegraphics[width=\textwidth]{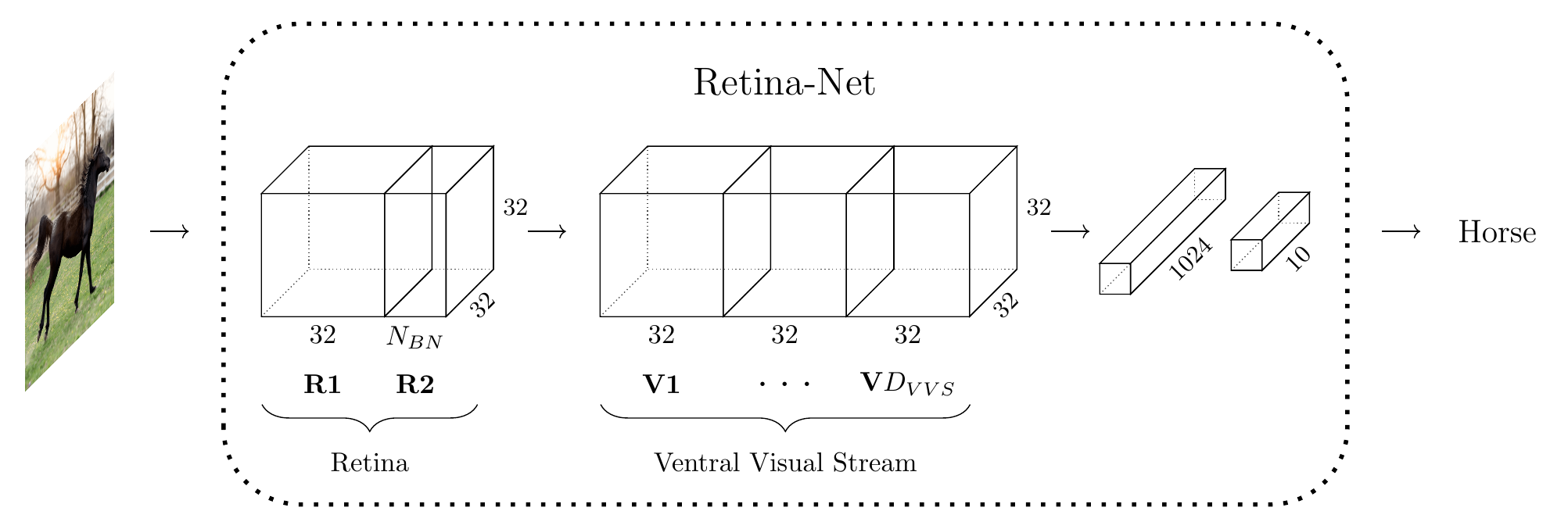}
        \caption{Retina-Net schematic}
        \label{fig:retinanet}
    \end{subfigure}
    \begin{subfigure}{0.45\linewidth}
        \includegraphics[width=\linewidth]{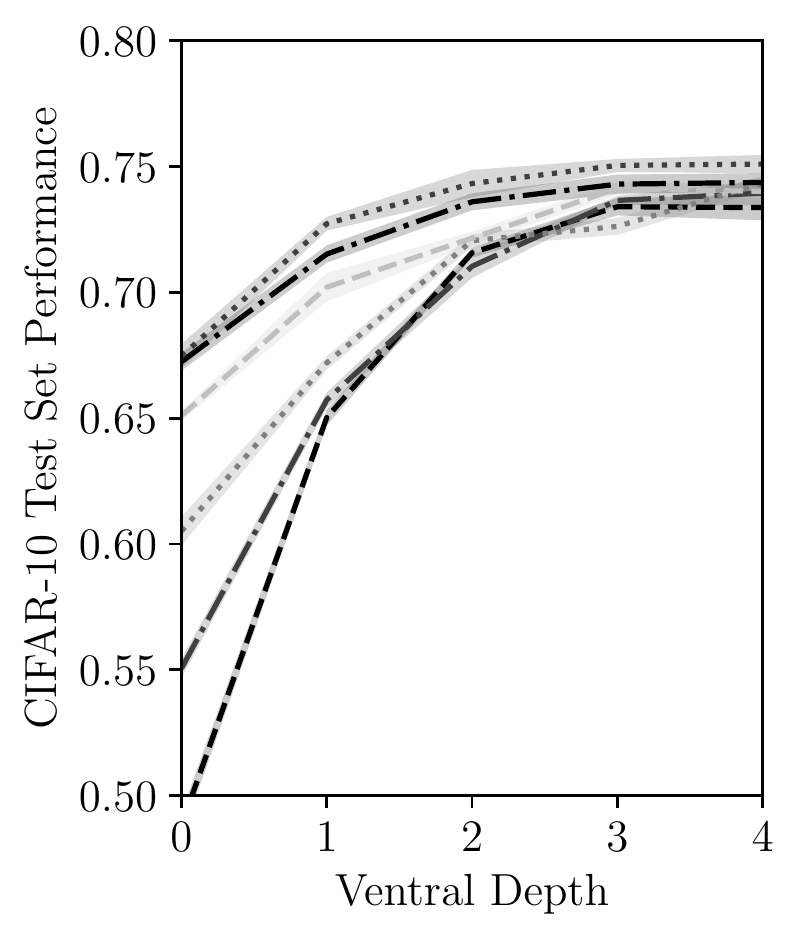}
        \caption{Greyscale}\label{fig:model-acc:grey}
    \end{subfigure}
    \hspace{0.05\linewidth}
    \begin{subfigure}{0.45\linewidth}
        \includegraphics[width=\linewidth]{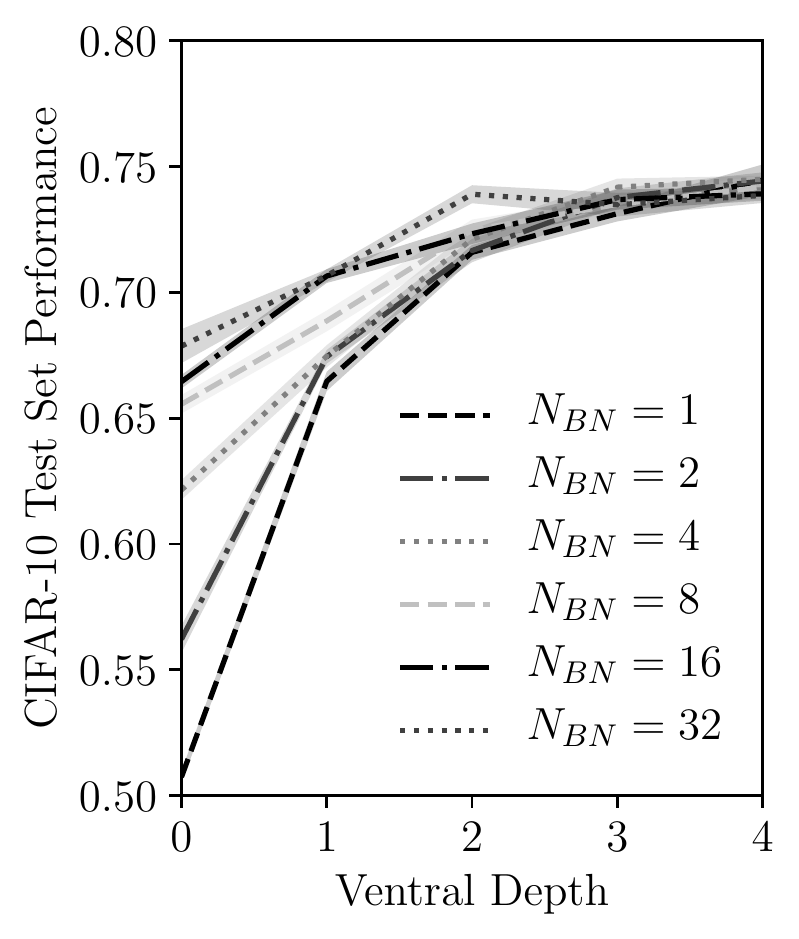}
        \caption{Colour}\label{fig:model-acc:colour}
    \end{subfigure}
    \caption{(\protect\subref{fig:retinanet}) Schematic of the Retina-Net model from \citep{lindsey2019unified}. (\protect\subref{fig:model-acc:grey}, \protect\subref{fig:model-acc:colour}) CIFAR-10 test accuracy for the different combinations of retinal bottleneck and ventral depth explored in the experiments. Mean and standard error given over $10$ trials.}
    % \label{fig:retinanet}
\end{figure}

% \begin{figure}
%     \centering
%     \begin{subfigure}{0.45\linewidth}
%         \includegraphics[width=\linewidth]{greyscale_accuracy.pdf}
%         \caption{Greyscale}
%     \end{subfigure}
%     \hspace{0.05\linewidth}
%     \begin{subfigure}{0.45\linewidth}
%         \includegraphics[width=\linewidth]{accuracy.pdf}
%         \caption{Colour}
%     \end{subfigure}
%     \caption{CIFAR-10 test accuracy for the different combinations of retinal bottleneck  and ventral stream depth explored in the experiments. Mean and standard error given over $10$ trials.}
%     \label{fig:model-acc}
% \end{figure}

Since we are interested in understanding the casual link between architectural constraints and learned representation, we adopt the same deep convolutional model of the visual system as \citet{lindsey2019unified}, referred to as Retina-Net.
This model, depicted in Figure~\ref{fig:retinanet}, consists of a model of the retina which feeds into a model of the visual cortex and Ventral Visual Stream (VVS).
% \todo{Make the figure central, it's annoyingly off-center}
The retina model consists of a pair of convolutional layers with ReLU nonlinearities.
% \todo{Cite? Define?}
The ventral network is a stack of convolutional layers (again with ReLUs) followed by a two layer MLP (with 1024 ReLU neurons in the hidden layer, and a 10-way softmax on the output layer).
Note that \citet{lindsey2019unified} additionally explore a model of the LGN, which can be considered as an extension of the retinal bottleneck \citep{ghodrati2017towards}. We do not include such an exploration in this work as we are primarily focused on opponency and colour tuning in the bottleneck layer.

As with \citeauthor{lindsey2019unified}'s work, the networks are trained to perform classification on the CIFAR-10 data set \citep{Krizhevsky09learningmultiple}, the only difference being that our model expects RGB inputs rather than greyscale. The choice of an object categorisation task is validated by previous studies which show there is a strong correlation between neural unit responses of CNNs trained on such a task and the neural activity observed in the primate visual stream \citep{yamins2014performance, gucclu2015deep, Cadena201764}. For further discussion of these results, please refer to \citet{lindsey2019unified}. Note that there may be many other learning tasks that are biologically valid in the sense that they yield similar functional properties. For example, self-supervised learning through deep information maximisation \citep{hjelm2018learning} and contrastive predictive coding \citep{henaff2019data} may present viable alternatives to the supervised object recognition used here.

We train models across the same range of hyperparameters as \citet{lindsey2019unified}. Specifically these are: bottleneck width $N_{BN} \in \{1, 2, 4, 8, 16, 32\}$, and ventral depth $D_{VVS} \in \{0, 1, 2, 3, 4\}$. Again following \citet{lindsey2019unified} we perform 10 repeats, with error bars denoting the standard deviation in result across all repeats. Networks were trained for $20$ epochs with the RMSProp optimizer and a learning rate of $1e-4$ with initial weights sampled via the Xavier method \citep{glorot2010understanding}. We note that in order to replicate the results from \citet{lindsey2019unified}, we required additional regularisation. Specifically, we use a weight decay of $1e-6$ and data augmentation (random translations of $10\%$ of the image width/height, and random horizontal flipping). Figures~\ref{fig:model-acc:grey} and \ref{fig:model-acc:colour} give the average terminal accuracy for models trained both on greyscale and colour images respectively. The greyscale accuracy curves match those given in \citet{lindsey2019unified}. The accuracy for networks trained on colour images is generally higher, particularly for networks with no ventral layers.
We will later discuss additional training settings that are variants of the above.
% In total we have trained over $1800$ models, all of which have been made publicly available through PyTorch-Hub at \url{https://github.com/ecs-vlc/opponency}.
% \todo{Maybe not have this twice, it's distracting...}
% - point to the model code in the repo?

\subsection{Characterising Single Cells}
\begin{figure}
    \centering
    \begin{subfigure}{\linewidth}
        \includegraphics[width=\linewidth]{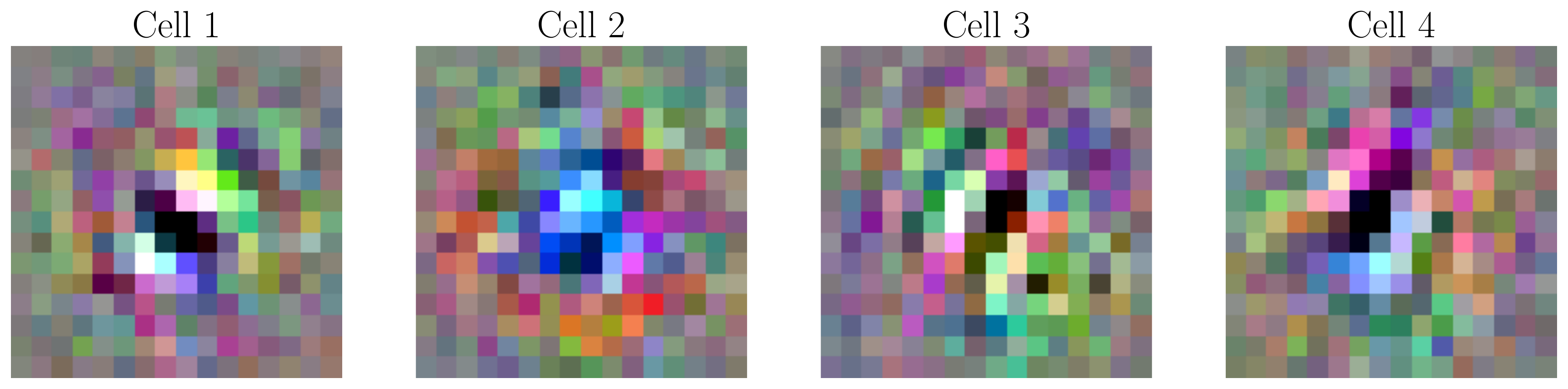}
        \caption{First order receptive field approximation}
        % \label{subfig:spect}
    \end{subfigure}
    \begin{subfigure}{\linewidth}
        \includegraphics[width=\linewidth]{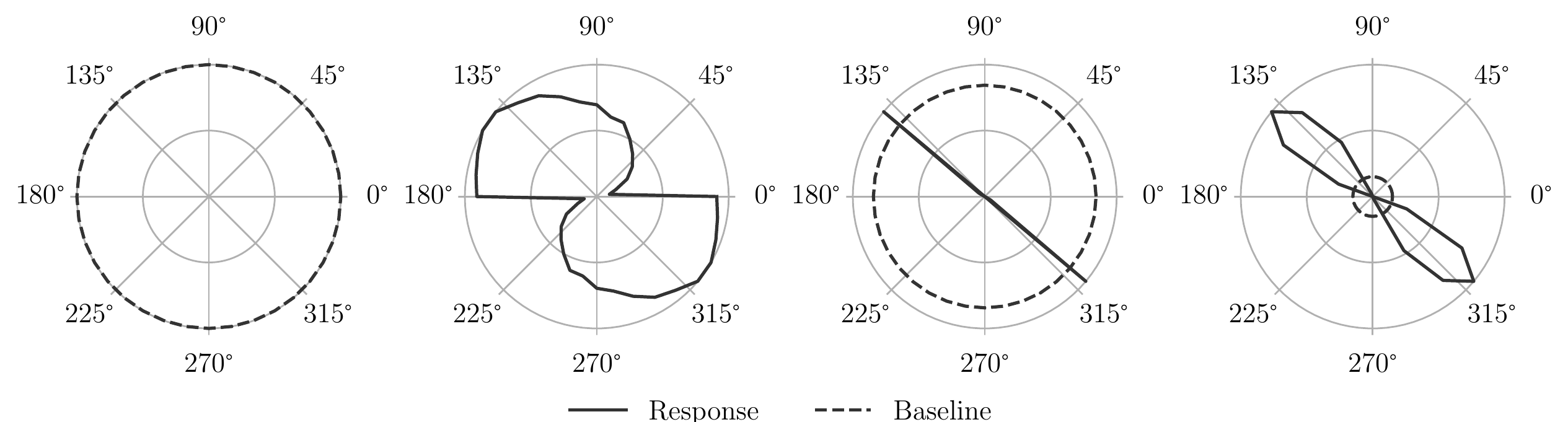}
        \caption{Orientation tuning curve for frequency and phase with minimal response}
        \label{subfig:spatial_min}
    \end{subfigure}
    \begin{subfigure}{\linewidth}
        \includegraphics[width=\linewidth]{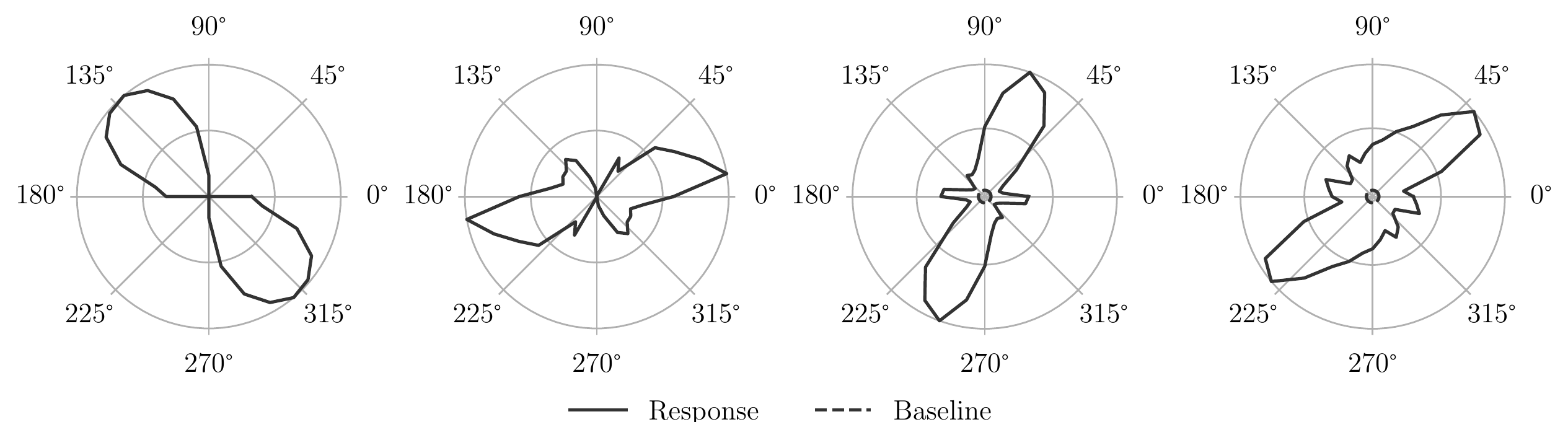}
        \caption{Orientation tuning curve for frequency and phase with maximal response}
        \label{subfig:spatial_max}
    \end{subfigure}
    \begin{subfigure}{\linewidth}
        \includegraphics[width=\linewidth, trim=0 0 0 1.5em, clip]{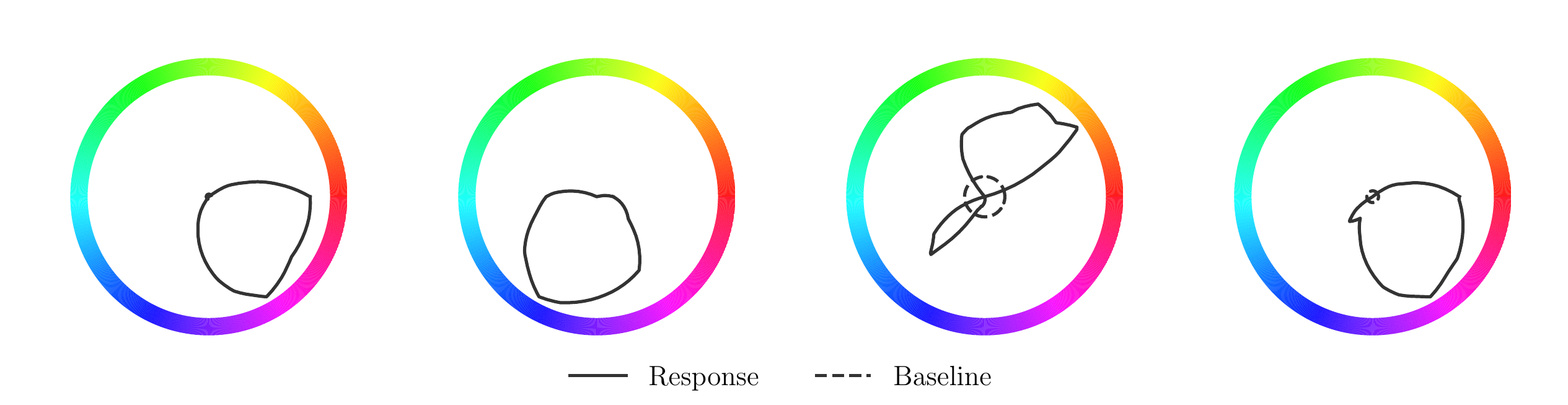}
        \caption{Colour tuning curve}
        \label{subfig:spect}
    \end{subfigure}
    \caption{Characterisation of the 4 cells in the second retinal layer of a network with $N_{BN} = 4$ and $D_{VVS} = 2$. (a) The receptive field approximation obtained from the gradient of the cell with respect to a blank image. (b) \& (c) Orientation tuning curves for the frequency and phase combination that yielded the smallest and largest response respectively. (d) Colour tuning curve over the hue wheel. Cells 1, 3 and 4 are double opponent, cell 2 is non-opponent.}
    \label{fig:cellcharac}
\end{figure}

To begin, we illustrate our framework for characterising single cells. Figure~\ref{fig:cellcharac} shows the first order receptive field approximations, orientation tuning curves, and colour tuning curves for four cells in the bottleneck layer of a network with $N_{BN} = 4$ and $D_{VVS} = 2$.
Following \citet{lindsey2019unified}, the receptive field approximation is the gradient (obtained through back-propagation) of the output of a single convolutional filter in a single spatial position (that is, a single convolutional `neuron') with respect to a blank input with a constant value of $0.01$. This small positive amount is required to ensure that each of the cells is in the linear region of the ReLU activation function (that is, the gradient is non-zero).
The gradient image is then normalised and scaled so that it can be interpreted visually.
Visually, cells 1, 3, and 4 appear to be greyscale edge filters, whereas cell 2 is red / blue or magenta / cyan centre-surround. However, the limitation of this analysis is the noise in the approximation. For example, one could argue that cell 1 is centre-surround with a dark centre and a magenta surround. Assessments given for any of the cells will be similarly contentious. Furthermore, this representation permits no understanding of inhibition. For example, cell 2 may be better described as tuned to blue hues in the interval $(\ang{180}, \ang{270})$, rather than centre-surround opponent.

To further characterise each cell, we employ our described approach. To characterise spatial opponency, in Figures \ref{subfig:spatial_min} and \ref{subfig:spatial_max} we provide orientation tuning curves for the frequency and phase which elicit the weakest and strongest responses respectively. If the cell responds above the baseline in one tuning curve and below in the other, or if either curve crosses the baseline, then the cell is spatially opponent. We can therefore say that, by our definition, cells 1, 3 and 4 are spatially opponent. In contrast, cell 2 is merely spatially non-opponent, always responding above the baseline for any choice of rotation, frequency, and phase. In addition to classifying opponency, we can identify the orientation tuning of each cell by further study of the curves in Figure \ref{subfig:spatial_max}. Figure \ref{subfig:spect} gives the colour tuning curves for each cell. As hue is the only parameter to consider, classification here is simpler; the cell is hue opponent if the tuning curve crosses the baseline. Given this definition, we can say that cells 1, 3 and 4 are hue opponent, although the extent of inhibition is different in each case. Furthermore, for every cell we can identify the range of hues to which it is tuned. 

Following interpretation of the tuning curves we can now state that cells 1, 3 and 4 are double opponent and that cell 2 is non-opponent both spatially and with regard to hue. Furthermore, for each cell we can state the orientation and hue to which it is tuned. For example, cell 2 is broadly excited by blue stimuli but with a distinct peak at a hue of around $\ang{240}$. Cell 2 is spatially tuned to lines oriented in the interval $(\ang{0}, \ang{45})$. Although it is true that this approach gives us a deeper understanding of each cell, the real value is in the fact that each of the above steps can trivially be automated over the whole cell population. We therefore transition away from studying single cells, and instead consider the distributions of different cell types for the remainder of the paper.

% all of the cells are spectrally opponent
% they may also be spatially opponent just not for this frequency / phase
% we additionally show the receptive field approximation, following lindsey this is the gradient of the cell with respect to a blank stimulus

\subsection{Characterising Cell Populations}

For each result in this section, we automate cell classification following our described method and present the distribution of each cell type as a function of retinal bottleneck width and ventral depth. This allows us to understand the effect that these two architectural variables have on the kinds of cells that are learned and where they are found in the network. Note, however, that cells in deeper layers are expected to have a highly non-linear response and thus may have receptive field properties that are quite different to the opponent cells observed in shallower layers. As such, observations regarding these deeper layers (`Ventral 2' in particular) should be considered only in the context of our approach and may not generally apply to the broader understanding of opponency.

\paragraph{Spatial opponency}

\begin{figure}
    \centering
    \includegraphics[width=\textwidth]{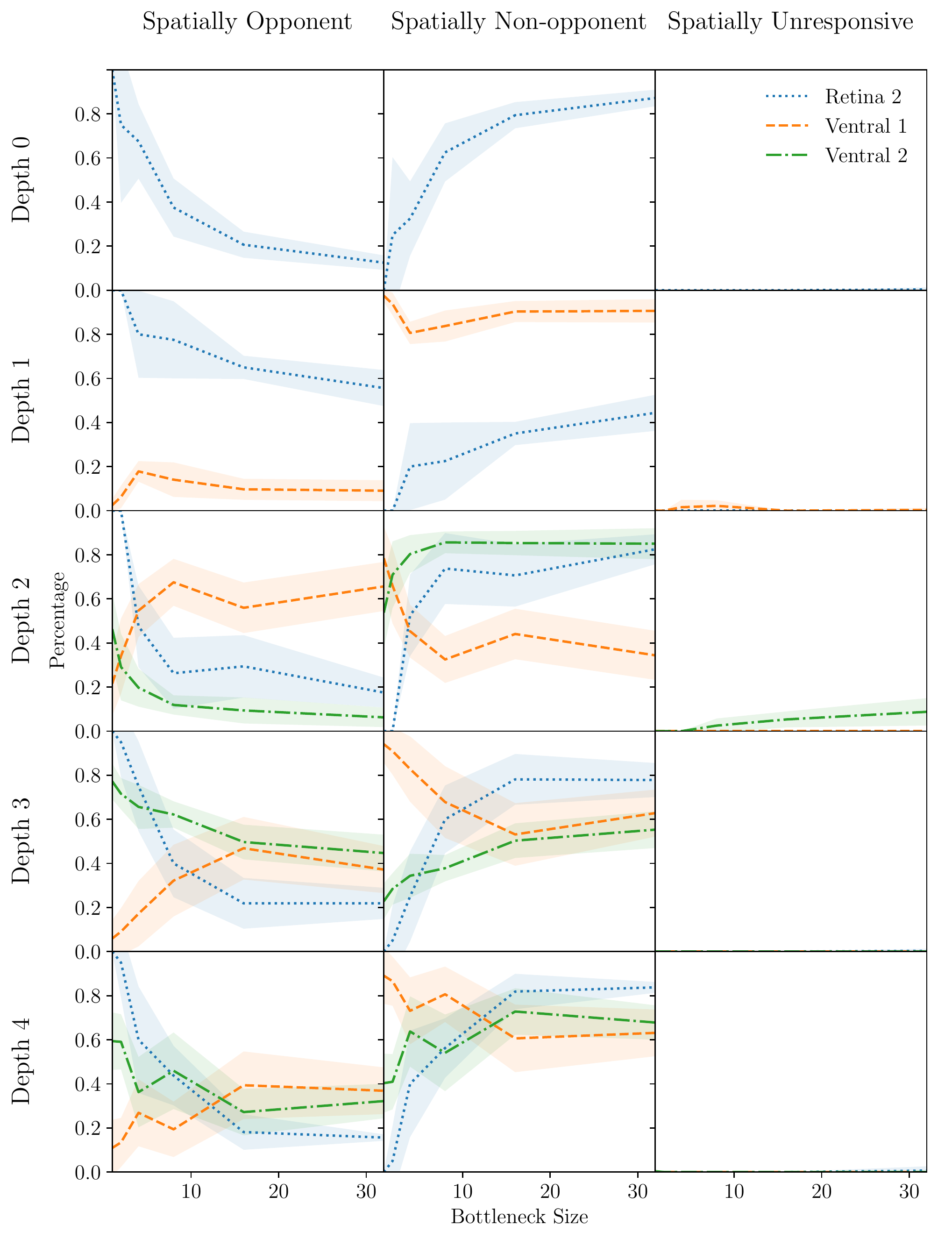}
    \caption{Distribution of spatially opponent, non-opponent, and unresponsive cells in different layers of our model as a function of bottleneck width, for a range of ventral depths. Functional organisation emerges for networks with tight bottlenecks. The last convolutional layer (`Retina 2' when depth is 0, `Ventral 1' when depth is 1 etc.) exhibits a reduction in spatial opponency. The penultimate convolutional layer (`Retina 2' when depth is 1, `Ventral 1' when depth is 2 etc.) exhibits an increase.}
    \label{fig:spatial}
\end{figure}

Figure \ref{fig:spatial} gives the distribution of spatially opponent, spatially non-opponent, and spatially unresponsive cells as a function of bottleneck width for a range of ventral depths. For a small bottleneck, the vast majority of cells in the second retinal layer are spatially opponent. Conversely, cells in the first ventral layer are predominantly spatially non-opponent.
For deeper networks with less constrained bottlenecks the distributions are approximately equal in each of the layers. Almost all cells respond to some configuration of the grating stimulus, with only a small fraction of the population being spatially unresponsive. These findings are consistent with the observations that unresponsiveness has not been observed in the neuroscience literature and that the majority of cells in primate V1 are orientation tuned \citep{livingstone1984anatomy}. Regarding ventral depth, the results show a consistent 
reduction in spatial opponency in the last convolutional layer (`Retina 2' when depth is 0, `Ventral 1' when depth is 1 etc.).
There is a corresponding spike in spatial opponency in the penultimate convolutional layer (`Retina 2' when depth is 1, `Ventral 1' when depth is 2 etc.).
The average number of opponent cells in each layer does not differ greatly.

\paragraph{Colour opponency}

\begin{figure}
    \centering
    \includegraphics[width=\textwidth]{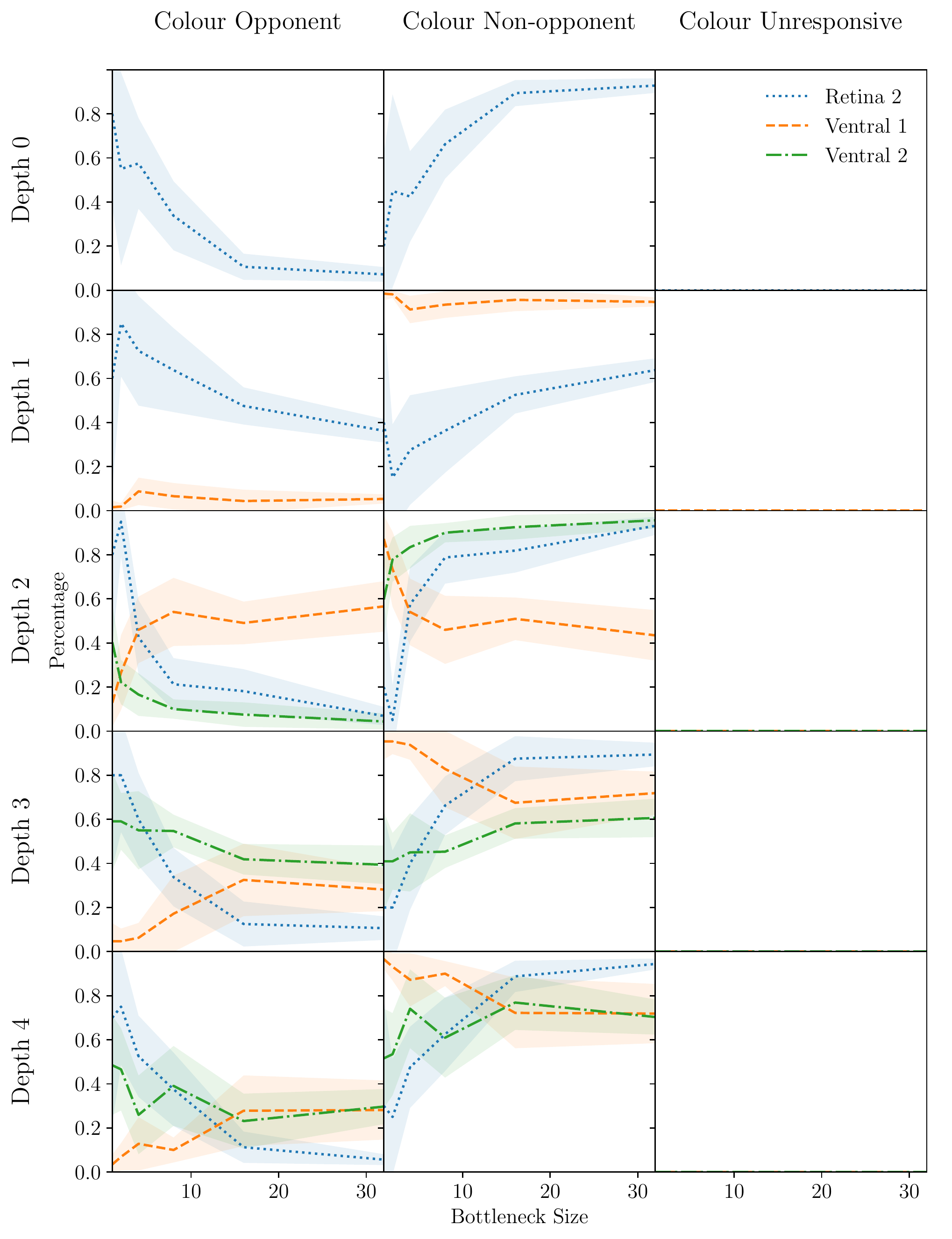}
    \caption{Distribution of colour opponent, non-opponent, and unresponsive cells in different layers of our model as a function of bottleneck width, for a range of ventral depths. Functional organisation again emerges for networks with tight bottlenecks. Furthermore, the last and penultimate convolutional layers exhibit a reduction and increase in colour opponency respectively. The echoes the spatial findings from Figure~\ref{fig:spatial}}
    \label{fig:devalois}
\end{figure}

Curves showing how the distributions of the colour opponent classes change for the second retinal and first two ventral layers as the bottleneck is increased, for a range of ventral depths, are given in Figure~\ref{fig:devalois}. As the bottleneck decreases, the second retina layer exhibits a strong increase in hue opponency, nearing $100\%$ for a bottleneck of one. Conversely, cells in the first ventral layer show a decrease in hue opponency over the same region.
For all but the tightest bottlenecks, up to half of the cells are hue non-opponent. Hue non-opponent cells show almost the exact opposite pattern to hue opponent cells.
The implication of this result is that networks with strong hue opponent representations in the bottleneck layer exhibit an increase in hue non-opponent cells in `Ventral 1'. Since this spike in opponency returns in `Ventral 2', we speculate that `Ventral 1' merely preserves the opponent code from `Retina 2' for downstream processing, and learns a set of filters that are tuned but non-opponent.
This is inconsistent with the evidence that spatially tuned cells in primate V1 are also colour opponent \citep{lennie1990chromatic,johnson2001spatial}. However, it should be stressed that our model of the primary visual cortex and ventral stream is highly simplified. In particular, we do not explicitly model the LGN or subsequent projections to different layers of V1 and greater similarity may well be observed in such a case.
Similarly to the results for spatial opponency, there is a consistent reduction / spike in hue opponency in the last and penultimate convolutional layers respectively. Averaged over bottleneck width, the number of hue opponent cells is generally lower than the number of spatially opponent cells.
\paragraph{Double opponency}

\begin{figure}
    \centering
    % \begin{subfigure}{0.375\linewidth}
    \includegraphics[width=0.45\textwidth]{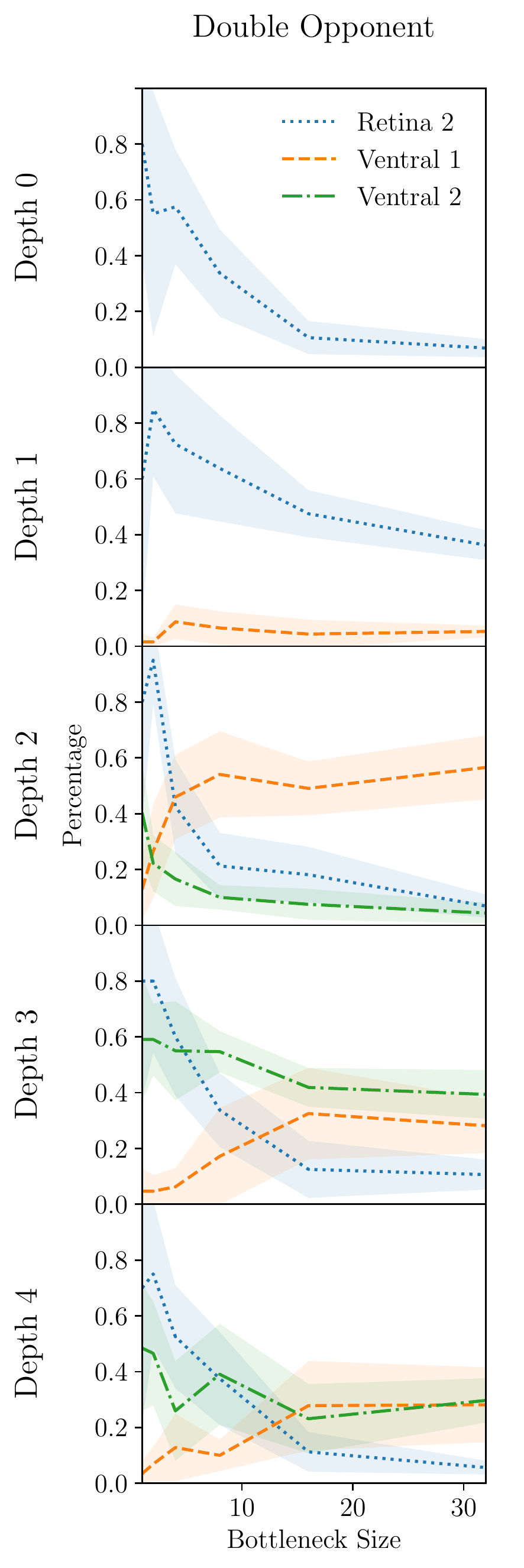}
        % \caption{Bottleneck width}  % \label{fig:double}
    % \end{subfigure}
    % \begin{subfigure}{0.375\linewidth}
    %     \includegraphics[width=\linewidth]{double_opponent_depth.pdf}
    %     \caption{Ventral depth}  % \label{fig:double}
    % \end{subfigure}
    \caption{Distribution of double opponent cells in different layers of our model as a function of bottleneck width and ventral depth. Most spatially opponent cells are also colour opponent and so these distributions bare strong similarity to those in Figures \ref{fig:spatial} and \ref{fig:devalois}}
    \label{fig:double}
\end{figure}

% 
% How
Figure \ref{fig:double} shows the distribution of double opponent cells as a function of bottleneck size and ventral depth, giving a similar picture to the spatial and hue opponency plots. The results suggest that the majority of hue opponent cells are also spatially opponent. This finding is in alignment with the observation that most hue opponent cells in the macaque V1 are also orientation selective \citep{johnson2008orientation}.

\paragraph{Types of opponency}

\begin{figure}
    \centering
    \includegraphics[width=\linewidth]{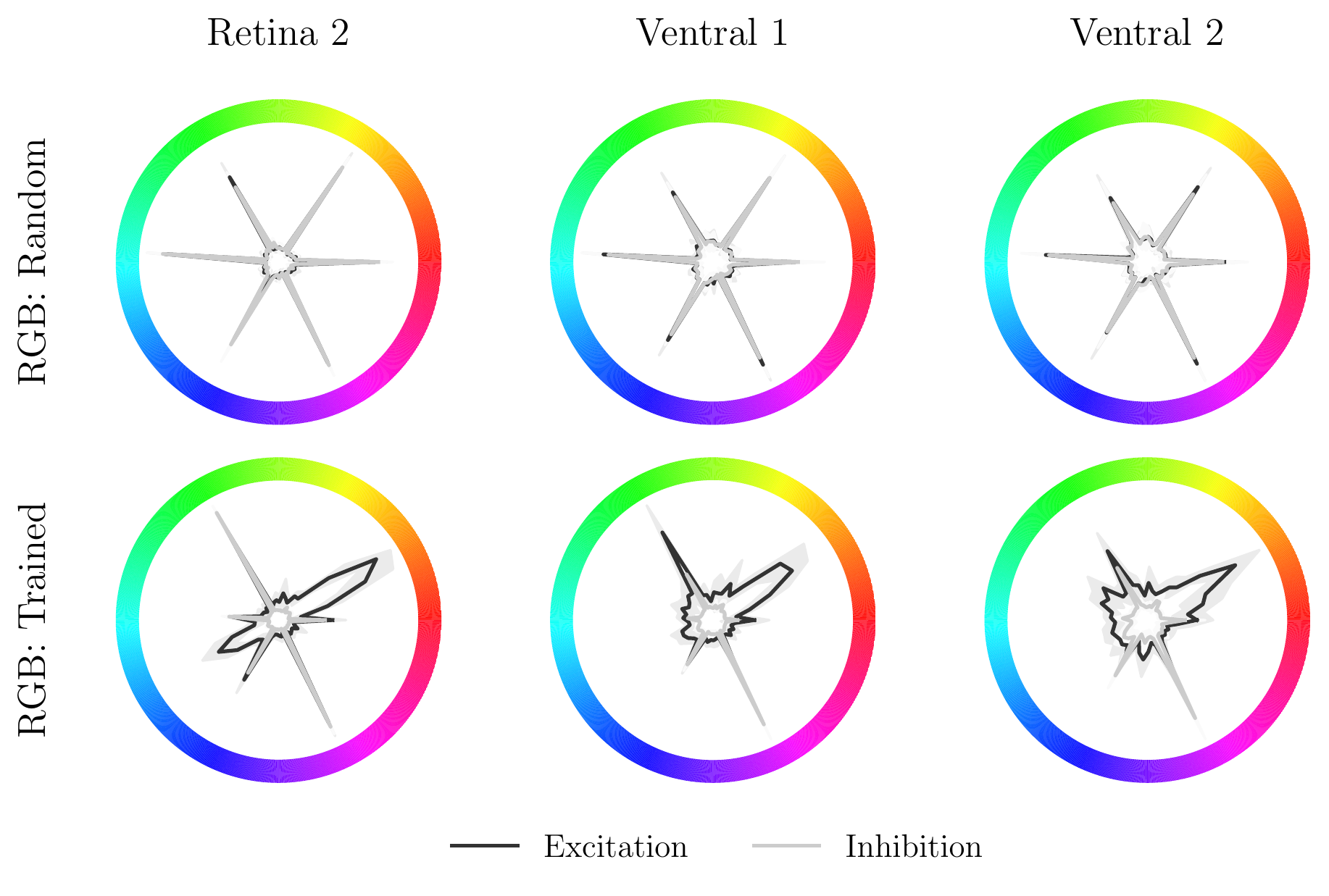}
    % \centering
    \caption{Distribution of excitatory and inhibitory hues for cells in different layers of networks with random weights, and networks trained on RGB images. Maximal excitation and inhibition before training is naturally aligned to the hues that correspond with RGB values of $255$ or $0$. Trained networks show a preference for green and magenta. Some cells are highly non-linear, maximally excited by orange / red and cyan / blue.}
    \label{fig:colours}
\end{figure}

\begin{figure}
    \centering
    \includegraphics[width=\linewidth]{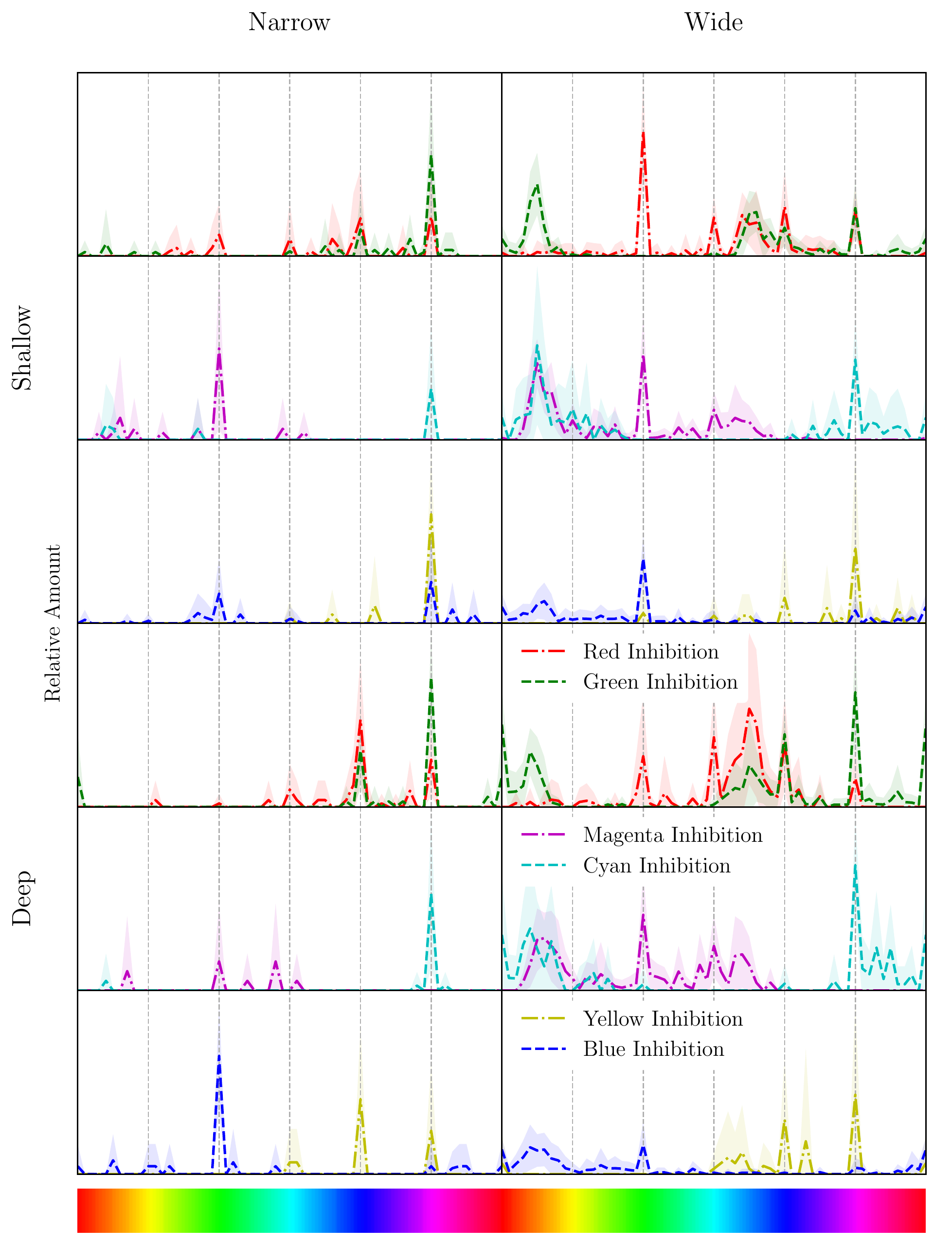}
    \caption{Conditional distribution of excitatory hues for cells that are most inhibited by red ($[315\si{\degree},45\si{\degree})$), yellow ($[45\si{\degree},75\si{\degree})$), green ($[75\si{\degree},165\si{\degree})$), cyan ($[165\si{\degree},195\si{\degree})$), blue ($[195\si{\degree},285\si{\degree})$), and magenta ($[285\si{\degree},315\si{\degree})$) for `Shallow' ($D_{VVS}\in \{0, 1\}$) and `Deep' ($D_{VVS}\in \{3, 4\}$) networks with `Narrow' ($N_{BN}\in \{1, 2, 4\}$) and `Wide' ($N_{BN}\in \{8, 16, 32\}$) bottlenecks. `Narrow' networks learn a simple colour system, with cells that are maximally excited / inhibited by extreme RGB values (dashed vertical lines). `Deep' networks show an increase in cells that are most excited by blue.}
    \label{fig:opponency_type}
\end{figure}

The plots in Figure~\ref{fig:colours} show the distribution over the hue wheel of the most excitatory and most inhibitory colours for cells in our models before and after training.
% For cells which are inhibited below zero for a range Hues, we make an unbiased estimate of the most inhibitory stimulus by choosing one at random. 
The key observation here is that maximal excitation and inhibition before training is naturally aligned to the hues that correspond with RGB values of $255$ or $0$. This is a quirk of the convolutional architecture. Since at initialisation the function of the network is smooth, if the cell is excited by a particular channel, it will be most excited when that channel is maximised and vice-versa. The effect of training, regarding both excitation and inhibition, is to reduce the proportion of cells that are tuned to red, yellow, cyan, and blue, and increase the proportion of cells that are tuned to green and magenta. In addition some cells in the bottleneck layer (`Retina 2') become most excited by orange / red and cyan / blue. Unlike the random networks, this changes as a function of depth, tending to broaden the range of excitatory and inhibitory hues. Note that this corresponds to the network learning a complex, non-linear, colour system. Cells in deeper layers are not only excited by particular channels but by the specific hue of the input.

One could speculate that the distribution in Figure~\ref{fig:colours} indicates that the type of opponency that is learned corresponds well with the cone opponency observed in primates.
However, as discussed, Figure~\ref{fig:colours} does not permit an understanding of the discrete types of opponency that are learned. Furthermore, the figure does not differentiate between the different model architectures.
In Figure~\ref{fig:opponency_type}, we additionally plot the distribution of excitatory colours for all cells in the bottleneck layer, given that they are most inhibited by red, green, magenta, cyan, yellow, and blue respectively. These are plotted for `Shallow' ($D_{VVS}\in \{0, 1\}$) and `Deep' ($D_{VVS}\in \{3, 4\}$) networks with `Narrow' ($N_{BN}\in \{1, 2, 4\}$) and `Wide' ($N_{BN}\in \{8, 16, 32\}$) bottlenecks.
% These settings simulate the two discussed settings observed in nature, with `Wide + Shallow' networks corresponding to a simple (e.g. mouse) visual system, and `Narrow + Deep' networks corresponding to a complex (e.g. primate) visual system. 

We can now observe that the primary opponent axis in our networks is green / magenta, with cells that are inhibited by red or magenta and excited by green being unique to the `Wide' / `Shallow' networks.
In addition, we can say that the majority of hue opponent cells (that is, cells in the `Narrow' networks) are channel opponent.
In the `Wide' networks, we find cells which are broadly excited by orange / red and cyan / blue. These cells persist in the first ventral layer, and are not typically present in `Narrow' networks. This suggests that the `Wide' networks are responsible for the peaks in Figure~\ref{fig:colours}. We find the presence of cells which are excited by blue and inhibited by yellow, red, and green more prominently in the `Deep' networks, with particular prevalence in the `Narrow + Deep' networks. In general, the range of excitatory and inhibitory hues is greater in the `Wide' networks, suggesting increased prevalence of complex, non-linear cells. This mirrors the finding from \citet{lindsey2019unified} that cells in this setting tend to have a non-linear receptive field.
Note that we have found that cells in the ventral layer (not included in the figure) are excited and inhibited by a much wider range of hues, particularly in the `Narrow' networks. This suggests that the bottleneck induces an efficient colour code that enables cells in later layers to become attuned to highly specific hues. Recall that we observe an increase in the proportion of colour tuned but non-opponent cells in `Ventral 1' in models with tight bottlenecks, corroborating this assertion.

% The presence of cyan and magenta corresponds well with the observable colour opponents in biological vision \citep{pridmore2011complementary}, potentially in support of the theory of complementary colours \citep{pridmore2005theory}.

\paragraph{Hue Sensitivity}

\begin{figure}
    \centering
    \includegraphics[width=\linewidth]{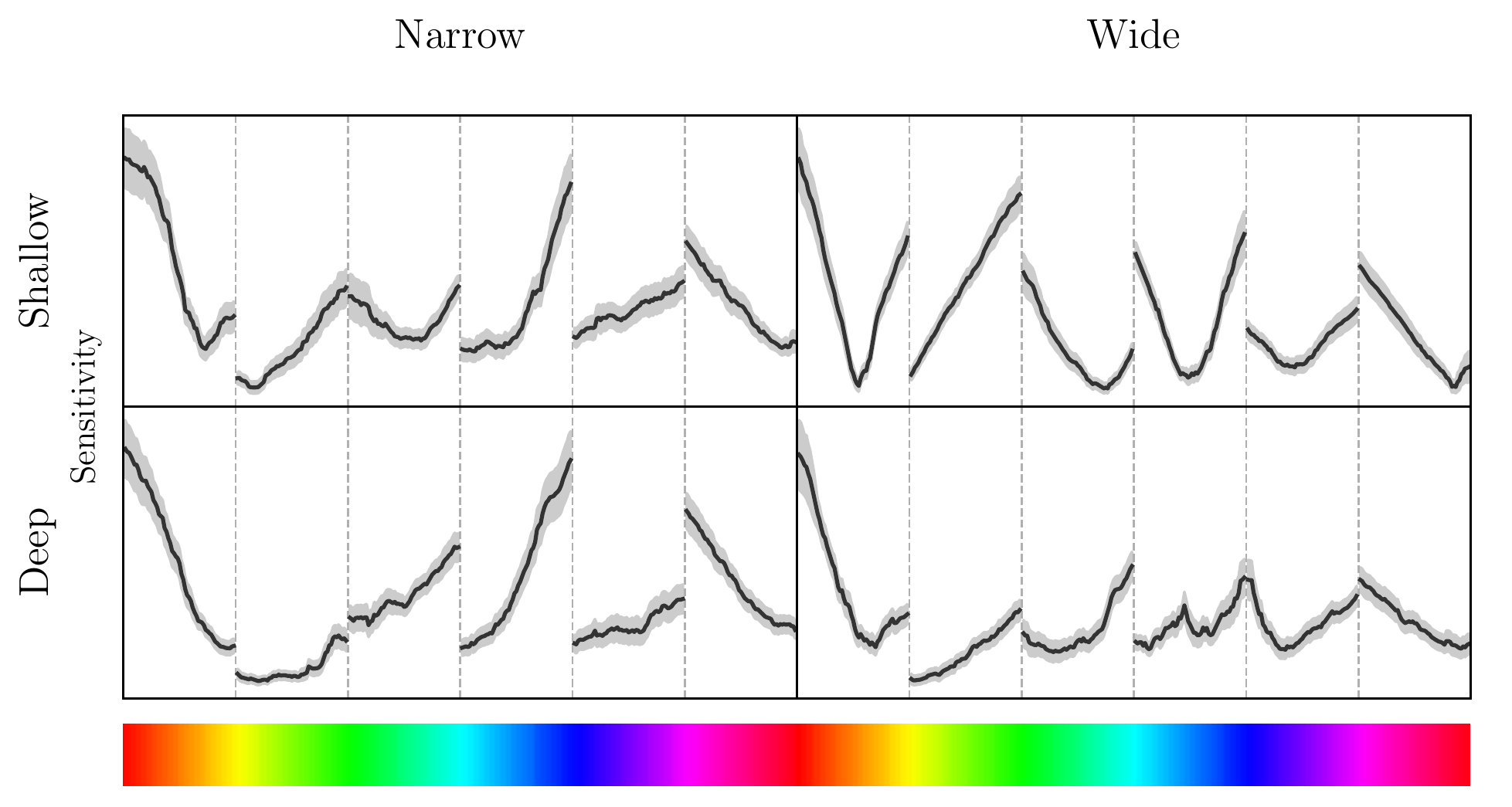}
    % \hspace{0.15em}
    \caption{Mean gradient of the sum of the bottleneck layer response with respect to hue for `Shallow' ($D_{VVS}\in \{0, 1\}$) and `Deep' ($D_{VVS}\in \{3, 4\}$) networks with `Narrow' ($N_{BN}\in \{1, 2, 4\}$) and `Wide' ($N_{BN}\in \{8, 16, 32\}$) bottlenecks. The shaded region indicates the standard error across the trained models. Discontinuities derive from the conversion from HSL to RGB. Sensitivity is an approximately linear function of hue for `Narrow' networks, and particularly in the `Narrow + Deep' setting, again showing a simple colour code in the bottleneck layer. Conversely, `Wide + Shallow' networks exhibit a highly non-linear sensitivity to hue.}\label{fig:sensitivity}
\end{figure}

% \begin{figure}
%     \centering
%     \begin{subfigure}{0.36\linewidth}
%         \includegraphics[width=\linewidth]{similarity_narrow.pdf}
%         \caption{Narrow}\label{fig:perceptual:narrow}
%     \end{subfigure}
%     % \hspace{0.15em}
%     \begin{subfigure}{0.36\linewidth}
%         \includegraphics[width=\linewidth]{similarity_wide.pdf}
%         \caption{Wide}\label{fig:perceptual:wide}
%     \end{subfigure}\\
%     \begin{subfigure}{0.36\linewidth}
%         \includegraphics[width=\linewidth]{similarity_bedford.pdf}
%         \caption{Human Observer}\label{fig:perceptual:bedford}
%     \end{subfigure}
%     \begin{subfigure}{0.36\linewidth}
%         \includegraphics[width=\linewidth]{similarity_long.pdf}
%         \caption{Natural Scenes}\label{fig:perceptual:long}
%     \end{subfigure}
%     \caption{ (\protect\subref{fig:perceptual:narrow}, \protect\subref{fig:perceptual:wide}) Mean gradient of the bottleneck layer with respect to wavelength for narrow ($N_{BN}\in \{1, 2, 4\}$) and wide ($N_{BN}\in \{8, 16, 32\}$) networks (shaded region indicates the standard error across the trained models). (\protect\subref{fig:perceptual:bedford}) Wavelength change needed to elicit a just-noticeable difference in hue for a human observer from \citet{bedford1958wavelength}. (\protect\subref{fig:perceptual:long}) Predicted similarity derived from images of natural scenes from \citet{long2006spectral}.}
%     \label{fig:perceptual}
% \end{figure}

Figure~\ref{fig:sensitivity} gives the results for the hue sensitivity experiment.
% alongside the results from \citet{bedford1958wavelength} and \citet{long2006spectral} for reference.
We again provide plots for `Shallow' and `Deep' networks with `Narrow' and `Wide' bottlenecks so that these results can be understood in the context of the previous section.
Since we are taking the gradient of the response, the sensitivity is undefined where there are discontinuities in the conversion from HSL to RGB (dotted vertical lines).
The first point to note is that the straight lines in the sensitivity curves correspond to at most a quadratic response to hue. In contrast, non-linear sensitivity curves suggest a higher order hue response.
In light of this, we can observe a general transition from highly non-linear hue response in the `Wide + Shallow' networks to a more linear hue response in the `Narrow + Deep' networks. This is in line with our findings in the previous section and again mirrors the findings from \citet{lindsey2019unified}. Regarding tuning to specific colours, \citet{lehky1990neural} note that gradient of the tuning curve (such as the curves in Figure~\ref{subfig:spect}) is maximal when the stimulus is to either side of the peak. As such, where there are peaks in the distribution of cells that are excited by a particular hue we should expect corresponding troughs in the sensitivity curve. As an example, note that the troughs in sensitivity to orange / red in the `Wide + Shallow' (and to a lesser extent in the `Narrow + Shallow' and `Wide + Deep') networks matches the peaks in excitation observed in Figure~\ref{fig:opponency_type}. A similar observation can be made regarding the trough in sensitivity to cyan / blue in the `Wide + Shallow' networks. The blue excitation that is a uniquely prominent feature in the `Narrow' networks has also resulted in a corresponding dip in sensitivity. However, this has manifested as a sudden drop rather than a smooth transition since it lies at the discontinuous boundary between blue with a green component and blue with a red component. Ultimately these results demonstrate that low level analysis of the colour tuning distributions provides valid insights into the high level functional properties of the networks.

\section{Control Experiments}

In this section we perform a series of targeted experiments to assess how well our results extend to different settings. These experiments are intended to improve our understanding of the conditions under which the various forms of opponency emerge, supporting a comprehensive discussion.

% Grey
% Random
% Imagenet
% Mosaic
% Distorted
% CIELAB
% Shuffled

\paragraph{Random weights}

\begin{figure}
    \centering
    \includegraphics[width=\textwidth]{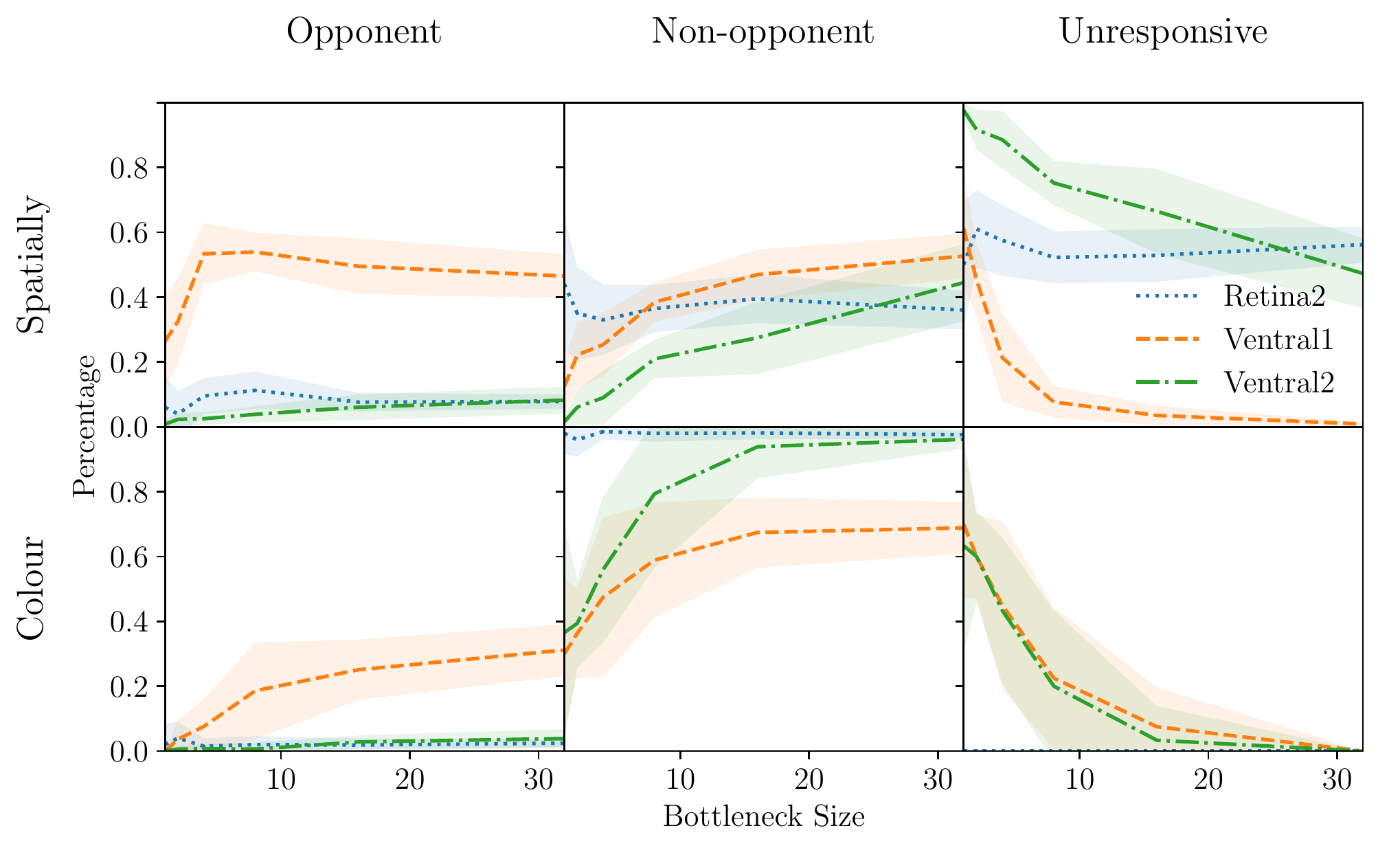}
    \caption{Distribution of spatially and colour opponent, non-opponent, and unresponsive cells in different layers of our models with Gaussian weights (mean and variance from filters of the same depth in a reference pre-trained model with $N_{BN} = 32$ and $D_{VVS} = 4$) as a function of bottleneck width. Some opponency is explained by simple statistics of the filters. Functional organisation emerges only as a result of training.}
    \label{fig:iid}
\end{figure}

Although we have presented strong evidence that cells in trained networks exhibit spatial, colour, and double opponency, we have not yet demonstrated that this is learned. To determine if this opponency is learned, we require a demonstration that it is not present at initialisation (when the weights are random). We have therefore performed our experiments on randomly initialised models. We find that networks with random weights (that is, following the Xavier initialisation \citep{glorot2010understanding}) never exhibit spatial or hue opponency. Instead, most cells are non-opponent and their distribution over the layers does not change significantly with the bottleneck size. These results demonstrate that all of the opponency in our networks is learned. However, it could be the case that opponency derives from simple statistics of the convolutional filters. In order to understand this further we experimented with networks whose filter weights are Gaussian with the same mean and variance as the filters of the same depth in a reference pre-trained model. The results for this experiment are given in Figure~\ref{fig:iid}. Although we do find some opponency in this case, we do not find the same structure. Of particular note are the unresponsive, not present in the trained networks. These findings reflect the fact that a degree of structure in the receptive fields is required in order for a cell to exhibit a consistent opponent or non-opponent response. 

\paragraph{Greyscale}

% \begin{subfigure}{0.15\linewidth}
%         \includegraphics[width=\linewidth]{135_grey.pdf}
%         \caption{$\theta=\ang{135}$}
%     \end{subfigure}

\begin{figure}
    \centering
    \begin{subfigure}{\linewidth}
        \includegraphics[width=\textwidth]{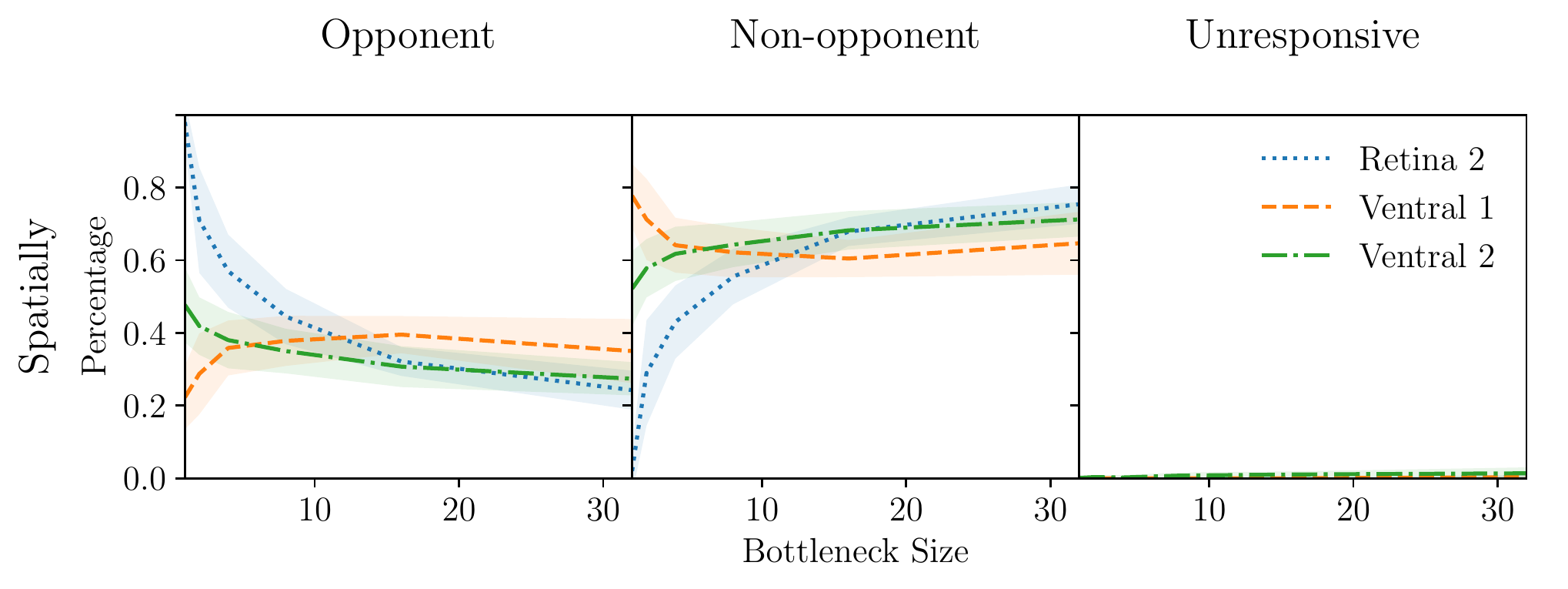}
        \caption{Greyscale}
        \label{fig:greyscale}
    \end{subfigure}
    
    % \begin{subfigure}{\linewidth}
    %     \includegraphics[width=\textwidth]{opponency_random.pdf}
    %     \caption{Random weights}
    %     \label{fig:random}
    % \end{subfigure}
    
    \begin{subfigure}{\linewidth}
        \includegraphics[width=\textwidth]{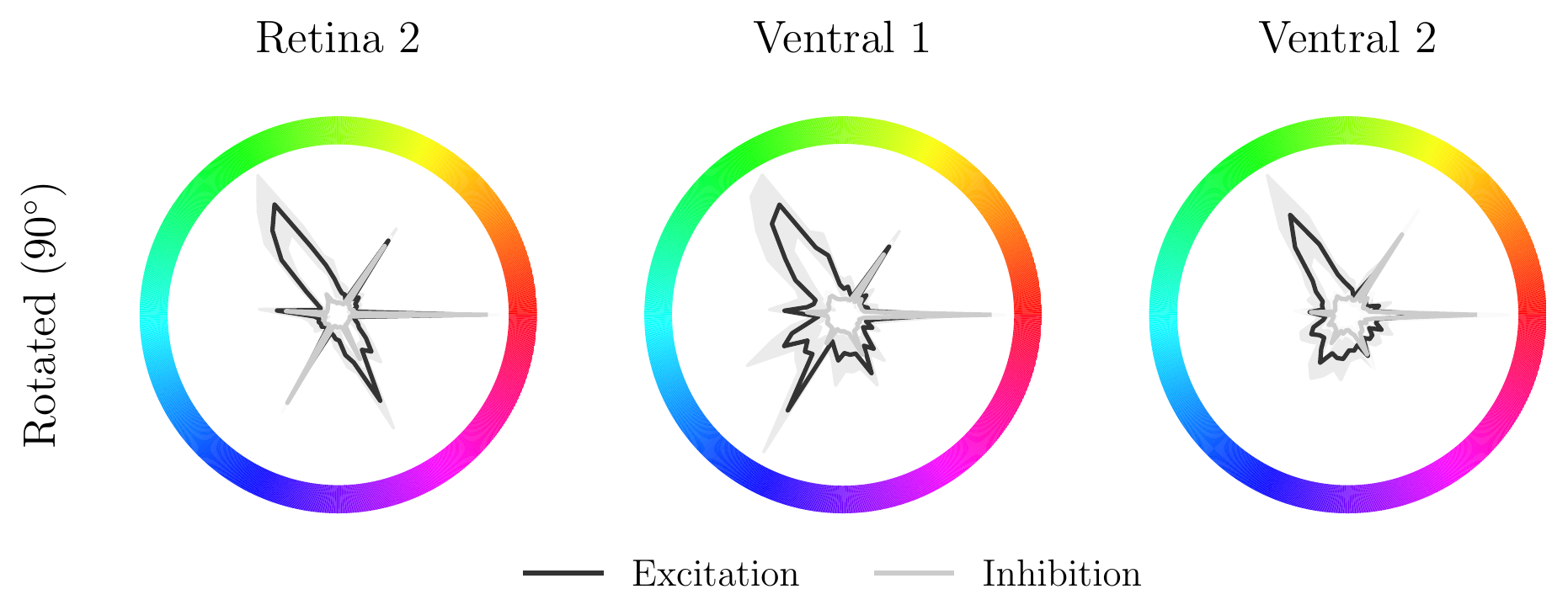}
        \caption{Distorted colour}
        \label{fig:distorted}
    \end{subfigure}
    \caption{(\protect\subref{fig:greyscale}) Distribution of spatially opponent, non-opponent, and unresponsive cells in different layers of our model as a function of bottleneck width, for models trained with greyscale images showing that the known spatial opponency from \citet{lindsey2019unified} is detected by our method. (\protect\subref{fig:distorted}) Distribution of excitatory and inhibitory hues for cells in different layers of networks trained on images with distorted colour (hue rotation of $90\si{\degree}$). The most prevalent excitatory and inhibitory colours are aligned with the RGB extremes closest to a $90\si{\degree}$ rotation of the peaks in Figure~\ref{fig:colours}.}
\end{figure}

As previously mentioned, in addition to the colour models we trained a batch of models with greyscale inputs. The results in Figure~\ref{fig:greyscale} validate that spatially opponent cells still emerge and have a similar distribution throughout the layers as that of cells in models trained with RGB inputs. Furthermore, this validates our classification approach since, from \citet{lindsey2019unified}, it is known that the Retina-Net model learns centre-surround and oriented edge (both spatially opponent) filters with greyscale input.

\paragraph{Distorted colour}

To further explore the idea that the opponency in our networks derives from the statistics of the data, we trained a batch of models on images with distorted colour. Specifically, we convert the images into HSV space and offset the hue channel by $90\si{\degree}$, before converting back into RGB and forwarding to the network. Our interest here is not in whether opponency emerges, but in the effect this distortion has on it. Figure~\ref{fig:distorted} shows the distribution of excitatory and inhibitory colours in networks trained with distorted inputs. Here, the most prevalent excitatory and inhibitory colours are aligned with the RGB extremes closest to a $90\si{\degree}$ rotation of the peaks in Figure~\ref{fig:colours}.
This is consistent with our observation that the vast majority of colour opponent neurons are channel opponent.
In contrast, the additional excitation peak has been rotated by exactly $90\si{\degree}$ from orange / red to green. This demonstrates that the cells which are excited by specific hues emerge as a result of the statistics of the data, not of the input colour space.

\paragraph{CIELAB space}

\begin{figure}
    \centering
    \begin{subfigure}{\linewidth}
        \includegraphics[width=\textwidth]{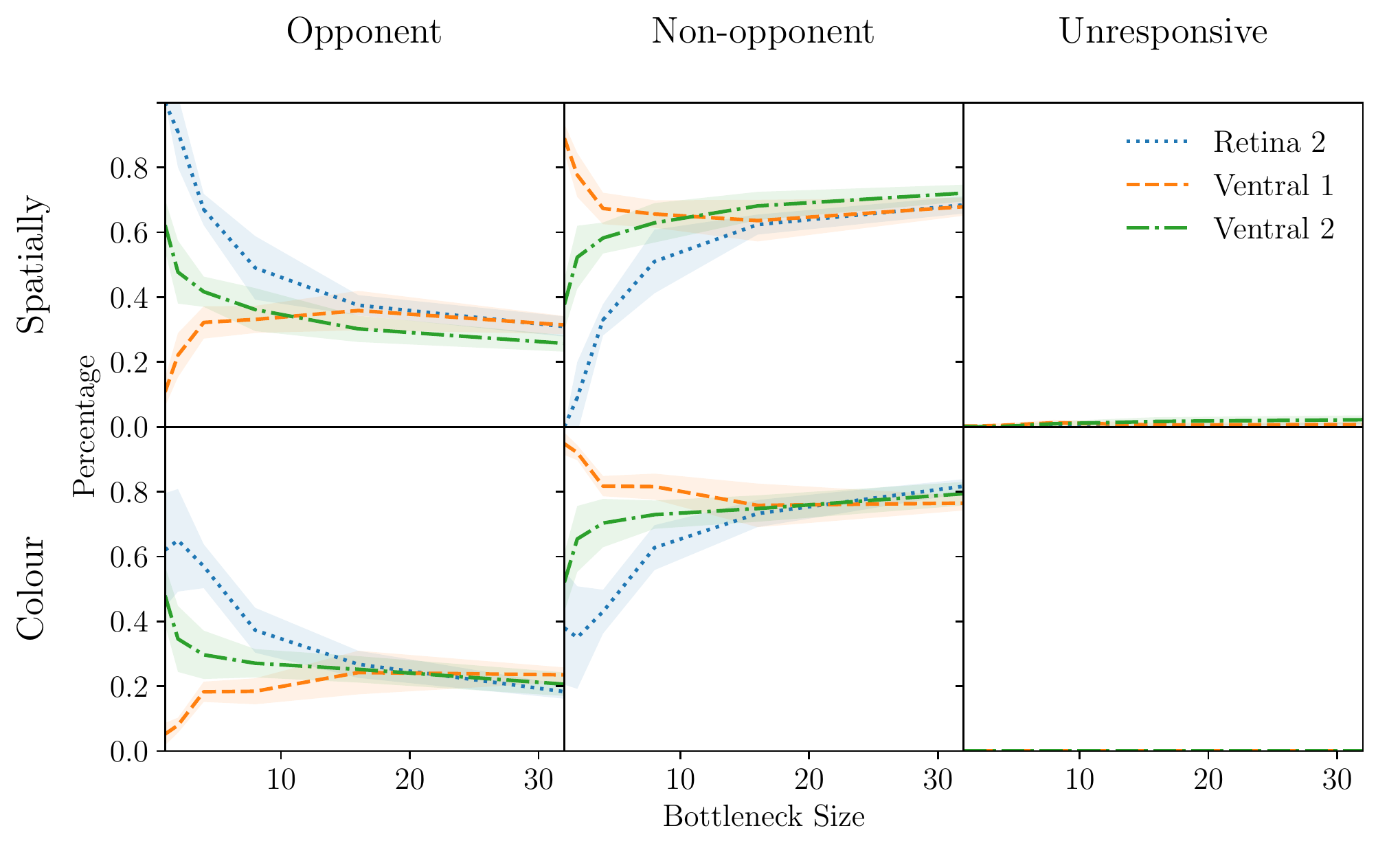}
        \caption{Opponency}
        \label{fig:lab}
    \end{subfigure}
    \begin{subfigure}{\linewidth}
        \includegraphics[width=\textwidth]{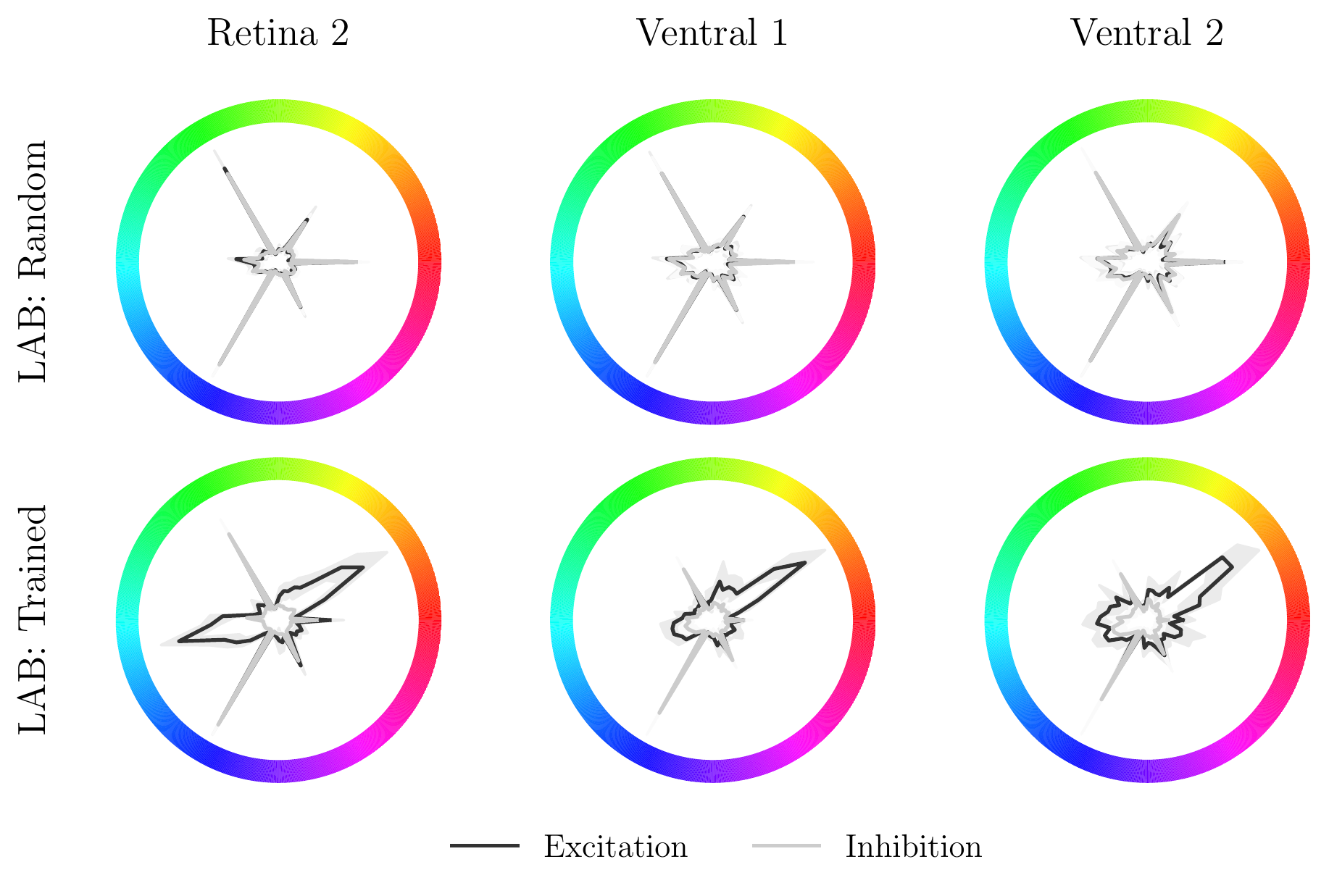}
        \caption{Excitatory and inhibitory colours}
        \label{fig:colours_lab}
    \end{subfigure}
    \caption{(\protect\subref{fig:lab}) Distribution of spatially and colour opponent, non-opponent, and unresponsive cells in different layers of models trained on images in LAB space as a function of bottleneck width, showing that functional organisation is not unique to RGB. (\protect\subref{fig:colours_lab}) Excitatory / inhibitory hues in LAB space for random and trained networks. Training increases prevalence of blue / green, and excitation by orange / red and cyan / blue.}
\end{figure}

% \begin{figure}
%     \centering
%     \includegraphics[width=\linewidth]{excitatory_colours_lab.pdf}
%     % \centering
%     \caption{Distribution of excitatory and inhibitory hues for cells in different layers of networks with random weights, and networks trained on CIELAB images.}
%     \label{fig:colours_lab}
% \end{figure}

In a similar vein to our experiments with distorted colour, we now perform experiments to validate whether opponency is still a feature in networks trained on images in the CIELAB colour space. The CIELAB colour space encodes colour in terms of lightness (L*), and two opponent axes: green / red (a*), and blue / yellow (b*). Each axis is non-linear such that uniform changes in CIELAB space correspond to uniform perceptual changes in colour.
This will allow us to understand if functional organisation still emerges when receptive fields are naturally opponent, that is, structure would require the cells to learn to `ignore' the inherent opponency of the a* and b* channels.
Figure \ref{fig:lab} shows the distribution of spatially and colour opponent cells in this setting. The distribution is nearly identical to that of networks trained on images in RGB space, with the same characteristic functional organisation. In Figure \ref{fig:colours_lab}, we plot the distribution of most excitatory and inhibitory colours in CIELAB space for random and trained networks. We again find that the distribution for random networks naturally aligns to hues which represent extreme values in the input colour space; CIELAB encodes colour on green / red and blue / yellow axes. Following training, we find that the majority of cells are most excited or inhibited by either green or blue. This bares some similarity to the RGB networks, which aligned to green and magenta following training. Again in accordance with the RGB networks, we find cells which are most excited by orange / red or cyan / blue, further showing that tuning to these particular colours is an artefact of the data rather than the colour space.

\paragraph{Street view house numbers}

\begin{figure}
    \centering
    % \begin{subfigure}{\linewidth}
    \includegraphics[width=\textwidth]{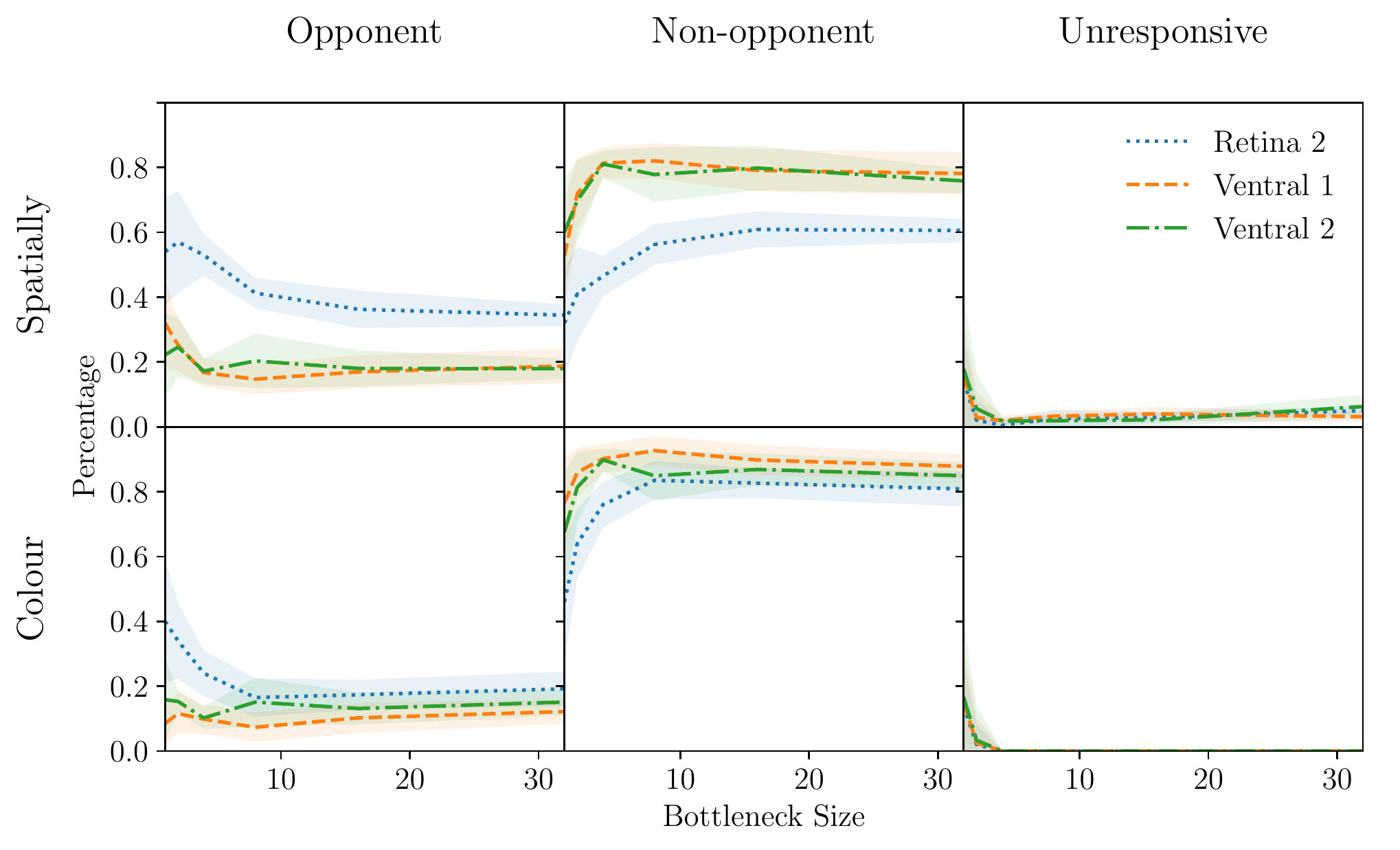}
        % \caption{Spatial and spectral opponency}
        % \label{fig:svhn}
    % \end{subfigure}
    % \begin{subfigure}{0.375\linewidth}
    %     \includegraphics[width=\linewidth]{double_opponent_svhn.pdf}
    %     \caption{Double: bottleneck width}
    %     \label{fig:svhn:double}
    % \end{subfigure}
    % \begin{subfigure}{0.375\linewidth}
    %     \includegraphics[width=\linewidth]{double_opponent_depth_svhn.pdf}
    %     \caption{Double: ventral depth}
    %     \label{fig:svhn:depth}
    % \end{subfigure}
    \caption{Distribution of spatially and colour opponent, non-opponent, and unresponsive cells in different layers of models trained on Street View House Numbers (SVHN)~\citep{netzer2011reading} as a function of bottleneck width. Spatial opponency is present, with a similar distribution to the networks trained on CIFAR-10. Colour opponency is generally lower, increasing only slightly for networks with narrow bottlenecks. 
    % (\protect\subref{fig:imagenet:double}, \protect\subref{fig:imagenet:depth}) Distribution of double opponent cells in different layers of models trained on SVHN as a function of bottleneck width and ventral depth respectively.
    }\label{fig:svhn}
\end{figure}

In addition to our experiments with CIFAR-10, we have trained a batch of networks on the Street View House Numbers (SVHN) data set~\citep{netzer2011reading}. This is a digit recognition (10 classes) problem with the same spatial resolution (32 $\times$ 32) as CIFAR-10. The distributions of the different cell types for these models are shown in Figure~\ref{fig:svhn}. The results show that spatial opponency is abundant in the second retina layer and increases for models with tight bottlenecks. In the first two ventral layers, the proportion of spatially opponent cells is generally lower ($\approx 20\%$) and does not change significantly with bottleneck size. Colour opponency is muted in comparison to the CIFAR-10 experiments and increases in the second retina layer only slightly for tight bottlenecks.
This is unsurprising since colour is not expected to be an important feature in the house number recognition problem.
The largest networks in this setting achieved over $90\%$ accuracy.
% The double opponent results show that
% - most spectrally opponent cells also spatially opponent
% - many spatially opponent cells that are not spectrally opponent

\paragraph{ImageNet}

\begin{figure}
    \centering
    % \begin{subfigure}{\linewidth}
    \includegraphics[width=\textwidth]{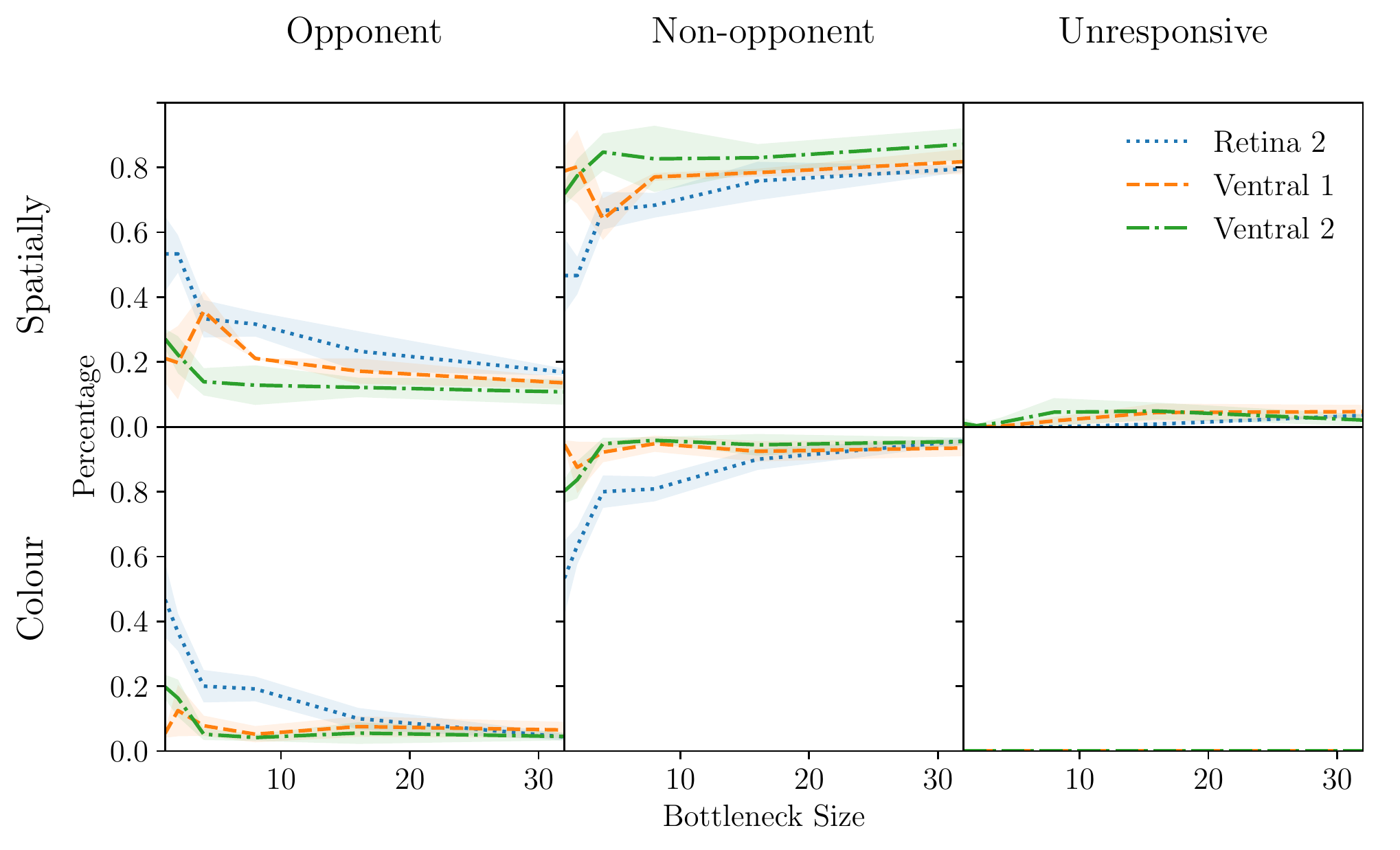}
        % \caption{Spatial and spectral opponency}
        % \label{fig:imagenet}
    % \end{subfigure}
    % \begin{subfigure}{0.375\linewidth}
    %     \includegraphics[width=\linewidth]{double_opponent_imagenet.pdf}
    %     \caption{Double: bottleneck width}
    %     \label{fig:imagenet:double}
    % \end{subfigure}
    % \begin{subfigure}{0.375\linewidth}
    %     \includegraphics[width=\linewidth]{double_opponent_depth_imagenet.pdf}
    %     \caption{Double: ventral depth}
    %     \label{fig:imagenet:depth}
    % \end{subfigure}
    \caption{Distribution of spatially and colour opponent, non-opponent, and unresponsive cells in different layers of models trained on ImageNet~\citep{russakovsky2015imagenet} as a function of bottleneck width, showing how our findings transfer to a higher resolution setting. There is an increase in opponency for narrow bottlenecks which decays rapidly. Emergent organisation is observed only partially in the networks with the tightest bottlenecks.
    % (\protect\subref{fig:imagenet:double}, \protect\subref{fig:imagenet:depth}) Distribution of double opponent cells in different layers of models trained on ImageNet as a function of bottleneck width and ventral depth respectively.
    }\label{fig:imagenet}
\end{figure}

Our experiments thus far have focused on low resolution (32 $\times$ 32) images.
It is now important to understand whether our findings generalise to a higher resolution setting.
To that end, we have trained networks on the ImageNet Large Scale Visual Recognition Challenge (ILSVRC) data set \citep{russakovsky2015imagenet} at a resolution of 128 $\times$ 128. Due to hardware constraints, we perform only $3$ repeats across the full range of ventral depths and bottleneck widths. We adapt the Retina-Net model slightly, adding average pooling with a window of size $4$ before the first fully connected layer. Figure~\ref{fig:imagenet} gives the distributions of the different spatial and colour cell types respectively for models trained on ImageNet. As with CIFAR-10, the results show an increase in the proportion of spatially and colour opponent cells in the bottleneck layer of networks with a tight bottleneck. Unlike the CIFAR-10 results, this opponency decays rapidly and the emergent organisation is observed only partially in the networks with the tightest bottlenecks. The percentage of opponent cells was generally lower than in networks trained on CIFAR-10. This could be related to the fact that the Retina-Net model does not effectively `fit' to ImageNet; the trained models achieved an accuracy between $5\%$ and $20\%$.
Although the number of double opponent cells in this setting (not shown in the figure) is much lower, the vast majority of spatially opponent cells are also colour opponent. We do not find a spike in opponency in the penultimate convolutional layer in general in the ImageNet trained models.

\paragraph{Intel scene classification}

\begin{figure}
    \centering
    % \begin{subfigure}{\linewidth}
    \includegraphics[width=\textwidth]{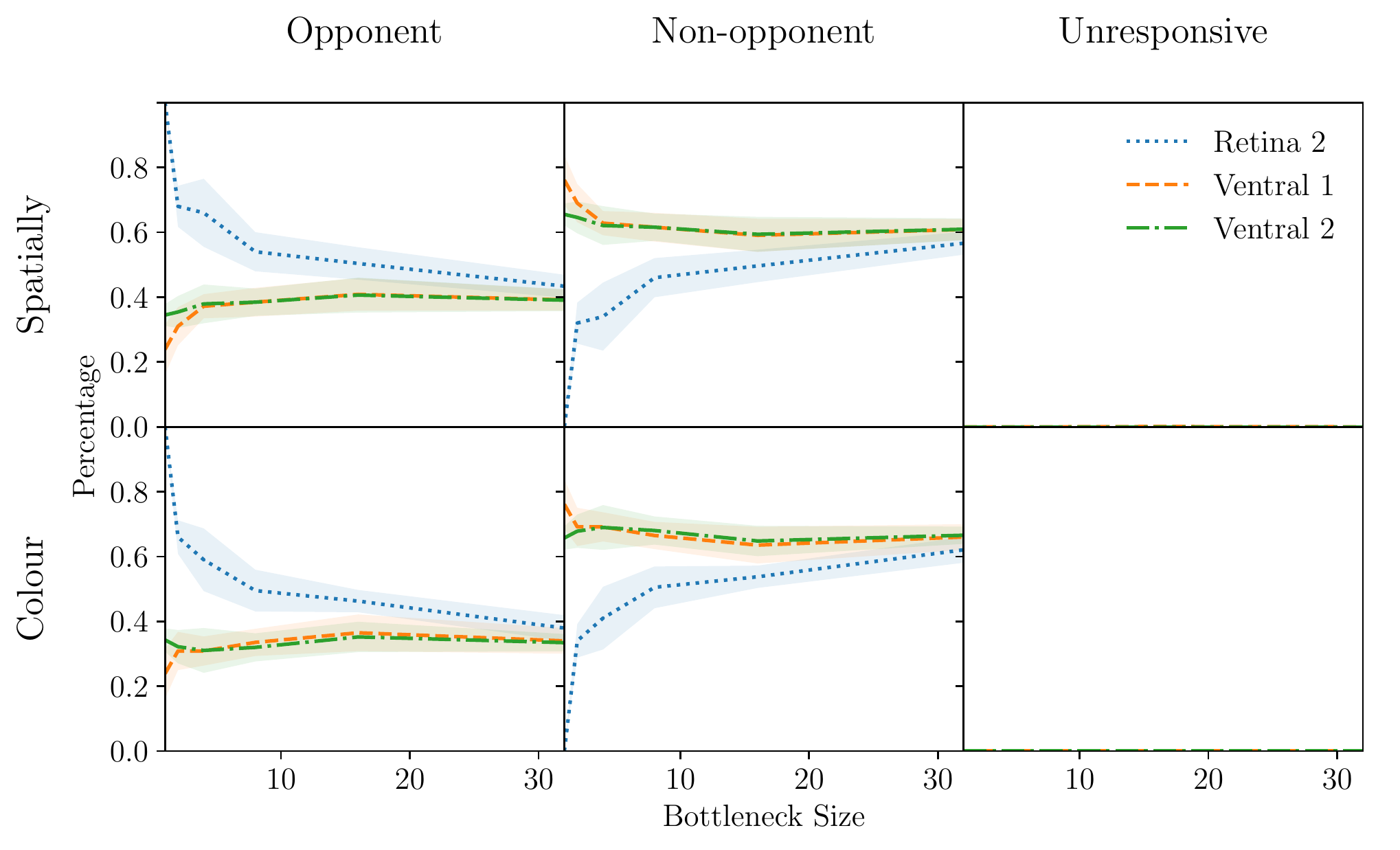}
        % \caption{Spatial and spectral opponency}
        % \label{fig:intel}
    % \end{subfigure}
    % \begin{subfigure}{0.375\linewidth}
    %     \includegraphics[width=\linewidth]{double_opponent_intel.pdf}
    %     \caption{Double: bottleneck width}
    %     \label{fig:intel:double}
    % \end{subfigure}
    % \begin{subfigure}{0.375\linewidth}
    %     \includegraphics[width=\linewidth]{double_opponent_depth_intel.pdf}
    %     \caption{Double: ventral depth}
    %     \label{fig:intel:depth}
    % \end{subfigure}
    \caption{Distribution of spatially and colour opponent, non-opponent, and unresponsive cells in different layers of models trained on the Intel scene classification challenge data set~\citep{intel2018} as a function of bottleneck width. With fewer classes (6 in this case), the number of opponent cells is much higher. The distribution of opponent cells in `Retina 2' bares strong similarity with the results from CIFAR-10. This does not extend to the ventral layers, which have near-identical cell distributions. 
    % (\protect\subref{fig:imagenet:double}, \protect\subref{fig:imagenet:depth}) Distribution of double opponent cells in different layers of models trained on the Intel data set as a function of bottleneck width and ventral depth respectively.
    }\label{fig:intel}
\end{figure}

In addition to our experiments with ImageNet, we have trained a batch of models on the Intel scene classification challenge~\citep{intel2018} data set. This is a natural scene classification problem with 6 classes and the same spatial resolution as ImageNet. As such, these models allow us to explore opponent cell types in a high resolution setting where the model obtains stronger performance (up to $80\%$ for the largest models). The results in Figure~\ref{fig:intel} show that models trained in this setting exhibit a much higher percentage of spatially and colour opponent cells than in models trained on ImageNet, particularly in the second retina layer. We again find that the percentage of opponent cells in the first two ventral layers is equal and broadly independent of bottleneck size, not showing the emergent structure observed in CIFAR-10 except in the extreme case of $N_{BN} = 1$.

\paragraph{Classifying mosaics}

\begin{figure}
    \centering
    \includegraphics[width=\linewidth]{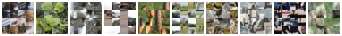}\\[0.4cm]
    \includegraphics[width=\linewidth]{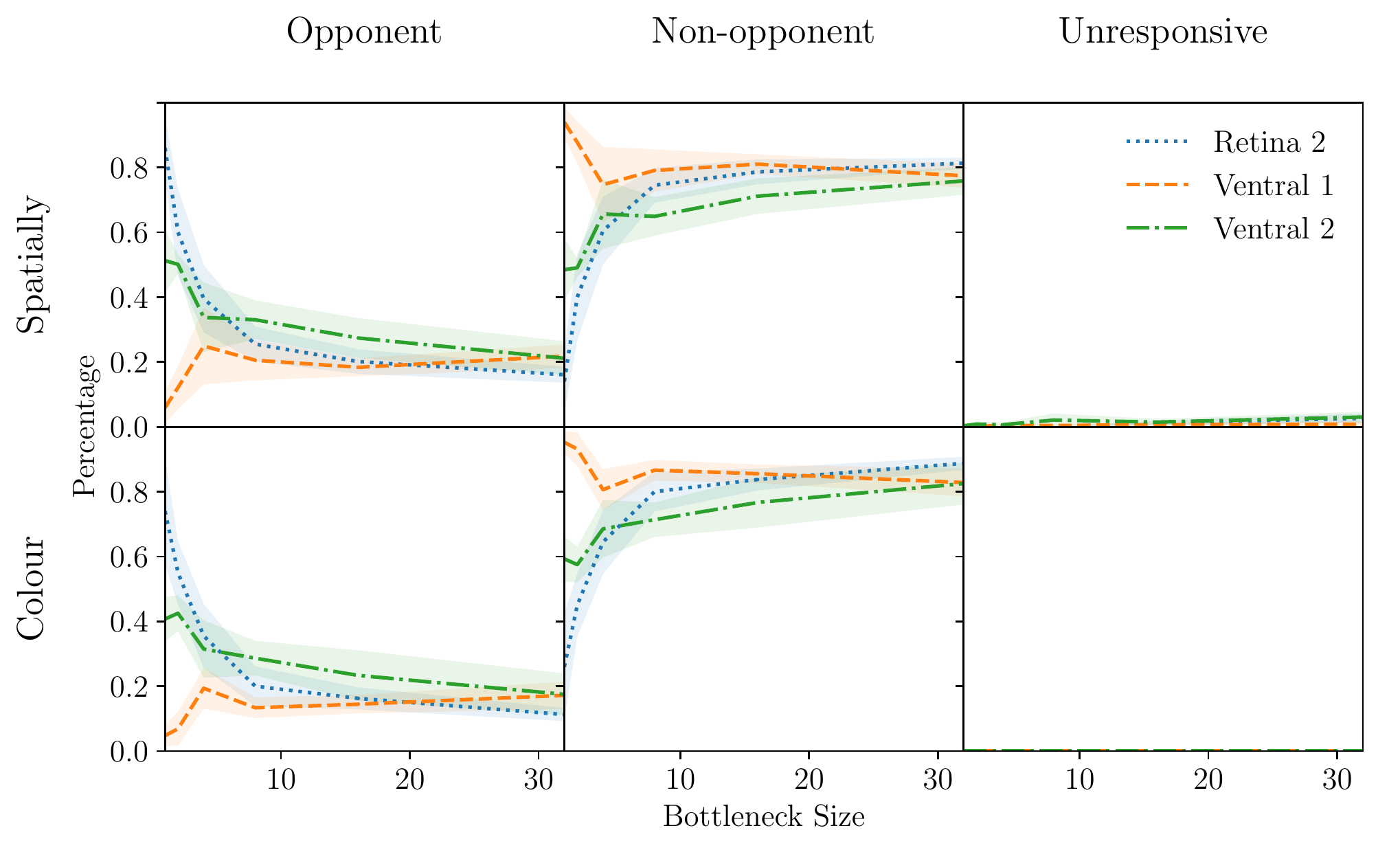}
    \caption{Distribution of spatially and colour opponent, non-opponent, and unresponsive cells in different layers of models trained on mosaic images as a function of bottleneck width with example mosaic images. These results show that when the spatial structure of the input is removed some spatial opponency, particularly in `Retina 2', is removed also. Colour opponency is similarly affected, suggesting a complex dependence between spatial and colour processing.}\label{fig:mos}
\end{figure}

We also performed experiments to determine whether there are conditions under which certain types of opponency can be removed. To attempt to ablate spatial oppponency, we trained a batch of models to classify mosaic images. These are images that have been separated into smaller squares which have then been shuffled (see examples in Figure~\ref{fig:mos} for reference). Figure~\ref{fig:mos} gives the distribution of spatially and colour opponent cells in these networks. The figure shows that some of the spatial opponency is removed in this setting. Notably, `Retina 2' exhibits a moderately lower proportion of spatial opponency, such that it is now in line with `Ventral 2'. This could be due to the fact that the impact of the mosaic images depends heavily on the size of the receptive field. We further note that colour opponency is affected to the same extent as spatial. This suggests that the efficacy of colour opponent cells is in some way reduced in the mosaic setting.

\paragraph{Shuffled colour channels}

\begin{figure}
    \centering
    \includegraphics[width=\linewidth]{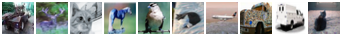}\\[0.4cm]
    \includegraphics[width=\linewidth]{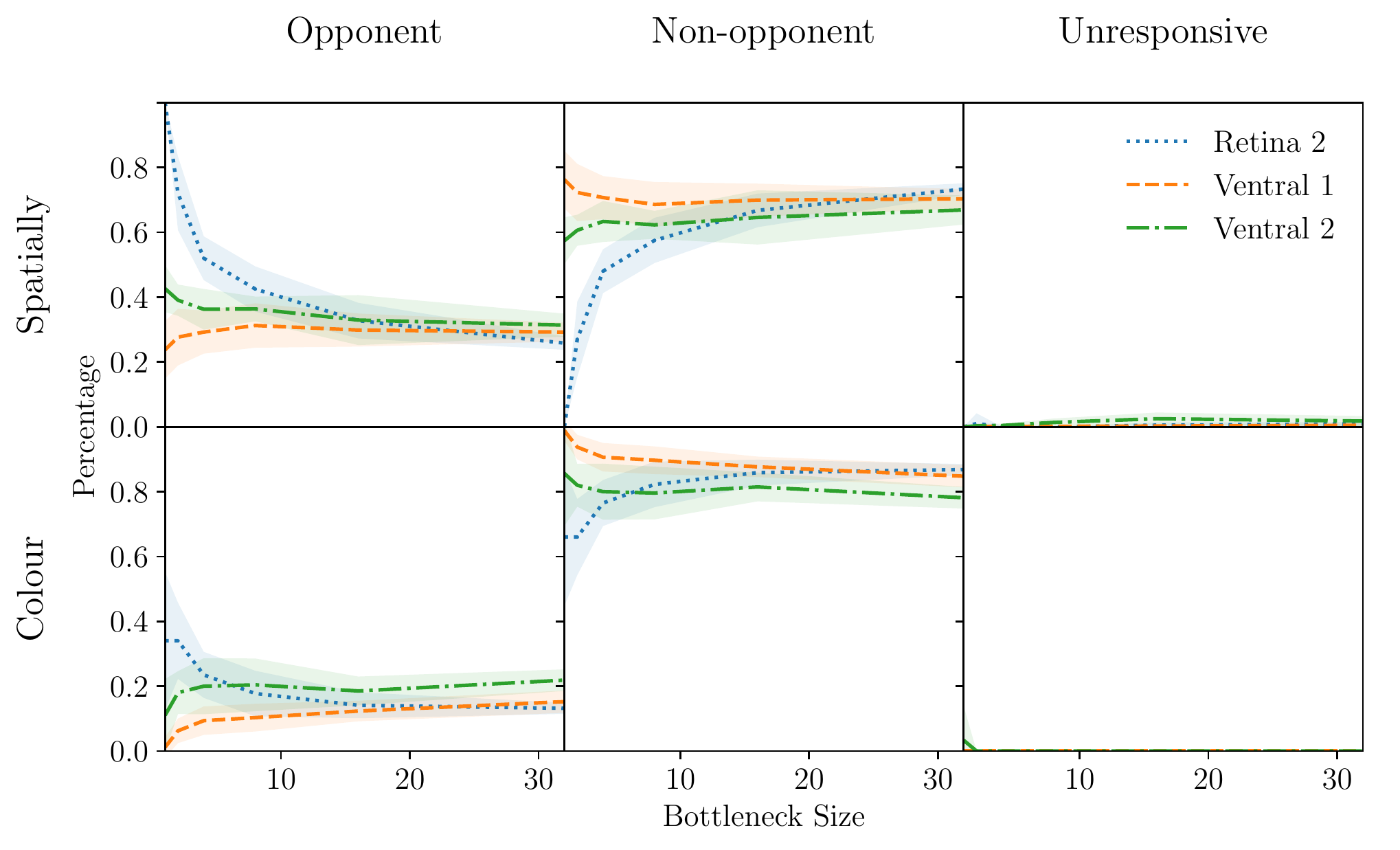}
    \caption{Distribution of spatially and colour opponent, non-opponent, and unresponsive cells in different layers of models trained on images with shuffled colour channels as a function of bottleneck width with example shuffled images. When consistent colour information is removed, most colour opponency is also removed. Spatial opponency remains.}\label{fig:shuffled}
\end{figure}

To attempt to ablate colour opponency, we remove colour information by randomly shuffling the channels of inputs to the network (see examples in Figure~\ref{fig:shuffled} for reference). The resultant distribution plots in Figure~\ref{fig:shuffled} show that this alteration removes the vast majority of colour opponent cells, whilst spatial opponency remains.
Since the information present in shuffled images is the same, this experiment demonstrates that colour opponency arises out of a need to consistently infer the colours in the inputs. We speculate that this aids in classification since each class will be associated with a set of features that vary both spatially and in hue. By shuffling the channels we remove the ability to repeatedly associate a particular input tuple with a particular class. This view is supported by the fact that the models in this setting generally reached a lower accuracy than the models trained on standard colour images and sometimes failed to match the models trained on greyscale images.

\section{Discussion}

Equipped with the results of our experiments, we now discuss the conclusions which can be drawn regarding spatial and colour processing in convolutional neural networks. In addition, we suggest possible directions for future work building on these findings. Our primary finding is that the addition of a bottleneck in the Retina-Net model induces functional organisation when trained on colour (RGB) CIFAR-10.
We have further shown that this finding generalises to networks trained on images in the CIELAB colour space. There is some evidence that this result differs when the networks are trained on other data sets; although the key finding, that structure emerges only with the tightest bottlenecks, remains.
In the case of ImageNet, more experimentation with a model capable of fitting to the data would be required to understand this fully. Regarding network depth, our experiments have uncovered an increase in the number of opponent cells in the penultimate convolutional layer of the network and a corresponding decrease in the last convolutional layer. Our experiments with random networks demonstrate that all of the discussed opponency is learned and that most opponency is not a result of simple statistics of the weights.

In addition to these high level observations, we have shown that an analysis based on approaches from neuroscience can yield a rich understanding of the function performed by a trained network. For example, we have shown that the deep Retina-Net model with a tight bottleneck learns a set of double opponent filters in the bottleneck layer, followed by a set of spatially and colour tuned but non-opponent filters in the first ventral layer, with opponency returning in the second ventral layer.
Cells which are maximally excited by blue are a unique feature of these networks not present when the bottleneck is relaxed. Furthermore, these networks tend to learn linear, channel opponent, neurons rather than neurons which are opponent to specific hues. We speculate that this is due to the increased need to learn an efficient colour code in the tight bottleneck case.

The key implication of our core findings is that the model architecture can be the source of an inductive bias towards the number of opponent cells. While this finding alone may be of interest, whether it is of any practical significance depends on whether opponency is desirable. By virtue of the fact that opponent cells represent a more efficient encoding of the input, one might speculate that an increase in opponency could lead to increased generalisation performance. This view is mildly supported by the plot in Figure~\ref{fig:model-acc:colour}, where the networks with $N_{BN} = 2$ and $D_{VVS} = 4$ obtained the highest accuracy.
We further suggest that opponent cells may be of greater utility in applications such as transfer learning.
Specifically, one can envisage a scenario where the pre-bottleneck weights are fixed, and the post-bottleneck weights are updated to fit a new data set. Before such a setting could be considered, our findings would need to be demonstrated on a much more capable network architecture that can obtain competitive performance on standard data sets. The finding that the penultimate layer exhibits a spike in opponency may provide insight to the efficacy of layer-wise training procedures such as deep cascade learning \citep{marquez2018deep}.
Note that cascade learning has been found to work well with transfer learning \citep{du2019transfer}. Based on the evidence presented in this work, one might speculate that cascade learning increases the number of opponent cells, and that these cells perform well for the transfer learning task. That said, and as previously discussed, whether the opponent cells in later layers inherit the same properties as opponent cells in earlier layers remains to be determined.

We have also demonstrated a number of similarities between the learned representations of our networks and representations observed in nature.
The large amount of double opponent cells we find in the retina layer of networks with tight bottlenecks is consistent with what is known about cells in the retina and LGN \citep{hubel2004brain}. There are some consistencies and some inconsistencies between the ventral layers of the model and what is known about spatial and colour processing in the visual cortex. However, as discussed, it is not clear that the ventral convolutional architecture is a good analogue of the structure of the visual cortex and so such comparisons should be treated with scepticism.
Our finding that the type of opponency learned is aligned with extreme values in the input colour space accords with the physiological finding that opponency in early stages of the visual pathway is aligned with cone responses \citep{Shevell:17}.
% We have additionally shown accordance between the spectral sensitivity of our networks and that of human observers.
% We have additionally demonstrated that the hue sensitivity of networks designed to mimic the primate visual system (deep networks with tight bottlenecks) exhibits the strongest similarity to the spectral sensitivity of a human observer.

The consequence of these demonstrations is not to suggest that convolutional neurons and biological neurons are similar. Instead, we have shown that similarity in the data space, architecture, and problem setting can give rise to similarity in the emergent functional properties.
% As such, we have validated the use of methods from the neuroscience literature for the analysis of deep CNNs.
In addition to the above, we have demonstrated some settings in which opponency is either hindered or removed entirely. This kind of controlled experiment may enable the exploration of hypotheses relating to the neuroscience of vision. Specifically, through construction of a data set which mimics an environment, or an architecture which mimics an anatomy, one might seek a better explanation of the differences in visual processing between species. This potential is hinted at by our experiments with SVHN, which show that networks trained on the digit recognition task have fewer colour opponent cells.

In conclusion, our considerations here provide a strong mandate for future research across a range of interests. As mentioned, work should be conducted to understand whether the presence of opponent cells promotes increased adversarial robustness. Such research will require the ability to apply our methods to state-of-the-art architectures in order to be of practical relevance. In particular, it remains to be seen whether the introduction of a bottleneck is enough to promote opponency in more complex architectures. Indeed, this may require more sophisticated approaches such as the aforementioned cascade learning. Additionally, future research should attempt to further explore the connection between the problem space and the nature of learned visual processing. For example, it could be possible to construct a model which permits a notion of learnable monochromacy or dichromacy. This would make it possible to better understand the connection between problem complexity and the need for colour acuity. Finally, experimentation with networks trained on hyperspectral images, where a complete spectrum is collected for each pixel, may enable more fine-grained comparison with physiological data.

\bibliographystyle{apacite}
\bibliography{main}

\begin{thebibliography}{}

\bibitem [\protect \citeauthoryear {%
Adrian%
\ \BBA {} Matthews%
}{%
Adrian%
\ \BBA {} Matthews%
}{%
{\protect \APACyear {1927}}%
{\protect \APACexlab {{\protect \BCnt {1}}}}}]{%
AdrianMatthewsPt1}
\APACinsertmetastar {%
AdrianMatthewsPt1}%
\begin{APACrefauthors}%
Adrian, E\BPBI D.%
\BCBT {}\ \BBA {} Matthews, R.%
\end{APACrefauthors}%
\unskip\
\newblock
\APACrefYearMonthDay{1927{\protect \BCnt {1}}}{}{}.
\newblock
{\BBOQ}\APACrefatitle {The action of light on the eye} {The action of light on
  the eye}.{\BBCQ}
\newblock
\APACjournalVolNumPages{The Journal of Physiology}{63}{4}{378-414}.
\newblock
\begin{APACrefURL}
  \url{https://physoc.onlinelibrary.wiley.com/doi/abs/10.1113/jphysiol.1927.sp002410}
  \end{APACrefURL}
\newblock
\begin{APACrefDOI} \doi{10.1113/jphysiol.1927.sp002410} \end{APACrefDOI}
\PrintBackRefs{\CurrentBib}

\bibitem [\protect \citeauthoryear {%
Adrian%
\ \BBA {} Matthews%
}{%
Adrian%
\ \BBA {} Matthews%
}{%
{\protect \APACyear {1927}}%
{\protect \APACexlab {{\protect \BCnt {2}}}}}]{%
AdrianMatthewsPt2}
\APACinsertmetastar {%
AdrianMatthewsPt2}%
\begin{APACrefauthors}%
Adrian, E\BPBI D.%
\BCBT {}\ \BBA {} Matthews, R.%
\end{APACrefauthors}%
\unskip\
\newblock
\APACrefYearMonthDay{1927{\protect \BCnt {2}}}{}{}.
\newblock
{\BBOQ}\APACrefatitle {The action of light on the eye} {The action of light on
  the eye}.{\BBCQ}
\newblock
\APACjournalVolNumPages{The Journal of Physiology}{64}{3}{279-301}.
\newblock
\begin{APACrefURL}
  \url{https://physoc.onlinelibrary.wiley.com/doi/abs/10.1113/jphysiol.1927.sp002437}
  \end{APACrefURL}
\newblock
\begin{APACrefDOI} \doi{10.1113/jphysiol.1927.sp002437} \end{APACrefDOI}
\PrintBackRefs{\CurrentBib}

\bibitem [\protect \citeauthoryear {%
Adrian%
\ \BBA {} Matthews%
}{%
Adrian%
\ \BBA {} Matthews%
}{%
{\protect \APACyear {1928}}%
}]{%
AdrianMatthewsPt3}
\APACinsertmetastar {%
AdrianMatthewsPt3}%
\begin{APACrefauthors}%
Adrian, E\BPBI D.%
\BCBT {}\ \BBA {} Matthews, R.%
\end{APACrefauthors}%
\unskip\
\newblock
\APACrefYearMonthDay{1928}{}{}.
\newblock
{\BBOQ}\APACrefatitle {The action of light on the eye} {The action of light on
  the eye}.{\BBCQ}
\newblock
\APACjournalVolNumPages{The Journal of Physiology}{65}{3}{273-298}.
\newblock
\begin{APACrefURL}
  \url{https://physoc.onlinelibrary.wiley.com/doi/abs/10.1113/jphysiol.1928.sp002475}
  \end{APACrefURL}
\newblock
\begin{APACrefDOI} \doi{10.1113/jphysiol.1928.sp002475} \end{APACrefDOI}
\PrintBackRefs{\CurrentBib}

\bibitem [\protect \citeauthoryear {%
Barlow%
}{%
Barlow%
}{%
{\protect \APACyear {1953}}%
}]{%
doi:10.1113/jphysiol.1953.sp004829}
\APACinsertmetastar {%
doi:10.1113/jphysiol.1953.sp004829}%
\begin{APACrefauthors}%
Barlow, H\BPBI B.%
\end{APACrefauthors}%
\unskip\
\newblock
\APACrefYearMonthDay{1953}{}{}.
\newblock
{\BBOQ}\APACrefatitle {Summation and inhibition in the frog's retina}
  {Summation and inhibition in the frog's retina}.{\BBCQ}
\newblock
\APACjournalVolNumPages{The Journal of Physiology}{119}{1}{69-88}.
\newblock
\begin{APACrefURL}
  \url{https://physoc.onlinelibrary.wiley.com/doi/abs/10.1113/jphysiol.1953.sp004829}
  \end{APACrefURL}
\newblock
\begin{APACrefDOI} \doi{10.1113/jphysiol.1953.sp004829} \end{APACrefDOI}
\PrintBackRefs{\CurrentBib}

\bibitem [\protect \citeauthoryear {%
Bedford%
\ \BBA {} Wyszecki%
}{%
Bedford%
\ \BBA {} Wyszecki%
}{%
{\protect \APACyear {1958}}%
}]{%
bedford1958wavelength}
\APACinsertmetastar {%
bedford1958wavelength}%
\begin{APACrefauthors}%
Bedford, R.%
\BCBT {}\ \BBA {} Wyszecki, G\BPBI W.%
\end{APACrefauthors}%
\unskip\
\newblock
\APACrefYearMonthDay{1958}{}{}.
\newblock
{\BBOQ}\APACrefatitle {Wavelength discrimination for point sources} {Wavelength
  discrimination for point sources}.{\BBCQ}
\newblock
\APACjournalVolNumPages{JOSA}{48}{2}{129--135}.
\PrintBackRefs{\CurrentBib}

\bibitem [\protect \citeauthoryear {%
Bell%
\ \BBA {} Sejnowski%
}{%
Bell%
\ \BBA {} Sejnowski%
}{%
{\protect \APACyear {1997}}%
}]{%
BELL19973327}
\APACinsertmetastar {%
BELL19973327}%
\begin{APACrefauthors}%
Bell, A\BPBI J.%
\BCBT {}\ \BBA {} Sejnowski, T\BPBI J.%
\end{APACrefauthors}%
\unskip\
\newblock
\APACrefYearMonthDay{1997}{}{}.
\newblock
{\BBOQ}\APACrefatitle {The “independent components” of natural scenes are
  edge filters} {The “independent components” of natural scenes are edge
  filters}.{\BBCQ}
\newblock
\APACjournalVolNumPages{Vision Research}{37}{23}{3327 - 3338}.
\newblock
\begin{APACrefURL}
  \url{http://www.sciencedirect.com/science/article/pii/S0042698997001211}
  \end{APACrefURL}
\newblock
\begin{APACrefDOI} \doi{https://doi.org/10.1016/S0042-6989(97)00121-1}
  \end{APACrefDOI}
\PrintBackRefs{\CurrentBib}

\bibitem [\protect \citeauthoryear {%
Bilotta%
\ \BBA {} Abramov%
}{%
Bilotta%
\ \BBA {} Abramov%
}{%
{\protect \APACyear {1989}}%
}]{%
bilotta1989spatial}
\APACinsertmetastar {%
bilotta1989spatial}%
\begin{APACrefauthors}%
Bilotta, J.%
\BCBT {}\ \BBA {} Abramov, I.%
\end{APACrefauthors}%
\unskip\
\newblock
\APACrefYearMonthDay{1989}{}{}.
\newblock
{\BBOQ}\APACrefatitle {Spatial properties of goldfish ganglion cells.} {Spatial
  properties of goldfish ganglion cells.}{\BBCQ}
\newblock
\APACjournalVolNumPages{The Journal of general physiology}{93}{6}{1147--1169}.
\PrintBackRefs{\CurrentBib}

\bibitem [\protect \citeauthoryear {%
Bottou%
\ \protect \BOthers {.}}{%
Bottou%
\ \protect \BOthers {.}}{%
{\protect \APACyear {1994}}%
}]{%
bottou1994comparison}
\APACinsertmetastar {%
bottou1994comparison}%
\begin{APACrefauthors}%
Bottou, L.%
, Cortes, C.%
, Denker, J\BPBI S.%
, Drucker, H.%
, Guyon, I.%
, Jackel, L\BPBI D.%
\BDBL {}others%
\end{APACrefauthors}%
\unskip\
\newblock
\APACrefYearMonthDay{1994}{}{}.
\newblock
{\BBOQ}\APACrefatitle {Comparison of classifier methods: a case study in
  handwritten digit recognition} {Comparison of classifier methods: a case
  study in handwritten digit recognition}.{\BBCQ}
\newblock
\BIn{} \APACrefbtitle {International conference on pattern recognition}
  {International conference on pattern recognition}\ (\BPGS\ 77--77).
\PrintBackRefs{\CurrentBib}

\bibitem [\protect \citeauthoryear {%
Boynton%
}{%
Boynton%
}{%
{\protect \APACyear {2002}}%
}]{%
BOYNTON2002R838}
\APACinsertmetastar {%
BOYNTON2002R838}%
\begin{APACrefauthors}%
Boynton, G\BPBI M.%
\end{APACrefauthors}%
\unskip\
\newblock
\APACrefYearMonthDay{2002}{}{}.
\newblock
{\BBOQ}\APACrefatitle {Color Vision: How the Cortex Represents Color} {Color
  vision: How the cortex represents color}.{\BBCQ}
\newblock
\APACjournalVolNumPages{Current Biology}{12}{24}{R838 - R840}.
\newblock
\begin{APACrefURL}
  \url{http://www.sciencedirect.com/science/article/pii/S0960982202013477}
  \end{APACrefURL}
\newblock
\begin{APACrefDOI} \doi{https://doi.org/10.1016/S0960-9822(02)01347-7}
  \end{APACrefDOI}
\PrintBackRefs{\CurrentBib}

\bibitem [\protect \citeauthoryear {%
Cadena%
\ \protect \BOthers {.}}{%
Cadena%
\ \protect \BOthers {.}}{%
{\protect \APACyear {2017}}%
}]{%
Cadena201764}
\APACinsertmetastar {%
Cadena201764}%
\begin{APACrefauthors}%
Cadena, S\BPBI A.%
, Denfield, G\BPBI H.%
, Walker, E\BPBI Y.%
, Gatys, L\BPBI A.%
, Tolias, A\BPBI S.%
, Bethge, M.%
\BCBL {}\ \BBA {} Ecker, A\BPBI S.%
\end{APACrefauthors}%
\unskip\
\newblock
\APACrefYearMonthDay{2017}{}{}.
\newblock
{\BBOQ}\APACrefatitle {Deep convolutional models improve predictions of macaque
  V1 responses to natural images} {Deep convolutional models improve
  predictions of macaque v1 responses to natural images}.{\BBCQ}
\newblock
\APACjournalVolNumPages{bioRxiv}{}{}{}.
\newblock
\begin{APACrefURL}
  \url{https://www.biorxiv.org/content/early/2017/10/11/201764}
  \end{APACrefURL}
\newblock
\begin{APACrefDOI} \doi{10.1101/201764} \end{APACrefDOI}
\PrintBackRefs{\CurrentBib}

\bibitem [\protect \citeauthoryear {%
Dacey%
}{%
Dacey%
}{%
{\protect \APACyear {1993}}%
}]{%
dacey1993mosaic}
\APACinsertmetastar {%
dacey1993mosaic}%
\begin{APACrefauthors}%
Dacey, D\BPBI M.%
\end{APACrefauthors}%
\unskip\
\newblock
\APACrefYearMonthDay{1993}{}{}.
\newblock
{\BBOQ}\APACrefatitle {The mosaic of midget ganglion cells in the human retina}
  {The mosaic of midget ganglion cells in the human retina}.{\BBCQ}
\newblock
\APACjournalVolNumPages{Journal of Neuroscience}{13}{12}{5334--5355}.
\PrintBackRefs{\CurrentBib}

\bibitem [\protect \citeauthoryear {%
Dalal%
\ \BBA {} Triggs%
}{%
Dalal%
\ \BBA {} Triggs%
}{%
{\protect \APACyear {2005}}%
}]{%
dalal2005histograms}
\APACinsertmetastar {%
dalal2005histograms}%
\begin{APACrefauthors}%
Dalal, N.%
\BCBT {}\ \BBA {} Triggs, B.%
\end{APACrefauthors}%
\unskip\
\newblock
\APACrefYearMonthDay{2005}{}{}.
\newblock
{\BBOQ}\APACrefatitle {Histograms of oriented gradients for human detection}
  {Histograms of oriented gradients for human detection}.{\BBCQ}
\newblock
\BIn{} \APACrefbtitle {2005 IEEE computer society conference on computer vision
  and pattern recognition (CVPR'05)} {2005 ieee computer society conference on
  computer vision and pattern recognition (cvpr'05)}\ (\BVOL~1, \BPGS\
  886--893).
\PrintBackRefs{\CurrentBib}

\bibitem [\protect \citeauthoryear {%
Daw%
}{%
Daw%
}{%
{\protect \APACyear {1967}}%
}]{%
daw1967goldfish}
\APACinsertmetastar {%
daw1967goldfish}%
\begin{APACrefauthors}%
Daw, N\BPBI W.%
\end{APACrefauthors}%
\unskip\
\newblock
\APACrefYearMonthDay{1967}{}{}.
\newblock
{\BBOQ}\APACrefatitle {Goldfish retina: organization for simultaneous color
  contrast} {Goldfish retina: organization for simultaneous color
  contrast}.{\BBCQ}
\newblock
\APACjournalVolNumPages{Science}{158}{3803}{942--944}.
\PrintBackRefs{\CurrentBib}

\bibitem [\protect \citeauthoryear {%
{De Valois}%
, Abramov%
\BCBL {}\ \BBA {} {Jacobs}%
}{%
{De Valois}%
\ \protect \BOthers {.}}{%
{\protect \APACyear {1966}}%
}]{%
DeValois:66}
\APACinsertmetastar {%
DeValois:66}%
\begin{APACrefauthors}%
{De Valois}, R\BPBI L.%
, Abramov, I.%
\BCBL {}\ \BBA {} {Jacobs}, G\BPBI H.%
\end{APACrefauthors}%
\unskip\
\newblock
\APACrefYearMonthDay{1966}{Jul}{}.
\newblock
{\BBOQ}\APACrefatitle {Analysis of Response Patterns of LGN Cells$\ast$}
  {Analysis of response patterns of lgn cells$\ast$}.{\BBCQ}
\newblock
\APACjournalVolNumPages{J. Opt. Soc. Am.}{56}{7}{966--977}.
\newblock
\begin{APACrefURL}
  \url{http://www.osapublishing.org/abstract.cfm?URI=josa-56-7-966}
  \end{APACrefURL}
\newblock
\begin{APACrefDOI} \doi{10.1364/JOSA.56.000966} \end{APACrefDOI}
\PrintBackRefs{\CurrentBib}

\bibitem [\protect \citeauthoryear {%
{De Valois}%
, Albrecht%
\BCBL {}\ \BBA {} Thorell%
}{%
{De Valois}%
\ \protect \BOthers {.}}{%
{\protect \APACyear {1982}}%
}]{%
de1982spatial}
\APACinsertmetastar {%
de1982spatial}%
\begin{APACrefauthors}%
{De Valois}, R\BPBI L.%
, Albrecht, D\BPBI G.%
\BCBL {}\ \BBA {} Thorell, L\BPBI G.%
\end{APACrefauthors}%
\unskip\
\newblock
\APACrefYearMonthDay{1982}{}{}.
\newblock
{\BBOQ}\APACrefatitle {Spatial frequency selectivity of cells in macaque visual
  cortex} {Spatial frequency selectivity of cells in macaque visual
  cortex}.{\BBCQ}
\newblock
\APACjournalVolNumPages{Vision research}{22}{5}{545--559}.
\PrintBackRefs{\CurrentBib}

\bibitem [\protect \citeauthoryear {%
{De Valois}%
, Smith%
, Kitai%
\BCBL {}\ \BBA {} Karoly%
}{%
{De Valois}%
, Smith%
, Kitai%
\BCBL {}\ \BBA {} Karoly%
}{%
{\protect \APACyear {1958}}%
}]{%
de1958response}
\APACinsertmetastar {%
de1958response}%
\begin{APACrefauthors}%
{De Valois}, R\BPBI L.%
, Smith, C.%
, Kitai, S\BPBI T.%
\BCBL {}\ \BBA {} Karoly, A.%
\end{APACrefauthors}%
\unskip\
\newblock
\APACrefYearMonthDay{1958}{}{}.
\newblock
{\BBOQ}\APACrefatitle {Response of single cells in monkey lateral geniculate
  nucleus to monochromatic light.} {Response of single cells in monkey lateral
  geniculate nucleus to monochromatic light.}{\BBCQ}
\newblock
\APACjournalVolNumPages{Science}{}{}{}.
\PrintBackRefs{\CurrentBib}

\bibitem [\protect \citeauthoryear {%
{De Valois}%
, Smith%
, Karoly%
\BCBL {}\ \BBA {} Kitai%
}{%
{De Valois}%
, Smith%
, Karoly%
\BCBL {}\ \BBA {} Kitai%
}{%
{\protect \APACyear {1958}}%
}]{%
de1958electrical}
\APACinsertmetastar {%
de1958electrical}%
\begin{APACrefauthors}%
{De Valois}, R\BPBI L.%
, Smith, C\BPBI J.%
, Karoly, A.%
\BCBL {}\ \BBA {} Kitai, S.%
\end{APACrefauthors}%
\unskip\
\newblock
\APACrefYearMonthDay{1958}{}{}.
\newblock
{\BBOQ}\APACrefatitle {Electrical responses of primate visual system: I.
  Different layers of macaque lateral geniculate nucleus.} {Electrical
  responses of primate visual system: I. different layers of macaque lateral
  geniculate nucleus.}{\BBCQ}
\newblock
\APACjournalVolNumPages{Journal of comparative and physiological
  psychology}{51}{6}{662}.
\PrintBackRefs{\CurrentBib}

\bibitem [\protect \citeauthoryear {%
Derrington%
, Krauskopf%
\BCBL {}\ \BBA {} Lennie%
}{%
Derrington%
\ \protect \BOthers {.}}{%
{\protect \APACyear {1984}}%
}]{%
derrington1984chromatic}
\APACinsertmetastar {%
derrington1984chromatic}%
\begin{APACrefauthors}%
Derrington, A\BPBI M.%
, Krauskopf, J.%
\BCBL {}\ \BBA {} Lennie, P.%
\end{APACrefauthors}%
\unskip\
\newblock
\APACrefYearMonthDay{1984}{}{}.
\newblock
{\BBOQ}\APACrefatitle {Chromatic mechanisms in lateral geniculate nucleus of
  macaque.} {Chromatic mechanisms in lateral geniculate nucleus of
  macaque.}{\BBCQ}
\newblock
\APACjournalVolNumPages{The Journal of physiology}{357}{1}{241--265}.
\PrintBackRefs{\CurrentBib}

\bibitem [\protect \citeauthoryear {%
Du%
, Farrahi%
\BCBL {}\ \BBA {} Niranjan%
}{%
Du%
\ \protect \BOthers {.}}{%
{\protect \APACyear {2019}}%
}]{%
du2019transfer}
\APACinsertmetastar {%
du2019transfer}%
\begin{APACrefauthors}%
Du, X.%
, Farrahi, K.%
\BCBL {}\ \BBA {} Niranjan, M.%
\end{APACrefauthors}%
\unskip\
\newblock
\APACrefYearMonthDay{2019}{}{}.
\newblock
{\BBOQ}\APACrefatitle {Transfer learning across human activities using a
  cascade neural network architecture} {Transfer learning across human
  activities using a cascade neural network architecture}.{\BBCQ}
\newblock
\BIn{} \APACrefbtitle {Proceedings of the 23rd International Symposium on
  Wearable Computers} {Proceedings of the 23rd international symposium on
  wearable computers}\ (\BPGS\ 35--44).
\PrintBackRefs{\CurrentBib}

\bibitem [\protect \citeauthoryear {%
Engel%
, Zhang%
\BCBL {}\ \BBA {} Wandell%
}{%
Engel%
\ \protect \BOthers {.}}{%
{\protect \APACyear {1997}}%
}]{%
Engel199768}
\APACinsertmetastar {%
Engel199768}%
\begin{APACrefauthors}%
Engel, S.%
, Zhang, X.%
\BCBL {}\ \BBA {} Wandell, B.%
\end{APACrefauthors}%
\unskip\
\newblock
\APACrefYearMonthDay{1997}{}{}.
\newblock
{\BBOQ}\APACrefatitle {Colour tuning in human visual cortex measured with
  functional magnetic resonance imaging} {Colour tuning in human visual cortex
  measured with functional magnetic resonance imaging}.{\BBCQ}
\newblock
\APACjournalVolNumPages{Nature}{388}{6637}{68-71}.
\newblock
\begin{APACrefURL}
  \url{https://www.scopus.com/inward/record.uri?eid=2-s2.0-0030744384&doi=10.1038%2f40398&partnerID=40&md5=d42786f821eb1652af0ad297996452d6}
  \end{APACrefURL}
\newblock
\APACrefnote{cited By 266}
\newblock
\begin{APACrefDOI} \doi{10.1038/40398} \end{APACrefDOI}
\PrintBackRefs{\CurrentBib}

\bibitem [\protect \citeauthoryear {%
Freeman%
\ \BBA {} Roth%
}{%
Freeman%
\ \BBA {} Roth%
}{%
{\protect \APACyear {1995}}%
}]{%
freeman1995orientation}
\APACinsertmetastar {%
freeman1995orientation}%
\begin{APACrefauthors}%
Freeman, W\BPBI T.%
\BCBT {}\ \BBA {} Roth, M.%
\end{APACrefauthors}%
\unskip\
\newblock
\APACrefYearMonthDay{1995}{}{}.
\newblock
{\BBOQ}\APACrefatitle {Orientation histograms for hand gesture recognition}
  {Orientation histograms for hand gesture recognition}.{\BBCQ}
\newblock
\BIn{} \APACrefbtitle {International workshop on automatic face and gesture
  recognition} {International workshop on automatic face and gesture
  recognition}\ (\BVOL~12, \BPGS\ 296--301).
\PrintBackRefs{\CurrentBib}

\bibitem [\protect \citeauthoryear {%
Ghodrati%
, Khaligh-Razavi%
\BCBL {}\ \BBA {} Lehky%
}{%
Ghodrati%
\ \protect \BOthers {.}}{%
{\protect \APACyear {2017}}%
}]{%
ghodrati2017towards}
\APACinsertmetastar {%
ghodrati2017towards}%
\begin{APACrefauthors}%
Ghodrati, M.%
, Khaligh-Razavi, S\BHBI M.%
\BCBL {}\ \BBA {} Lehky, S\BPBI R.%
\end{APACrefauthors}%
\unskip\
\newblock
\APACrefYearMonthDay{2017}{}{}.
\newblock
{\BBOQ}\APACrefatitle {Towards building a more complex view of the lateral
  geniculate nucleus: recent advances in understanding its role} {Towards
  building a more complex view of the lateral geniculate nucleus: recent
  advances in understanding its role}.{\BBCQ}
\newblock
\APACjournalVolNumPages{Progress in Neurobiology}{156}{}{214--255}.
\PrintBackRefs{\CurrentBib}

\bibitem [\protect \citeauthoryear {%
Glorot%
\ \BBA {} Bengio%
}{%
Glorot%
\ \BBA {} Bengio%
}{%
{\protect \APACyear {2010}}%
}]{%
glorot2010understanding}
\APACinsertmetastar {%
glorot2010understanding}%
\begin{APACrefauthors}%
Glorot, X.%
\BCBT {}\ \BBA {} Bengio, Y.%
\end{APACrefauthors}%
\unskip\
\newblock
\APACrefYearMonthDay{2010}{}{}.
\newblock
{\BBOQ}\APACrefatitle {Understanding the difficulty of training deep
  feedforward neural networks} {Understanding the difficulty of training deep
  feedforward neural networks}.{\BBCQ}
\newblock
\BIn{} \APACrefbtitle {Proceedings of the thirteenth international conference
  on artificial intelligence and statistics} {Proceedings of the thirteenth
  international conference on artificial intelligence and statistics}\ (\BPGS\
  249--256).
\PrintBackRefs{\CurrentBib}

\bibitem [\protect \citeauthoryear {%
Goethe%
}{%
Goethe%
}{%
{\protect \APACyear {1840}}%
}]{%
goethe1840theory}
\APACinsertmetastar {%
goethe1840theory}%
\begin{APACrefauthors}%
Goethe, J\BPBI W\BPBI v.%
\end{APACrefauthors}%
\unskip\
\newblock
\APACrefYear{1840}.
\newblock
\APACrefbtitle {Theory of colours} {Theory of colours}\ (\BVOL~3).
\newblock
\APACaddressPublisher{}{MIT Press}.
\PrintBackRefs{\CurrentBib}

\bibitem [\protect \citeauthoryear {%
Gomez-Villa%
, Martin%
, Vazquez-Corral%
\BCBL {}\ \BBA {} Bertalmio%
}{%
Gomez-Villa%
\ \protect \BOthers {.}}{%
{\protect \APACyear {2019}}%
}]{%
gomez2018convolutional}
\APACinsertmetastar {%
gomez2018convolutional}%
\begin{APACrefauthors}%
Gomez-Villa, A.%
, Martin, A.%
, Vazquez-Corral, J.%
\BCBL {}\ \BBA {} Bertalmio, M.%
\end{APACrefauthors}%
\unskip\
\newblock
\APACrefYearMonthDay{2019}{June}{}.
\newblock
{\BBOQ}\APACrefatitle {{Convolutional Neural Networks Can Be Deceived by Visual
  Illusions}} {{Convolutional Neural Networks Can Be Deceived by Visual
  Illusions}}.{\BBCQ}
\newblock
\BIn{} \APACrefbtitle {{The IEEE Conference on Computer Vision and Pattern
  Recognition (CVPR)}.} {{The IEEE Conference on Computer Vision and Pattern
  Recognition (CVPR)}.}
\PrintBackRefs{\CurrentBib}

\bibitem [\protect \citeauthoryear {%
G{\"u}{\c{c}}l{\"u}%
\ \BBA {} van Gerven%
}{%
G{\"u}{\c{c}}l{\"u}%
\ \BBA {} van Gerven%
}{%
{\protect \APACyear {2015}}%
}]{%
gucclu2015deep}
\APACinsertmetastar {%
gucclu2015deep}%
\begin{APACrefauthors}%
G{\"u}{\c{c}}l{\"u}, U.%
\BCBT {}\ \BBA {} van Gerven, M\BPBI A.%
\end{APACrefauthors}%
\unskip\
\newblock
\APACrefYearMonthDay{2015}{}{}.
\newblock
{\BBOQ}\APACrefatitle {Deep neural networks reveal a gradient in the complexity
  of neural representations across the ventral stream} {Deep neural networks
  reveal a gradient in the complexity of neural representations across the
  ventral stream}.{\BBCQ}
\newblock
\APACjournalVolNumPages{Journal of Neuroscience}{35}{27}{10005--10014}.
\PrintBackRefs{\CurrentBib}

\bibitem [\protect \citeauthoryear {%
Hartline%
}{%
Hartline%
}{%
{\protect \APACyear {1938}}%
}]{%
doi:10.1152/ajplegacy.1938.121.2.400}
\APACinsertmetastar {%
doi:10.1152/ajplegacy.1938.121.2.400}%
\begin{APACrefauthors}%
Hartline, H\BPBI K.%
\end{APACrefauthors}%
\unskip\
\newblock
\APACrefYearMonthDay{1938}{}{}.
\newblock
{\BBOQ}\APACrefatitle {THE RESPONSE OF SINGLE OPTIC NERVE FIBERS OF THE
  VERTEBRATE EYE TO ILLUMINATION OF THE RETINA} {The response of single optic
  nerve fibers of the vertebrate eye to illumination of the retina}.{\BBCQ}
\newblock
\APACjournalVolNumPages{American Journal of Physiology-Legacy
  Content}{121}{2}{400-415}.
\newblock
\begin{APACrefDOI} \doi{10.1152/ajplegacy.1938.121.2.400} \end{APACrefDOI}
\PrintBackRefs{\CurrentBib}

\bibitem [\protect \citeauthoryear {%
Hartline%
}{%
Hartline%
}{%
{\protect \APACyear {1940}}%
}]{%
Hartline:40}
\APACinsertmetastar {%
Hartline:40}%
\begin{APACrefauthors}%
Hartline, H\BPBI K.%
\end{APACrefauthors}%
\unskip\
\newblock
\APACrefYearMonthDay{1940}{Jun}{}.
\newblock
{\BBOQ}\APACrefatitle {The Nerve Messages in the Fibers of the Visual
  Pathway$\ast$} {The nerve messages in the fibers of the visual
  pathway$\ast$}.{\BBCQ}
\newblock
\APACjournalVolNumPages{J. Opt. Soc. Am.}{30}{6}{239--247}.
\newblock
\begin{APACrefURL}
  \url{http://www.osapublishing.org/abstract.cfm?URI=josa-30-6-239}
  \end{APACrefURL}
\newblock
\begin{APACrefDOI} \doi{10.1364/JOSA.30.000239} \end{APACrefDOI}
\PrintBackRefs{\CurrentBib}

\bibitem [\protect \citeauthoryear {%
Hartline%
, Wagner%
\BCBL {}\ \BBA {} Macnichol%
}{%
Hartline%
\ \protect \BOthers {.}}{%
{\protect \APACyear {1952}}%
}]{%
Hartline1952ThePO}
\APACinsertmetastar {%
Hartline1952ThePO}%
\begin{APACrefauthors}%
Hartline, H\BPBI K.%
, Wagner, H\BPBI G.%
\BCBL {}\ \BBA {} Macnichol, E\BPBI F.%
\end{APACrefauthors}%
\unskip\
\newblock
\APACrefYearMonthDay{1952}{}{}.
\newblock
{\BBOQ}\APACrefatitle {The peripheral origin of nervous activity in the visual
  system.} {The peripheral origin of nervous activity in the visual
  system.}{\BBCQ}
\newblock
\APACjournalVolNumPages{Cold Spring Harbor symposia on quantitative
  biology}{17}{}{125-41}.
\PrintBackRefs{\CurrentBib}

\bibitem [\protect \citeauthoryear {%
He%
, Zhang%
, Ren%
\BCBL {}\ \BBA {} Sun%
}{%
He%
\ \protect \BOthers {.}}{%
{\protect \APACyear {2016}}%
}]{%
he2016deep}
\APACinsertmetastar {%
he2016deep}%
\begin{APACrefauthors}%
He, K.%
, Zhang, X.%
, Ren, S.%
\BCBL {}\ \BBA {} Sun, J.%
\end{APACrefauthors}%
\unskip\
\newblock
\APACrefYearMonthDay{2016}{}{}.
\newblock
{\BBOQ}\APACrefatitle {Deep residual learning for image recognition} {Deep
  residual learning for image recognition}.{\BBCQ}
\newblock
\BIn{} \APACrefbtitle {Proceedings of the IEEE conference on computer vision
  and pattern recognition} {Proceedings of the ieee conference on computer
  vision and pattern recognition}\ (\BPGS\ 770--778).
\PrintBackRefs{\CurrentBib}

\bibitem [\protect \citeauthoryear {%
Helmholtz%
}{%
Helmholtz%
}{%
{\protect \APACyear {1852}}%
}]{%
helmholtz1852lxxxi}
\APACinsertmetastar {%
helmholtz1852lxxxi}%
\begin{APACrefauthors}%
Helmholtz, H\BPBI v.%
\end{APACrefauthors}%
\unskip\
\newblock
\APACrefYearMonthDay{1852}{}{}.
\newblock
{\BBOQ}\APACrefatitle {{LXXXI. On the theory of compound colours}} {{LXXXI. On
  the theory of compound colours}}.{\BBCQ}
\newblock
\APACjournalVolNumPages{The London, Edinburgh, and Dublin Philosophical
  Magazine and Journal of Science}{4}{28}{519--534}.
\PrintBackRefs{\CurrentBib}

\bibitem [\protect \citeauthoryear {%
H{\'e}naff%
\ \protect \BOthers {.}}{%
H{\'e}naff%
\ \protect \BOthers {.}}{%
{\protect \APACyear {2019}}%
}]{%
henaff2019data}
\APACinsertmetastar {%
henaff2019data}%
\begin{APACrefauthors}%
H{\'e}naff, O\BPBI J.%
, Srinivas, A.%
, De~Fauw, J.%
, Razavi, A.%
, Doersch, C.%
, Eslami, S.%
\BCBL {}\ \BBA {} Oord, A\BPBI v\BPBI d.%
\end{APACrefauthors}%
\unskip\
\newblock
\APACrefYearMonthDay{2019}{}{}.
\newblock
{\BBOQ}\APACrefatitle {Data-efficient image recognition with contrastive
  predictive coding} {Data-efficient image recognition with contrastive
  predictive coding}.{\BBCQ}
\newblock
\APACjournalVolNumPages{arXiv preprint arXiv:1905.09272}{}{}{}.
\PrintBackRefs{\CurrentBib}

\bibitem [\protect \citeauthoryear {%
Hering%
}{%
Hering%
}{%
{\protect \APACyear {1920}}%
}]{%
hering1920grundzuge}
\APACinsertmetastar {%
hering1920grundzuge}%
\begin{APACrefauthors}%
Hering, E.%
\end{APACrefauthors}%
\unskip\
\newblock
\APACrefYear{1920}.
\newblock
\APACrefbtitle {Grundz{\"u}ge der Lehre vom Lichtsinn} {Grundz{\"u}ge der lehre
  vom lichtsinn}.
\newblock
\APACaddressPublisher{}{Springer}.
\PrintBackRefs{\CurrentBib}

\bibitem [\protect \citeauthoryear {%
Hjelm%
\ \protect \BOthers {.}}{%
Hjelm%
\ \protect \BOthers {.}}{%
{\protect \APACyear {2018}}%
}]{%
hjelm2018learning}
\APACinsertmetastar {%
hjelm2018learning}%
\begin{APACrefauthors}%
Hjelm, R\BPBI D.%
, Fedorov, A.%
, Lavoie-Marchildon, S.%
, Grewal, K.%
, Bachman, P.%
, Trischler, A.%
\BCBL {}\ \BBA {} Bengio, Y.%
\end{APACrefauthors}%
\unskip\
\newblock
\APACrefYearMonthDay{2018}{}{}.
\newblock
{\BBOQ}\APACrefatitle {Learning deep representations by mutual information
  estimation and maximization} {Learning deep representations by mutual
  information estimation and maximization}.{\BBCQ}
\newblock
\APACjournalVolNumPages{arXiv preprint arXiv:1808.06670}{}{}{}.
\PrintBackRefs{\CurrentBib}

\bibitem [\protect \citeauthoryear {%
Huang%
, Liu%
, Van Der~Maaten%
\BCBL {}\ \BBA {} Weinberger%
}{%
Huang%
\ \protect \BOthers {.}}{%
{\protect \APACyear {2017}}%
}]{%
huang2017densely}
\APACinsertmetastar {%
huang2017densely}%
\begin{APACrefauthors}%
Huang, G.%
, Liu, Z.%
, Van Der~Maaten, L.%
\BCBL {}\ \BBA {} Weinberger, K\BPBI Q.%
\end{APACrefauthors}%
\unskip\
\newblock
\APACrefYearMonthDay{2017}{}{}.
\newblock
{\BBOQ}\APACrefatitle {Densely connected convolutional networks} {Densely
  connected convolutional networks}.{\BBCQ}
\newblock
\BIn{} \APACrefbtitle {Proceedings of the IEEE conference on computer vision
  and pattern recognition} {Proceedings of the ieee conference on computer
  vision and pattern recognition}\ (\BPGS\ 4700--4708).
\PrintBackRefs{\CurrentBib}

\bibitem [\protect \citeauthoryear {%
Huang%
, Sun%
, Liu%
, Sedra%
\BCBL {}\ \BBA {} Weinberger%
}{%
Huang%
\ \protect \BOthers {.}}{%
{\protect \APACyear {2016}}%
}]{%
huang2016deep}
\APACinsertmetastar {%
huang2016deep}%
\begin{APACrefauthors}%
Huang, G.%
, Sun, Y.%
, Liu, Z.%
, Sedra, D.%
\BCBL {}\ \BBA {} Weinberger, K\BPBI Q.%
\end{APACrefauthors}%
\unskip\
\newblock
\APACrefYearMonthDay{2016}{}{}.
\newblock
{\BBOQ}\APACrefatitle {Deep networks with stochastic depth} {Deep networks with
  stochastic depth}.{\BBCQ}
\newblock
\BIn{} \APACrefbtitle {European conference on computer vision} {European
  conference on computer vision}\ (\BPGS\ 646--661).
\PrintBackRefs{\CurrentBib}

\bibitem [\protect \citeauthoryear {%
Hubel%
\ \BBA {} Wiesel%
}{%
Hubel%
\ \BBA {} Wiesel%
}{%
{\protect \APACyear {1962}}%
}]{%
hubel1962receptive}
\APACinsertmetastar {%
hubel1962receptive}%
\begin{APACrefauthors}%
Hubel, D\BPBI H.%
\BCBT {}\ \BBA {} Wiesel, T\BPBI N.%
\end{APACrefauthors}%
\unskip\
\newblock
\APACrefYearMonthDay{1962}{}{}.
\newblock
{\BBOQ}\APACrefatitle {Receptive fields, binocular interaction and functional
  architecture in the cat's visual cortex} {Receptive fields, binocular
  interaction and functional architecture in the cat's visual cortex}.{\BBCQ}
\newblock
\APACjournalVolNumPages{The Journal of physiology}{160}{1}{106--154}.
\PrintBackRefs{\CurrentBib}

\bibitem [\protect \citeauthoryear {%
Hubel%
\ \BBA {} Wiesel%
}{%
Hubel%
\ \BBA {} Wiesel%
}{%
{\protect \APACyear {2004}}%
}]{%
hubel2004brain}
\APACinsertmetastar {%
hubel2004brain}%
\begin{APACrefauthors}%
Hubel, D\BPBI H.%
\BCBT {}\ \BBA {} Wiesel, T\BPBI N.%
\end{APACrefauthors}%
\unskip\
\newblock
\APACrefYear{2004}.
\newblock
\APACrefbtitle {Brain and visual perception: the story of a 25-year
  collaboration} {Brain and visual perception: the story of a 25-year
  collaboration}.
\newblock
\APACaddressPublisher{}{Oxford University Press}.
\PrintBackRefs{\CurrentBib}

\bibitem [\protect \citeauthoryear {%
Intel%
}{%
Intel%
}{%
{\protect \APACyear {2018}}%
}]{%
intel2018}
\APACinsertmetastar {%
intel2018}%
\begin{APACrefauthors}%
Intel.%
\end{APACrefauthors}%
\unskip\
\newblock
\APACrefYearMonthDay{2018}{}{}.
\newblock
\APACrefbtitle {Intel Scene Classification Challenge.} {Intel scene
  classification challenge.}
\newblock
\begin{APACrefURL}
  \url{https://datahack.analyticsvidhya.com/contest/practice-problem-intel-scene-classification-challe/}
  \end{APACrefURL}
\PrintBackRefs{\CurrentBib}

\bibitem [\protect \citeauthoryear {%
Jacobs%
}{%
Jacobs%
}{%
{\protect \APACyear {1964}}%
}]{%
jacobs1964single}
\APACinsertmetastar {%
jacobs1964single}%
\begin{APACrefauthors}%
Jacobs, G\BPBI H.%
\end{APACrefauthors}%
\unskip\
\newblock
\APACrefYearMonthDay{1964}{}{}.
\newblock
{\BBOQ}\APACrefatitle {Single cells in squirrel monkey lateral geniculate
  nucleus with broad spectral sensitivity} {Single cells in squirrel monkey
  lateral geniculate nucleus with broad spectral sensitivity}.{\BBCQ}
\newblock
\APACjournalVolNumPages{Vision Research}{4}{3-4}{221--IN3}.
\PrintBackRefs{\CurrentBib}

\bibitem [\protect \citeauthoryear {%
Johnson%
, Hawken%
\BCBL {}\ \BBA {} Shapley%
}{%
Johnson%
\ \protect \BOthers {.}}{%
{\protect \APACyear {2001}}%
}]{%
johnson2001spatial}
\APACinsertmetastar {%
johnson2001spatial}%
\begin{APACrefauthors}%
Johnson, E\BPBI N.%
, Hawken, M\BPBI J.%
\BCBL {}\ \BBA {} Shapley, R.%
\end{APACrefauthors}%
\unskip\
\newblock
\APACrefYearMonthDay{2001}{}{}.
\newblock
{\BBOQ}\APACrefatitle {The spatial transformation of color in the primary
  visual cortex of the macaque monkey} {The spatial transformation of color in
  the primary visual cortex of the macaque monkey}.{\BBCQ}
\newblock
\APACjournalVolNumPages{Nature neuroscience}{4}{4}{409}.
\PrintBackRefs{\CurrentBib}

\bibitem [\protect \citeauthoryear {%
Johnson%
, Hawken%
\BCBL {}\ \BBA {} Shapley%
}{%
Johnson%
\ \protect \BOthers {.}}{%
{\protect \APACyear {2008}}%
}]{%
johnson2008orientation}
\APACinsertmetastar {%
johnson2008orientation}%
\begin{APACrefauthors}%
Johnson, E\BPBI N.%
, Hawken, M\BPBI J.%
\BCBL {}\ \BBA {} Shapley, R.%
\end{APACrefauthors}%
\unskip\
\newblock
\APACrefYearMonthDay{2008}{}{}.
\newblock
{\BBOQ}\APACrefatitle {The orientation selectivity of color-responsive neurons
  in macaque V1} {The orientation selectivity of color-responsive neurons in
  macaque v1}.{\BBCQ}
\newblock
\APACjournalVolNumPages{Journal of Neuroscience}{28}{32}{8096--8106}.
\PrintBackRefs{\CurrentBib}

\bibitem [\protect \citeauthoryear {%
Kanan%
}{%
Kanan%
}{%
{\protect \APACyear {2013}}%
}]{%
kanan2013recognizing}
\APACinsertmetastar {%
kanan2013recognizing}%
\begin{APACrefauthors}%
Kanan, C.%
\end{APACrefauthors}%
\unskip\
\newblock
\APACrefYearMonthDay{2013}{}{}.
\newblock
{\BBOQ}\APACrefatitle {Recognizing sights, smells, and sounds with gnostic
  fields} {Recognizing sights, smells, and sounds with gnostic fields}.{\BBCQ}
\newblock
\APACjournalVolNumPages{PloS one}{8}{1}{}.
\PrintBackRefs{\CurrentBib}

\bibitem [\protect \citeauthoryear {%
Kanan%
}{%
Kanan%
}{%
{\protect \APACyear {2014}}%
}]{%
kanan2014fine}
\APACinsertmetastar {%
kanan2014fine}%
\begin{APACrefauthors}%
Kanan, C.%
\end{APACrefauthors}%
\unskip\
\newblock
\APACrefYearMonthDay{2014}{}{}.
\newblock
{\BBOQ}\APACrefatitle {Fine-grained object recognition with gnostic fields}
  {Fine-grained object recognition with gnostic fields}.{\BBCQ}
\newblock
\BIn{} \APACrefbtitle {IEEE Winter Conference on Applications of Computer
  Vision} {Ieee winter conference on applications of computer vision}\ (\BPGS\
  23--30).
\PrintBackRefs{\CurrentBib}

\bibitem [\protect \citeauthoryear {%
Karklin%
\ \BBA {} Lewicki%
}{%
Karklin%
\ \BBA {} Lewicki%
}{%
{\protect \APACyear {2003}}%
}]{%
karklin2003learning}
\APACinsertmetastar {%
karklin2003learning}%
\begin{APACrefauthors}%
Karklin, Y.%
\BCBT {}\ \BBA {} Lewicki, M\BPBI S.%
\end{APACrefauthors}%
\unskip\
\newblock
\APACrefYearMonthDay{2003}{}{}.
\newblock
{\BBOQ}\APACrefatitle {Learning higher-order structures in natural images}
  {Learning higher-order structures in natural images}.{\BBCQ}
\newblock
\APACjournalVolNumPages{Network: Computation in Neural
  Systems}{14}{3}{483--499}.
\PrintBackRefs{\CurrentBib}

\bibitem [\protect \citeauthoryear {%
Kleinschmidt%
, Lee%
, Requardt%
\BCBL {}\ \BBA {} Frahm%
}{%
Kleinschmidt%
\ \protect \BOthers {.}}{%
{\protect \APACyear {1996}}%
}]{%
kleinschmidt1996functional}
\APACinsertmetastar {%
kleinschmidt1996functional}%
\begin{APACrefauthors}%
Kleinschmidt, A.%
, Lee, B\BPBI B.%
, Requardt, M.%
\BCBL {}\ \BBA {} Frahm, J.%
\end{APACrefauthors}%
\unskip\
\newblock
\APACrefYearMonthDay{1996}{}{}.
\newblock
{\BBOQ}\APACrefatitle {Functional mapping of color processing by magnetic
  resonance imaging of responses to selective P-and M-pathway stimulation}
  {Functional mapping of color processing by magnetic resonance imaging of
  responses to selective p-and m-pathway stimulation}.{\BBCQ}
\newblock
\APACjournalVolNumPages{Experimental Brain Research}{110}{2}{279--288}.
\PrintBackRefs{\CurrentBib}

\bibitem [\protect \citeauthoryear {%
Krizhevsky%
}{%
Krizhevsky%
}{%
{\protect \APACyear {2009}}%
}]{%
Krizhevsky09learningmultiple}
\APACinsertmetastar {%
Krizhevsky09learningmultiple}%
\begin{APACrefauthors}%
Krizhevsky, A.%
\end{APACrefauthors}%
\unskip\
\newblock
\APACrefYearMonthDay{2009}{}{}.
\newblock
\APACrefbtitle {Learning multiple layers of features from tiny images}
  {Learning multiple layers of features from tiny images}\
  \APACbVolEdTR{}{\BTR{}}.
\PrintBackRefs{\CurrentBib}

\bibitem [\protect \citeauthoryear {%
Krizhevsky%
, Sutskever%
\BCBL {}\ \BBA {} Hinton%
}{%
Krizhevsky%
\ \protect \BOthers {.}}{%
{\protect \APACyear {2012}}%
}]{%
krizhevsky2012imagenet}
\APACinsertmetastar {%
krizhevsky2012imagenet}%
\begin{APACrefauthors}%
Krizhevsky, A.%
, Sutskever, I.%
\BCBL {}\ \BBA {} Hinton, G\BPBI E.%
\end{APACrefauthors}%
\unskip\
\newblock
\APACrefYearMonthDay{2012}{}{}.
\newblock
{\BBOQ}\APACrefatitle {Imagenet classification with deep convolutional neural
  networks} {Imagenet classification with deep convolutional neural
  networks}.{\BBCQ}
\newblock
\BIn{} \APACrefbtitle {Advances in neural information processing systems}
  {Advances in neural information processing systems}\ (\BPGS\ 1097--1105).
\PrintBackRefs{\CurrentBib}

\bibitem [\protect \citeauthoryear {%
Kuffler%
}{%
Kuffler%
}{%
{\protect \APACyear {1953}}%
}]{%
kuffler1953discharge}
\APACinsertmetastar {%
kuffler1953discharge}%
\begin{APACrefauthors}%
Kuffler, S\BPBI W.%
\end{APACrefauthors}%
\unskip\
\newblock
\APACrefYearMonthDay{1953}{}{}.
\newblock
{\BBOQ}\APACrefatitle {Discharge patterns and functional organization of
  mammalian retina} {Discharge patterns and functional organization of
  mammalian retina}.{\BBCQ}
\newblock
\APACjournalVolNumPages{Journal of neurophysiology}{16}{1}{37--68}.
\PrintBackRefs{\CurrentBib}

\bibitem [\protect \citeauthoryear {%
Le%
, Karpenko%
, Ngiam%
\BCBL {}\ \BBA {} Ng%
}{%
Le%
\ \protect \BOthers {.}}{%
{\protect \APACyear {2011}}%
}]{%
le2011ica}
\APACinsertmetastar {%
le2011ica}%
\begin{APACrefauthors}%
Le, Q\BPBI V.%
, Karpenko, A.%
, Ngiam, J.%
\BCBL {}\ \BBA {} Ng, A\BPBI Y.%
\end{APACrefauthors}%
\unskip\
\newblock
\APACrefYearMonthDay{2011}{}{}.
\newblock
{\BBOQ}\APACrefatitle {ICA with reconstruction cost for efficient overcomplete
  feature learning} {Ica with reconstruction cost for efficient overcomplete
  feature learning}.{\BBCQ}
\newblock
\BIn{} \APACrefbtitle {Advances in neural information processing systems}
  {Advances in neural information processing systems}\ (\BPGS\ 1017--1025).
\PrintBackRefs{\CurrentBib}

\bibitem [\protect \citeauthoryear {%
Le~Cun%
, Bengio%
\BCBL {}\ \protect \BOthers {.}}{%
Le~Cun%
\ \protect \BOthers {.}}{%
{\protect \APACyear {1995}}%
}]{%
lecun1995convolutional}
\APACinsertmetastar {%
lecun1995convolutional}%
\begin{APACrefauthors}%
Le~Cun, Y.%
, Bengio, Y.%
\BCBL {}\ \BOthersPeriod {.}\end{APACrefauthors}%
\unskip\
\newblock
\APACrefYearMonthDay{1995}{}{}.
\newblock
{\BBOQ}\APACrefatitle {Convolutional networks for images, speech, and time
  series} {Convolutional networks for images, speech, and time series}.{\BBCQ}
\newblock
\APACjournalVolNumPages{The handbook of brain theory and neural
  networks}{3361}{10}{1995}.
\PrintBackRefs{\CurrentBib}

\bibitem [\protect \citeauthoryear {%
Le~Cun%
\ \protect \BOthers {.}}{%
Le~Cun%
\ \protect \BOthers {.}}{%
{\protect \APACyear {1990}}%
}]{%
le1990handwritten}
\APACinsertmetastar {%
le1990handwritten}%
\begin{APACrefauthors}%
Le~Cun, Y.%
, Matan, O.%
, Boser, B.%
, Denker, J\BPBI S.%
, Henderson, D.%
, Howard, R\BPBI E.%
\BDBL {}Baird, H\BPBI S.%
\end{APACrefauthors}%
\unskip\
\newblock
\APACrefYearMonthDay{1990}{}{}.
\newblock
{\BBOQ}\APACrefatitle {Handwritten zip code recognition with multilayer
  networks} {Handwritten zip code recognition with multilayer networks}.{\BBCQ}
\newblock
\BIn{} \APACrefbtitle {Proc. 10th International Conference on Pattern
  Recognition} {Proc. 10th international conference on pattern recognition}\
  (\BVOL~2, \BPGS\ 35--40).
\PrintBackRefs{\CurrentBib}

\bibitem [\protect \citeauthoryear {%
Lehky%
\ \BBA {} Sejnowski%
}{%
Lehky%
\ \BBA {} Sejnowski%
}{%
{\protect \APACyear {1988}}%
}]{%
lehky1988network}
\APACinsertmetastar {%
lehky1988network}%
\begin{APACrefauthors}%
Lehky, S\BPBI R.%
\BCBT {}\ \BBA {} Sejnowski, T\BPBI J.%
\end{APACrefauthors}%
\unskip\
\newblock
\APACrefYearMonthDay{1988}{}{}.
\newblock
{\BBOQ}\APACrefatitle {Network model of shape-from-shading: Neural function
  arises from both receptive and projective fields} {Network model of
  shape-from-shading: Neural function arises from both receptive and projective
  fields}.{\BBCQ}
\newblock
\APACjournalVolNumPages{Nature}{333}{6172}{452--454}.
\PrintBackRefs{\CurrentBib}

\bibitem [\protect \citeauthoryear {%
Lehky%
\ \BBA {} Sejnowski%
}{%
Lehky%
\ \BBA {} Sejnowski%
}{%
{\protect \APACyear {1990}}%
}]{%
lehky1990neural}
\APACinsertmetastar {%
lehky1990neural}%
\begin{APACrefauthors}%
Lehky, S\BPBI R.%
\BCBT {}\ \BBA {} Sejnowski, T\BPBI J.%
\end{APACrefauthors}%
\unskip\
\newblock
\APACrefYearMonthDay{1990}{}{}.
\newblock

\newblock
\APACjournalVolNumPages{Journal of Neuroscience}{10}{7}{2281--2299}.
\PrintBackRefs{\CurrentBib}

\bibitem [\protect \citeauthoryear {%
Lehky%
\ \BBA {} Sejnowski%
}{%
Lehky%
\ \BBA {} Sejnowski%
}{%
{\protect \APACyear {1999}}%
}]{%
lehky1999seeing}
\APACinsertmetastar {%
lehky1999seeing}%
\begin{APACrefauthors}%
Lehky, S\BPBI R.%
\BCBT {}\ \BBA {} Sejnowski, T\BPBI J.%
\end{APACrefauthors}%
\unskip\
\newblock
\APACrefYearMonthDay{1999}{}{}.
\newblock
{\BBOQ}\APACrefatitle {Seeing white: Qualia in the context of decoding
  population codes} {Seeing white: Qualia in the context of decoding population
  codes}.{\BBCQ}
\newblock
\APACjournalVolNumPages{Neural computation}{11}{6}{1261--1280}.
\PrintBackRefs{\CurrentBib}

\bibitem [\protect \citeauthoryear {%
Lennie%
, Krauskopf%
\BCBL {}\ \BBA {} Sclar%
}{%
Lennie%
\ \protect \BOthers {.}}{%
{\protect \APACyear {1990}}%
}]{%
lennie1990chromatic}
\APACinsertmetastar {%
lennie1990chromatic}%
\begin{APACrefauthors}%
Lennie, P.%
, Krauskopf, J.%
\BCBL {}\ \BBA {} Sclar, G.%
\end{APACrefauthors}%
\unskip\
\newblock
\APACrefYearMonthDay{1990}{}{}.
\newblock
{\BBOQ}\APACrefatitle {Chromatic mechanisms in striate cortex of macaque}
  {Chromatic mechanisms in striate cortex of macaque}.{\BBCQ}
\newblock
\APACjournalVolNumPages{Journal of Neuroscience}{10}{2}{649--669}.
\PrintBackRefs{\CurrentBib}

\bibitem [\protect \citeauthoryear {%
{Lettvin}%
, {Maturana}%
, {McCulloch}%
\BCBL {}\ \BBA {} {Pitts}%
}{%
{Lettvin}%
\ \protect \BOthers {.}}{%
{\protect \APACyear {1959}}%
}]{%
Lettvin1959}
\APACinsertmetastar {%
Lettvin1959}%
\begin{APACrefauthors}%
{Lettvin}, J\BPBI Y.%
, {Maturana}, H\BPBI R.%
, {McCulloch}, W\BPBI S.%
\BCBL {}\ \BBA {} {Pitts}, W\BPBI H.%
\end{APACrefauthors}%
\unskip\
\newblock
\APACrefYearMonthDay{1959}{}{}.
\newblock
{\BBOQ}\APACrefatitle {What the Frog's Eye Tells the Frog's Brain} {What the
  frog's eye tells the frog's brain}.{\BBCQ}
\newblock
\APACjournalVolNumPages{Proceedings of the IRE}{47}{11}{1940-1951}.
\PrintBackRefs{\CurrentBib}

\bibitem [\protect \citeauthoryear {%
Levick%
\ \BBA {} Thibos%
}{%
Levick%
\ \BBA {} Thibos%
}{%
{\protect \APACyear {1982}}%
}]{%
levick1982analysis}
\APACinsertmetastar {%
levick1982analysis}%
\begin{APACrefauthors}%
Levick, W.%
\BCBT {}\ \BBA {} Thibos, L.%
\end{APACrefauthors}%
\unskip\
\newblock
\APACrefYearMonthDay{1982}{}{}.
\newblock
{\BBOQ}\APACrefatitle {Analysis of orientation bias in cat retina} {Analysis of
  orientation bias in cat retina}.{\BBCQ}
\newblock
\APACjournalVolNumPages{The Journal of Physiology}{329}{}{243}.
\PrintBackRefs{\CurrentBib}

\bibitem [\protect \citeauthoryear {%
Lindsey%
, Ocko%
, Ganguli%
\BCBL {}\ \BBA {} Deny%
}{%
Lindsey%
\ \protect \BOthers {.}}{%
{\protect \APACyear {2019}}%
}]{%
lindsey2019unified}
\APACinsertmetastar {%
lindsey2019unified}%
\begin{APACrefauthors}%
Lindsey, J.%
, Ocko, S\BPBI A.%
, Ganguli, S.%
\BCBL {}\ \BBA {} Deny, S.%
\end{APACrefauthors}%
\unskip\
\newblock
\APACrefYearMonthDay{2019}{}{}.
\newblock
{\BBOQ}\APACrefatitle {{A Unified Theory of Early Visual Representations from
  Retina to Cortex through Anatomically Constrained Deep CNNs}} {{A Unified
  Theory of Early Visual Representations from Retina to Cortex through
  Anatomically Constrained Deep CNNs}}.{\BBCQ}
\newblock
\BIn{} \APACrefbtitle {{International Conference on Learning Representations}.}
  {{International Conference on Learning Representations}.}
\newblock
\begin{APACrefURL} \url{https://openreview.net/forum?id=S1xq3oR5tQ}
  \end{APACrefURL}
\PrintBackRefs{\CurrentBib}

\bibitem [\protect \citeauthoryear {%
Livingstone%
\ \BBA {} Hubel%
}{%
Livingstone%
\ \BBA {} Hubel%
}{%
{\protect \APACyear {1984}}%
}]{%
livingstone1984anatomy}
\APACinsertmetastar {%
livingstone1984anatomy}%
\begin{APACrefauthors}%
Livingstone, M\BPBI S.%
\BCBT {}\ \BBA {} Hubel, D\BPBI H.%
\end{APACrefauthors}%
\unskip\
\newblock
\APACrefYearMonthDay{1984}{}{}.
\newblock
{\BBOQ}\APACrefatitle {Anatomy and physiology of a color system in the primate
  visual cortex} {Anatomy and physiology of a color system in the primate
  visual cortex}.{\BBCQ}
\newblock
\APACjournalVolNumPages{Journal of Neuroscience}{4}{1}{309--356}.
\PrintBackRefs{\CurrentBib}

\bibitem [\protect \citeauthoryear {%
Long%
, Yang%
\BCBL {}\ \BBA {} Purves%
}{%
Long%
\ \protect \BOthers {.}}{%
{\protect \APACyear {2006}}%
}]{%
long2006spectral}
\APACinsertmetastar {%
long2006spectral}%
\begin{APACrefauthors}%
Long, F.%
, Yang, Z.%
\BCBL {}\ \BBA {} Purves, D.%
\end{APACrefauthors}%
\unskip\
\newblock
\APACrefYearMonthDay{2006}{}{}.
\newblock
{\BBOQ}\APACrefatitle {Spectral statistics in natural scenes predict hue,
  saturation, and brightness} {Spectral statistics in natural scenes predict
  hue, saturation, and brightness}.{\BBCQ}
\newblock
\APACjournalVolNumPages{Proceedings of the National Academy of
  Sciences}{103}{15}{6013--6018}.
\PrintBackRefs{\CurrentBib}

\bibitem [\protect \citeauthoryear {%
Lowe%
}{%
Lowe%
}{%
{\protect \APACyear {1999}}%
}]{%
lowe1999object}
\APACinsertmetastar {%
lowe1999object}%
\begin{APACrefauthors}%
Lowe, D\BPBI G.%
\end{APACrefauthors}%
\unskip\
\newblock
\APACrefYearMonthDay{1999}{}{}.
\newblock
{\BBOQ}\APACrefatitle {Object recognition from local scale-invariant features}
  {Object recognition from local scale-invariant features}.{\BBCQ}
\newblock
\BIn{} \APACrefbtitle {Proceedings of the seventh IEEE international conference
  on computer vision} {Proceedings of the seventh ieee international conference
  on computer vision}\ (\BVOL~2, \BPGS\ 1150--1157).
\PrintBackRefs{\CurrentBib}

\bibitem [\protect \citeauthoryear {%
Marquez%
, Hare%
\BCBL {}\ \BBA {} Niranjan%
}{%
Marquez%
\ \protect \BOthers {.}}{%
{\protect \APACyear {2018}}%
}]{%
marquez2018deep}
\APACinsertmetastar {%
marquez2018deep}%
\begin{APACrefauthors}%
Marquez, E\BPBI S.%
, Hare, J\BPBI S.%
\BCBL {}\ \BBA {} Niranjan, M.%
\end{APACrefauthors}%
\unskip\
\newblock
\APACrefYearMonthDay{2018}{}{}.
\newblock
{\BBOQ}\APACrefatitle {Deep cascade learning} {Deep cascade learning}.{\BBCQ}
\newblock
\APACjournalVolNumPages{IEEE transactions on neural networks and learning
  systems}{29}{11}{5475--5485}.
\PrintBackRefs{\CurrentBib}

\bibitem [\protect \citeauthoryear {%
Marr%
\ \BBA {} Hildreth%
}{%
Marr%
\ \BBA {} Hildreth%
}{%
{\protect \APACyear {1980}}%
}]{%
marr1980theory}
\APACinsertmetastar {%
marr1980theory}%
\begin{APACrefauthors}%
Marr, D.%
\BCBT {}\ \BBA {} Hildreth, E.%
\end{APACrefauthors}%
\unskip\
\newblock
\APACrefYearMonthDay{1980}{}{}.
\newblock
{\BBOQ}\APACrefatitle {Theory of edge detection} {Theory of edge
  detection}.{\BBCQ}
\newblock
\APACjournalVolNumPages{Proceedings of the Royal Society of London. Series B.
  Biological Sciences}{207}{1167}{187--217}.
\PrintBackRefs{\CurrentBib}

\bibitem [\protect \citeauthoryear {%
Martinez%
\ \BBA {} Alonso%
}{%
Martinez%
\ \BBA {} Alonso%
}{%
{\protect \APACyear {2003}}%
}]{%
martinez2003complex}
\APACinsertmetastar {%
martinez2003complex}%
\begin{APACrefauthors}%
Martinez, L\BPBI M.%
\BCBT {}\ \BBA {} Alonso, J\BHBI M.%
\end{APACrefauthors}%
\unskip\
\newblock
\APACrefYearMonthDay{2003}{}{}.
\newblock
{\BBOQ}\APACrefatitle {Complex receptive fields in primary visual cortex}
  {Complex receptive fields in primary visual cortex}.{\BBCQ}
\newblock
\APACjournalVolNumPages{The neuroscientist}{9}{5}{317--331}.
\PrintBackRefs{\CurrentBib}

\bibitem [\protect \citeauthoryear {%
Maxwell%
}{%
Maxwell%
}{%
{\protect \APACyear {1860}}%
}]{%
maxwell1860iv}
\APACinsertmetastar {%
maxwell1860iv}%
\begin{APACrefauthors}%
Maxwell, J\BPBI C.%
\end{APACrefauthors}%
\unskip\
\newblock
\APACrefYearMonthDay{1860}{}{}.
\newblock
{\BBOQ}\APACrefatitle {{IV. On the theory of compound colours, and the
  relations of the colours of the spectrum}} {{IV. On the theory of compound
  colours, and the relations of the colours of the spectrum}}.{\BBCQ}
\newblock
\APACjournalVolNumPages{Philosophical Transactions of the Royal Society of
  London}{}{150}{57--84}.
\PrintBackRefs{\CurrentBib}

\bibitem [\protect \citeauthoryear {%
McConnell%
}{%
McConnell%
}{%
{\protect \APACyear {1986}}%
}]{%
mcconnell1986method}
\APACinsertmetastar {%
mcconnell1986method}%
\begin{APACrefauthors}%
McConnell, R\BPBI K.%
\end{APACrefauthors}%
\unskip\
\newblock
\APACrefYearMonthDay{1986}{{\APACmonth{01}}~28}{}.
\newblock
\APACrefbtitle {Method of and apparatus for pattern recognition.} {Method of
  and apparatus for pattern recognition.}
\newblock
\APACaddressPublisher{}{Google Patents}.
\newblock
\APACrefnote{US Patent 4,567,610}
\PrintBackRefs{\CurrentBib}

\bibitem [\protect \citeauthoryear {%
Naka%
\ \BBA {} Rushton%
}{%
Naka%
\ \BBA {} Rushton%
}{%
{\protect \APACyear {1966}}%
}]{%
naka1966s}
\APACinsertmetastar {%
naka1966s}%
\begin{APACrefauthors}%
Naka, K.%
\BCBT {}\ \BBA {} Rushton, W\BPBI A.%
\end{APACrefauthors}%
\unskip\
\newblock
\APACrefYearMonthDay{1966}{}{}.
\newblock
{\BBOQ}\APACrefatitle {S-potentials from colour units in the retina of fish
  (Cyprinidae)} {S-potentials from colour units in the retina of fish
  (cyprinidae)}.{\BBCQ}
\newblock
\APACjournalVolNumPages{The Journal of physiology}{185}{3}{536--555}.
\PrintBackRefs{\CurrentBib}

\bibitem [\protect \citeauthoryear {%
Netzer%
\ \protect \BOthers {.}}{%
Netzer%
\ \protect \BOthers {.}}{%
{\protect \APACyear {2011}}%
}]{%
netzer2011reading}
\APACinsertmetastar {%
netzer2011reading}%
\begin{APACrefauthors}%
Netzer, Y.%
, Wang, T.%
, Coates, A.%
, Bissacco, A.%
, Wu, B.%
\BCBL {}\ \BBA {} Ng, A\BPBI Y.%
\end{APACrefauthors}%
\unskip\
\newblock
\APACrefYearMonthDay{2011}{}{}.
\newblock
{\BBOQ}\APACrefatitle {Reading digits in natural images with unsupervised
  feature learning} {Reading digits in natural images with unsupervised feature
  learning}.{\BBCQ}
\newblock

\PrintBackRefs{\CurrentBib}

\bibitem [\protect \citeauthoryear {%
Olah%
\ \protect \BOthers {.}}{%
Olah%
\ \protect \BOthers {.}}{%
{\protect \APACyear {2020}}%
{\protect \APACexlab {{\protect \BCnt {1}}}}}]{%
olah2020overview}
\APACinsertmetastar {%
olah2020overview}%
\begin{APACrefauthors}%
Olah, C.%
, Cammarata, N.%
, Schubert, L.%
, Goh, G.%
, Petrov, M.%
\BCBL {}\ \BBA {} Carter, S.%
\end{APACrefauthors}%
\unskip\
\newblock
\APACrefYearMonthDay{2020{\protect \BCnt {1}}}{}{}.
\newblock
{\BBOQ}\APACrefatitle {An Overview of Early Vision in InceptionV1} {An overview
  of early vision in inceptionv1}.{\BBCQ}
\newblock
\APACjournalVolNumPages{Distill}{5}{4}{e00024--002}.
\PrintBackRefs{\CurrentBib}

\bibitem [\protect \citeauthoryear {%
Olah%
\ \protect \BOthers {.}}{%
Olah%
\ \protect \BOthers {.}}{%
{\protect \APACyear {2020}}%
{\protect \APACexlab {{\protect \BCnt {2}}}}}]{%
olah2020zoom}
\APACinsertmetastar {%
olah2020zoom}%
\begin{APACrefauthors}%
Olah, C.%
, Cammarata, N.%
, Schubert, L.%
, Goh, G.%
, Petrov, M.%
\BCBL {}\ \BBA {} Carter, S.%
\end{APACrefauthors}%
\unskip\
\newblock
\APACrefYearMonthDay{2020{\protect \BCnt {2}}}{}{}.
\newblock
{\BBOQ}\APACrefatitle {Zoom In: An Introduction to Circuits} {Zoom in: An
  introduction to circuits}.{\BBCQ}
\newblock
\APACjournalVolNumPages{Distill}{5}{3}{e00024--001}.
\PrintBackRefs{\CurrentBib}

\bibitem [\protect \citeauthoryear {%
Olah%
, Mordvintsev%
\BCBL {}\ \BBA {} Schubert%
}{%
Olah%
\ \protect \BOthers {.}}{%
{\protect \APACyear {2017}}%
}]{%
olah2017feature}
\APACinsertmetastar {%
olah2017feature}%
\begin{APACrefauthors}%
Olah, C.%
, Mordvintsev, A.%
\BCBL {}\ \BBA {} Schubert, L.%
\end{APACrefauthors}%
\unskip\
\newblock
\APACrefYearMonthDay{2017}{}{}.
\newblock
{\BBOQ}\APACrefatitle {Feature visualization} {Feature visualization}.{\BBCQ}
\newblock
\APACjournalVolNumPages{Distill}{2}{11}{e7}.
\PrintBackRefs{\CurrentBib}

\bibitem [\protect \citeauthoryear {%
Olah%
\ \protect \BOthers {.}}{%
Olah%
\ \protect \BOthers {.}}{%
{\protect \APACyear {2018}}%
}]{%
olah2018building}
\APACinsertmetastar {%
olah2018building}%
\begin{APACrefauthors}%
Olah, C.%
, Satyanarayan, A.%
, Johnson, I.%
, Carter, S.%
, Schubert, L.%
, Ye, K.%
\BCBL {}\ \BBA {} Mordvintsev, A.%
\end{APACrefauthors}%
\unskip\
\newblock
\APACrefYearMonthDay{2018}{}{}.
\newblock
{\BBOQ}\APACrefatitle {The building blocks of interpretability} {The building
  blocks of interpretability}.{\BBCQ}
\newblock
\APACjournalVolNumPages{Distill}{3}{3}{e10}.
\PrintBackRefs{\CurrentBib}

\bibitem [\protect \citeauthoryear {%
Olshausen%
\ \BBA {} Field%
}{%
Olshausen%
\ \BBA {} Field%
}{%
{\protect \APACyear {1996}}%
}]{%
olshausen1996emergence}
\APACinsertmetastar {%
olshausen1996emergence}%
\begin{APACrefauthors}%
Olshausen, B\BPBI A.%
\BCBT {}\ \BBA {} Field, D\BPBI J.%
\end{APACrefauthors}%
\unskip\
\newblock
\APACrefYearMonthDay{1996}{}{}.
\newblock
{\BBOQ}\APACrefatitle {Emergence of simple-cell receptive field properties by
  learning a sparse code for natural images} {Emergence of simple-cell
  receptive field properties by learning a sparse code for natural
  images}.{\BBCQ}
\newblock
\APACjournalVolNumPages{Nature}{381}{6583}{607--609}.
\PrintBackRefs{\CurrentBib}

\bibitem [\protect \citeauthoryear {%
Peirce%
\ \protect \BOthers {.}}{%
Peirce%
\ \protect \BOthers {.}}{%
{\protect \APACyear {2019}}%
}]{%
peirce2019psychopy2}
\APACinsertmetastar {%
peirce2019psychopy2}%
\begin{APACrefauthors}%
Peirce, J.%
, Gray, J\BPBI R.%
, Simpson, S.%
, MacAskill, M.%
, H{\"o}chenberger, R.%
, Sogo, H.%
\BDBL {}Lindel{\o}v, J\BPBI K.%
\end{APACrefauthors}%
\unskip\
\newblock
\APACrefYearMonthDay{2019}{}{}.
\newblock
{\BBOQ}\APACrefatitle {PsychoPy2: Experiments in behavior made easy}
  {Psychopy2: Experiments in behavior made easy}.{\BBCQ}
\newblock
\APACjournalVolNumPages{Behavior research methods}{51}{1}{195--203}.
\PrintBackRefs{\CurrentBib}

\bibitem [\protect \citeauthoryear {%
Pridmore%
}{%
Pridmore%
}{%
{\protect \APACyear {2005}}%
}]{%
pridmore2005theory}
\APACinsertmetastar {%
pridmore2005theory}%
\begin{APACrefauthors}%
Pridmore, R\BPBI W.%
\end{APACrefauthors}%
\unskip\
\newblock
\APACrefYearMonthDay{2005}{}{}.
\newblock
{\BBOQ}\APACrefatitle {Theory of corresponding colors as complementary sets}
  {Theory of corresponding colors as complementary sets}.{\BBCQ}
\newblock
\APACjournalVolNumPages{Color Research \& Application: Endorsed by
  Inter-Society Color Council, The Colour Group (Great Britain), Canadian
  Society for Color, Color Science Association of Japan, Dutch Society for the
  Study of Color, The Swedish Colour Centre Foundation, Colour Society of
  Australia, Centre Fran{\c{c}}ais de la Couleur}{30}{5}{371--381}.
\PrintBackRefs{\CurrentBib}

\bibitem [\protect \citeauthoryear {%
Pridmore%
}{%
Pridmore%
}{%
{\protect \APACyear {2011}}%
}]{%
pridmore2011complementary}
\APACinsertmetastar {%
pridmore2011complementary}%
\begin{APACrefauthors}%
Pridmore, R\BPBI W.%
\end{APACrefauthors}%
\unskip\
\newblock
\APACrefYearMonthDay{2011}{}{}.
\newblock
{\BBOQ}\APACrefatitle {Complementary colors theory of color vision: Physiology,
  color mixture, color constancy and color perception} {Complementary colors
  theory of color vision: Physiology, color mixture, color constancy and color
  perception}.{\BBCQ}
\newblock
\APACjournalVolNumPages{Color Research \& Application}{36}{6}{394--412}.
\PrintBackRefs{\CurrentBib}

\bibitem [\protect \citeauthoryear {%
Rafegas%
\ \BBA {} Vanrell%
}{%
Rafegas%
\ \BBA {} Vanrell%
}{%
{\protect \APACyear {2018}}%
}]{%
rafegas2018color}
\APACinsertmetastar {%
rafegas2018color}%
\begin{APACrefauthors}%
Rafegas, I.%
\BCBT {}\ \BBA {} Vanrell, M.%
\end{APACrefauthors}%
\unskip\
\newblock
\APACrefYearMonthDay{2018}{}{}.
\newblock
{\BBOQ}\APACrefatitle {Color encoding in biologically-inspired convolutional
  neural networks} {Color encoding in biologically-inspired convolutional
  neural networks}.{\BBCQ}
\newblock
\APACjournalVolNumPages{Vision research}{151}{}{7--17}.
\PrintBackRefs{\CurrentBib}

\bibitem [\protect \citeauthoryear {%
Russakovsky%
\ \protect \BOthers {.}}{%
Russakovsky%
\ \protect \BOthers {.}}{%
{\protect \APACyear {2015}}%
}]{%
russakovsky2015imagenet}
\APACinsertmetastar {%
russakovsky2015imagenet}%
\begin{APACrefauthors}%
Russakovsky, O.%
, Deng, J.%
, Su, H.%
, Krause, J.%
, Satheesh, S.%
, Ma, S.%
\BDBL {}others%
\end{APACrefauthors}%
\unskip\
\newblock
\APACrefYearMonthDay{2015}{}{}.
\newblock
{\BBOQ}\APACrefatitle {Imagenet large scale visual recognition challenge}
  {Imagenet large scale visual recognition challenge}.{\BBCQ}
\newblock
\APACjournalVolNumPages{International journal of computer
  vision}{115}{3}{211--252}.
\PrintBackRefs{\CurrentBib}

\bibitem [\protect \citeauthoryear {%
Schluppeck%
\ \BBA {} Engel%
}{%
Schluppeck%
\ \BBA {} Engel%
}{%
{\protect \APACyear {2002}}%
}]{%
schluppeck2002color}
\APACinsertmetastar {%
schluppeck2002color}%
\begin{APACrefauthors}%
Schluppeck, D.%
\BCBT {}\ \BBA {} Engel, S\BPBI A.%
\end{APACrefauthors}%
\unskip\
\newblock
\APACrefYearMonthDay{2002}{}{}.
\newblock
{\BBOQ}\APACrefatitle {Color opponent neurons in V1: a review and model
  reconciling results from imaging and single-unit recording} {Color opponent
  neurons in v1: a review and model reconciling results from imaging and
  single-unit recording}.{\BBCQ}
\newblock
\APACjournalVolNumPages{Journal of vision}{2}{6}{5--5}.
\PrintBackRefs{\CurrentBib}

\bibitem [\protect \citeauthoryear {%
Schrimpf%
\ \protect \BOthers {.}}{%
Schrimpf%
\ \protect \BOthers {.}}{%
{\protect \APACyear {2018}}%
}]{%
SchrimpfKubilius2018BrainScore}
\APACinsertmetastar {%
SchrimpfKubilius2018BrainScore}%
\begin{APACrefauthors}%
Schrimpf, M.%
, Kubilius, J.%
, Hong, H.%
, Majaj, N\BPBI J.%
, Rajalingham, R.%
, Issa, E\BPBI B.%
\BDBL {}DiCarlo, J\BPBI J.%
\end{APACrefauthors}%
\unskip\
\newblock
\APACrefYearMonthDay{2018}{}{}.
\newblock
{\BBOQ}\APACrefatitle {Brain-Score: Which Artificial Neural Network for Object
  Recognition is most Brain-Like?} {Brain-score: Which artificial neural
  network for object recognition is most brain-like?}{\BBCQ}
\newblock
\APACjournalVolNumPages{bioRxiv preprint}{}{}{}.
\PrintBackRefs{\CurrentBib}

\bibitem [\protect \citeauthoryear {%
Seymour%
, Williams%
\BCBL {}\ \BBA {} Rich%
}{%
Seymour%
\ \protect \BOthers {.}}{%
{\protect \APACyear {2015}}%
}]{%
10.1093/cercor/bhv021}
\APACinsertmetastar {%
10.1093/cercor/bhv021}%
\begin{APACrefauthors}%
Seymour, K\BPBI J.%
, Williams, M\BPBI A.%
\BCBL {}\ \BBA {} Rich, A\BPBI N.%
\end{APACrefauthors}%
\unskip\
\newblock
\APACrefYearMonthDay{2015}{02}{}.
\newblock
{\BBOQ}\APACrefatitle {{The Representation of Color across the Human Visual
  Cortex: Distinguishing Chromatic Signals Contributing to Object Form Versus
  Surface Color}} {{The Representation of Color across the Human Visual Cortex:
  Distinguishing Chromatic Signals Contributing to Object Form Versus Surface
  Color}}.{\BBCQ}
\newblock
\APACjournalVolNumPages{Cerebral Cortex}{26}{5}{1997-2005}.
\newblock
\begin{APACrefURL} \url{https://doi.org/10.1093/cercor/bhv021} \end{APACrefURL}
\newblock
\begin{APACrefDOI} \doi{10.1093/cercor/bhv021} \end{APACrefDOI}
\PrintBackRefs{\CurrentBib}

\bibitem [\protect \citeauthoryear {%
Shan%
, Zhang%
\BCBL {}\ \BBA {} Cottrell%
}{%
Shan%
\ \protect \BOthers {.}}{%
{\protect \APACyear {2007}}%
}]{%
NIPS2006_3018}
\APACinsertmetastar {%
NIPS2006_3018}%
\begin{APACrefauthors}%
Shan, H.%
, Zhang, L.%
\BCBL {}\ \BBA {} Cottrell, G\BPBI W.%
\end{APACrefauthors}%
\unskip\
\newblock
\APACrefYearMonthDay{2007}{}{}.
\newblock
{\BBOQ}\APACrefatitle {Recursive ICA} {Recursive ica}.{\BBCQ}
\newblock
\BIn{} B.~Sch\"{o}lkopf, J\BPBI C.~Platt\BCBL {}\ \BBA {} T.~Hoffman\ (\BEDS),
  \APACrefbtitle {Advances in Neural Information Processing Systems 19}
  {Advances in neural information processing systems 19}\ (\BPGS\ 1273--1280).
\newblock
\APACaddressPublisher{}{MIT Press}.
\newblock
\begin{APACrefURL} \url{http://papers.nips.cc/paper/3018-recursive-ica.pdf}
  \end{APACrefURL}
\PrintBackRefs{\CurrentBib}

\bibitem [\protect \citeauthoryear {%
Shapley%
\ \BBA {} Hawken%
}{%
Shapley%
\ \BBA {} Hawken%
}{%
{\protect \APACyear {2011}}%
}]{%
SHAPLEY2011701}
\APACinsertmetastar {%
SHAPLEY2011701}%
\begin{APACrefauthors}%
Shapley, R.%
\BCBT {}\ \BBA {} Hawken, M\BPBI J.%
\end{APACrefauthors}%
\unskip\
\newblock
\APACrefYearMonthDay{2011}{}{}.
\newblock
{\BBOQ}\APACrefatitle {{Color in the Cortex: single- and double-opponent
  cells}} {{Color in the Cortex: single- and double-opponent cells}}.{\BBCQ}
\newblock
\APACjournalVolNumPages{Vision Research}{51}{7}{701 - 717}.
\newblock
\begin{APACrefURL}
  \url{http://www.sciencedirect.com/science/article/pii/S0042698911000526}
  \end{APACrefURL}
\newblock
\APACrefnote{Vision Research 50th Anniversary Issue: Part 1}
\newblock
\begin{APACrefDOI} \doi{https://doi.org/10.1016/j.visres.2011.02.012}
  \end{APACrefDOI}
\PrintBackRefs{\CurrentBib}

\bibitem [\protect \citeauthoryear {%
Shevell%
\ \BBA {} Martin%
}{%
Shevell%
\ \BBA {} Martin%
}{%
{\protect \APACyear {2017}}%
}]{%
Shevell:17}
\APACinsertmetastar {%
Shevell:17}%
\begin{APACrefauthors}%
Shevell, S\BPBI K.%
\BCBT {}\ \BBA {} Martin, P\BPBI R.%
\end{APACrefauthors}%
\unskip\
\newblock
\APACrefYearMonthDay{2017}{Jul}{}.
\newblock
{\BBOQ}\APACrefatitle {Color opponency: tutorial} {Color opponency:
  tutorial}.{\BBCQ}
\newblock
\APACjournalVolNumPages{J. Opt. Soc. Am. A}{34}{7}{1099--1108}.
\PrintBackRefs{\CurrentBib}

\bibitem [\protect \citeauthoryear {%
Smith%
, Lee%
, Pokorny%
, Martin%
\BCBL {}\ \BBA {} Valberg%
}{%
Smith%
\ \protect \BOthers {.}}{%
{\protect \APACyear {1992}}%
}]{%
smith1992responses}
\APACinsertmetastar {%
smith1992responses}%
\begin{APACrefauthors}%
Smith, V\BPBI C.%
, Lee, B.%
, Pokorny, J.%
, Martin, P.%
\BCBL {}\ \BBA {} Valberg, A.%
\end{APACrefauthors}%
\unskip\
\newblock
\APACrefYearMonthDay{1992}{}{}.
\newblock
{\BBOQ}\APACrefatitle {Responses of macaque ganglion cells to the relative
  phase of heterochromatically modulated lights.} {Responses of macaque
  ganglion cells to the relative phase of heterochromatically modulated
  lights.}{\BBCQ}
\newblock
\APACjournalVolNumPages{The Journal of Physiology}{458}{1}{191--221}.
\PrintBackRefs{\CurrentBib}

\bibitem [\protect \citeauthoryear {%
Szegedy%
\ \protect \BOthers {.}}{%
Szegedy%
\ \protect \BOthers {.}}{%
{\protect \APACyear {2015}}%
}]{%
szegedy2015going}
\APACinsertmetastar {%
szegedy2015going}%
\begin{APACrefauthors}%
Szegedy, C.%
, Liu, W.%
, Jia, Y.%
, Sermanet, P.%
, Reed, S.%
, Anguelov, D.%
\BDBL {}Rabinovich, A.%
\end{APACrefauthors}%
\unskip\
\newblock
\APACrefYearMonthDay{2015}{}{}.
\newblock
{\BBOQ}\APACrefatitle {Going deeper with convolutions} {Going deeper with
  convolutions}.{\BBCQ}
\newblock
\BIn{} \APACrefbtitle {Proceedings of the IEEE conference on computer vision
  and pattern recognition} {Proceedings of the ieee conference on computer
  vision and pattern recognition}\ (\BPGS\ 1--9).
\PrintBackRefs{\CurrentBib}

\bibitem [\protect \citeauthoryear {%
Tan%
\ \BBA {} Le%
}{%
Tan%
\ \BBA {} Le%
}{%
{\protect \APACyear {2019}}%
}]{%
tan2019efficientnet}
\APACinsertmetastar {%
tan2019efficientnet}%
\begin{APACrefauthors}%
Tan, M.%
\BCBT {}\ \BBA {} Le, Q\BPBI V.%
\end{APACrefauthors}%
\unskip\
\newblock
\APACrefYearMonthDay{2019}{}{}.
\newblock
{\BBOQ}\APACrefatitle {Efficientnet: Rethinking model scaling for convolutional
  neural networks} {Efficientnet: Rethinking model scaling for convolutional
  neural networks}.{\BBCQ}
\newblock
\APACjournalVolNumPages{arXiv preprint arXiv:1905.11946}{}{}{}.
\PrintBackRefs{\CurrentBib}

\bibitem [\protect \citeauthoryear {%
Troy%
\ \BBA {} Shou%
}{%
Troy%
\ \BBA {} Shou%
}{%
{\protect \APACyear {2002}}%
}]{%
troy2002receptive}
\APACinsertmetastar {%
troy2002receptive}%
\begin{APACrefauthors}%
Troy, J\BPBI B.%
\BCBT {}\ \BBA {} Shou, T.%
\end{APACrefauthors}%
\unskip\
\newblock
\APACrefYearMonthDay{2002}{}{}.
\newblock
{\BBOQ}\APACrefatitle {The receptive fields of cat retinal ganglion cells in
  physiological and pathological states: where we are after half a century of
  research} {The receptive fields of cat retinal ganglion cells in
  physiological and pathological states: where we are after half a century of
  research}.{\BBCQ}
\newblock
\APACjournalVolNumPages{Progress in retinal and eye research}{21}{3}{263--302}.
\PrintBackRefs{\CurrentBib}

\bibitem [\protect \citeauthoryear {%
Wade%
, Augath%
, Logothetis%
\BCBL {}\ \BBA {} Wandell%
}{%
Wade%
\ \protect \BOthers {.}}{%
{\protect \APACyear {2008}}%
}]{%
Wade2008fMRIMO}
\APACinsertmetastar {%
Wade2008fMRIMO}%
\begin{APACrefauthors}%
Wade, A\BPBI R.%
, Augath, M.%
, Logothetis, N\BPBI K.%
\BCBL {}\ \BBA {} Wandell, B\BPBI A.%
\end{APACrefauthors}%
\unskip\
\newblock
\APACrefYearMonthDay{2008}{}{}.
\newblock
{\BBOQ}\APACrefatitle {fMRI measurements of color in macaque and human.} {fmri
  measurements of color in macaque and human.}{\BBCQ}
\newblock
\APACjournalVolNumPages{Journal of vision}{8 10}{}{6.1-19}.
\PrintBackRefs{\CurrentBib}

\bibitem [\protect \citeauthoryear {%
Wagner%
, MacNichol%
\BCBL {}\ \BBA {} Wolbarsht%
}{%
Wagner%
\ \protect \BOthers {.}}{%
{\protect \APACyear {1960}}%
}]{%
wagner1960opponent}
\APACinsertmetastar {%
wagner1960opponent}%
\begin{APACrefauthors}%
Wagner, H\BPBI G.%
, MacNichol, E.%
\BCBL {}\ \BBA {} Wolbarsht, M\BPBI L.%
\end{APACrefauthors}%
\unskip\
\newblock
\APACrefYearMonthDay{1960}{}{}.
\newblock
{\BBOQ}\APACrefatitle {Opponent color responses in retinal ganglion cells}
  {Opponent color responses in retinal ganglion cells}.{\BBCQ}
\newblock
\APACjournalVolNumPages{Science}{131}{3409}{1314--1314}.
\PrintBackRefs{\CurrentBib}

\bibitem [\protect \citeauthoryear {%
Wang%
, Cottrell%
\BCBL {}\ \BBA {} Kanan%
}{%
Wang%
\ \protect \BOthers {.}}{%
{\protect \APACyear {2015}}%
}]{%
Wang2015ModelingTO}
\APACinsertmetastar {%
Wang2015ModelingTO}%
\begin{APACrefauthors}%
Wang, P.%
, Cottrell, G\BPBI W.%
\BCBL {}\ \BBA {} Kanan, C.%
\end{APACrefauthors}%
\unskip\
\newblock
\APACrefYearMonthDay{2015}{}{}.
\newblock
{\BBOQ}\APACrefatitle {Modeling the Object Recognition Pathway: A Deep
  Hierarchical Model Using Gnostic Fields} {Modeling the object recognition
  pathway: A deep hierarchical model using gnostic fields}.{\BBCQ}
\newblock
\BIn{} \APACrefbtitle {CogSci.} {Cogsci.}
\PrintBackRefs{\CurrentBib}

\bibitem [\protect \citeauthoryear {%
Wiesel%
\ \BBA {} Hubel%
}{%
Wiesel%
\ \BBA {} Hubel%
}{%
{\protect \APACyear {1966}}%
}]{%
wiesel1966spatial}
\APACinsertmetastar {%
wiesel1966spatial}%
\begin{APACrefauthors}%
Wiesel, T\BPBI N.%
\BCBT {}\ \BBA {} Hubel, D\BPBI H.%
\end{APACrefauthors}%
\unskip\
\newblock
\APACrefYearMonthDay{1966}{}{}.
\newblock
{\BBOQ}\APACrefatitle {Spatial and chromatic interactions in the lateral
  geniculate body of the rhesus monkey.} {Spatial and chromatic interactions in
  the lateral geniculate body of the rhesus monkey.}{\BBCQ}
\newblock
\APACjournalVolNumPages{Journal of neurophysiology}{29}{6}{1115--1156}.
\PrintBackRefs{\CurrentBib}

\bibitem [\protect \citeauthoryear {%
Yamins%
\ \protect \BOthers {.}}{%
Yamins%
\ \protect \BOthers {.}}{%
{\protect \APACyear {2014}}%
}]{%
yamins2014performance}
\APACinsertmetastar {%
yamins2014performance}%
\begin{APACrefauthors}%
Yamins, D\BPBI L.%
, Hong, H.%
, Cadieu, C\BPBI F.%
, Solomon, E\BPBI A.%
, Seibert, D.%
\BCBL {}\ \BBA {} DiCarlo, J\BPBI J.%
\end{APACrefauthors}%
\unskip\
\newblock
\APACrefYearMonthDay{2014}{}{}.
\newblock
{\BBOQ}\APACrefatitle {Performance-optimized hierarchical models predict neural
  responses in higher visual cortex} {Performance-optimized hierarchical models
  predict neural responses in higher visual cortex}.{\BBCQ}
\newblock
\APACjournalVolNumPages{Proceedings of the National Academy of
  Sciences}{111}{23}{8619--8624}.
\PrintBackRefs{\CurrentBib}

\bibitem [\protect \citeauthoryear {%
Young%
}{%
Young%
}{%
{\protect \APACyear {1802}}%
}]{%
young1802ii}
\APACinsertmetastar {%
young1802ii}%
\begin{APACrefauthors}%
Young, T.%
\end{APACrefauthors}%
\unskip\
\newblock
\APACrefYearMonthDay{1802}{}{}.
\newblock
{\BBOQ}\APACrefatitle {II. The Bakerian Lecture. On the theory of light and
  colours} {Ii. the bakerian lecture. on the theory of light and
  colours}.{\BBCQ}
\newblock
\APACjournalVolNumPages{Philosophical transactions of the Royal Society of
  London}{}{92}{12--48}.
\PrintBackRefs{\CurrentBib}

\bibitem [\protect \citeauthoryear {%
Zagoruyko%
\ \BBA {} Komodakis%
}{%
Zagoruyko%
\ \BBA {} Komodakis%
}{%
{\protect \APACyear {2016}}%
}]{%
zagoruyko2016wide}
\APACinsertmetastar {%
zagoruyko2016wide}%
\begin{APACrefauthors}%
Zagoruyko, S.%
\BCBT {}\ \BBA {} Komodakis, N.%
\end{APACrefauthors}%
\unskip\
\newblock
\APACrefYearMonthDay{2016}{}{}.
\newblock
{\BBOQ}\APACrefatitle {Wide residual networks} {Wide residual networks}.{\BBCQ}
\newblock
\APACjournalVolNumPages{arXiv preprint arXiv:1605.07146}{}{}{}.
\PrintBackRefs{\CurrentBib}

\bibitem [\protect \citeauthoryear {%
Zeiler%
\ \BBA {} Fergus%
}{%
Zeiler%
\ \BBA {} Fergus%
}{%
{\protect \APACyear {2014}}%
}]{%
zeiler2014visualizing}
\APACinsertmetastar {%
zeiler2014visualizing}%
\begin{APACrefauthors}%
Zeiler, M\BPBI D.%
\BCBT {}\ \BBA {} Fergus, R.%
\end{APACrefauthors}%
\unskip\
\newblock
\APACrefYearMonthDay{2014}{}{}.
\newblock
{\BBOQ}\APACrefatitle {Visualizing and understanding convolutional networks}
  {Visualizing and understanding convolutional networks}.{\BBCQ}
\newblock
\BIn{} \APACrefbtitle {European conference on computer vision} {European
  conference on computer vision}\ (\BPGS\ 818--833).
\PrintBackRefs{\CurrentBib}

\bibitem [\protect \citeauthoryear {%
Zhang%
, Kim%
, Sanes%
\BCBL {}\ \BBA {} Meister%
}{%
Zhang%
\ \protect \BOthers {.}}{%
{\protect \APACyear {2012}}%
}]{%
zhang2012most}
\APACinsertmetastar {%
zhang2012most}%
\begin{APACrefauthors}%
Zhang, Y.%
, Kim, I\BHBI J.%
, Sanes, J\BPBI R.%
\BCBL {}\ \BBA {} Meister, M.%
\end{APACrefauthors}%
\unskip\
\newblock
\APACrefYearMonthDay{2012}{}{}.
\newblock
{\BBOQ}\APACrefatitle {The most numerous ganglion cell type of the mouse retina
  is a selective feature detector} {The most numerous ganglion cell type of the
  mouse retina is a selective feature detector}.{\BBCQ}
\newblock
\APACjournalVolNumPages{Proceedings of the National Academy of
  Sciences}{109}{36}{E2391--E2398}.
\PrintBackRefs{\CurrentBib}

\bibitem [\protect \citeauthoryear {%
Zhao%
, Chen%
, Liu%
\BCBL {}\ \BBA {} Cang%
}{%
Zhao%
\ \protect \BOthers {.}}{%
{\protect \APACyear {2013}}%
}]{%
zhao2013orientation}
\APACinsertmetastar {%
zhao2013orientation}%
\begin{APACrefauthors}%
Zhao, X.%
, Chen, H.%
, Liu, X.%
\BCBL {}\ \BBA {} Cang, J.%
\end{APACrefauthors}%
\unskip\
\newblock
\APACrefYearMonthDay{2013}{}{}.
\newblock
{\BBOQ}\APACrefatitle {Orientation-selective responses in the mouse lateral
  geniculate nucleus} {Orientation-selective responses in the mouse lateral
  geniculate nucleus}.{\BBCQ}
\newblock
\APACjournalVolNumPages{Journal of Neuroscience}{33}{31}{12751--12763}.
\PrintBackRefs{\CurrentBib}

\bibitem [\protect \citeauthoryear {%
Zhaoping%
, Geisler%
\BCBL {}\ \BBA {} May%
}{%
Zhaoping%
\ \protect \BOthers {.}}{%
{\protect \APACyear {2011}}%
}]{%
zhaoping2011human}
\APACinsertmetastar {%
zhaoping2011human}%
\begin{APACrefauthors}%
Zhaoping, L.%
, Geisler, W\BPBI S.%
\BCBL {}\ \BBA {} May, K\BPBI A.%
\end{APACrefauthors}%
\unskip\
\newblock
\APACrefYearMonthDay{2011}{}{}.
\newblock
{\BBOQ}\APACrefatitle {Human wavelength discrimination of monochromatic light
  explained by optimal wavelength decoding of light of unknown intensity}
  {Human wavelength discrimination of monochromatic light explained by optimal
  wavelength decoding of light of unknown intensity}.{\BBCQ}
\newblock
\APACjournalVolNumPages{PloS one}{6}{5}{e19248}.
\PrintBackRefs{\CurrentBib}

\bibitem [\protect \citeauthoryear {%
Zoph%
\ \BBA {} Le%
}{%
Zoph%
\ \BBA {} Le%
}{%
{\protect \APACyear {2016}}%
}]{%
zoph2016neural}
\APACinsertmetastar {%
zoph2016neural}%
\begin{APACrefauthors}%
Zoph, B.%
\BCBT {}\ \BBA {} Le, Q\BPBI V.%
\end{APACrefauthors}%
\unskip\
\newblock
\APACrefYearMonthDay{2016}{}{}.
\newblock
{\BBOQ}\APACrefatitle {Neural architecture search with reinforcement learning}
  {Neural architecture search with reinforcement learning}.{\BBCQ}
\newblock
\APACjournalVolNumPages{arXiv preprint arXiv:1611.01578}{}{}{}.
\PrintBackRefs{\CurrentBib}

\end{thebibliography}

\clearpage

\end{document}